\documentclass[journal]{IEEEtran}

\usepackage{array}
\usepackage{stfloats}
\usepackage{upgreek}
\usepackage{comment}
\usepackage{relsize}

\usepackage{LaTeX_macros_bib/macros}
\usepackage{pgfplots}
\pgfplotsset{compat=1.18}
\usepackage{tikz}

\usepackage{tikzexternal}
\tikzsetexternalprefix{tikz/}
\usepackage{tikzviolinplots}
\let\|=\Vert
\usetikzlibrary{pgfplots.statistics}
\usepgfplotslibrary{fillbetween}

\usepackage[caption=false,font=footnotesize]{subfig}

\begin{document}
\bstctlcite{IEEEexample:BSTcontrol}

\title{LPAC: Learnable Perception-Action-Communication Loops with Applications to Coverage Control}

\author{Saurav Agarwal, Ramya Muthukrishnan, Walker Gosrich, Vijay Kumar, and Alejandro Ribeiro%
  \thanks{This work was supported in part by grants ARL DCIST CRA W911NF-17-2-0181, NSF 2415249, and ONR N00014-20-1-2822.}%
  \thanks{The source code for the architecture implementation and simulation platform is available at \url{https://github.com/KumarRobotics/CoverageControl}.}%
  \thanks{S.\ Agarwal, W.\ Gosrich, and V.\ Kumar are with the GRASP Laboratory, University of Pennsylvania, USA.\\E-mail: {\tt\footnotesize \{sauravag,gosrich,kumar\}@seas.upenn.edu}}%
  \thanks{R.\ Muthukrishnan is with the CSAIL, Massachusetts Institute of Technology, USA. Her contributions were made while she was affiliated with the University of Pennsylvania. E-mail: {\tt\footnotesize ramyamut@mit.edu}}%
  \thanks{A.\ Ribeiro is with the Department of Electrical and Systems Engineering, University of Pennsylvania, USA. E-mail: {\tt\footnotesize aribeiro@seas.upenn.edu}}%
}
\maketitle

\begin{abstract}
  Coverage control is the problem of navigating a robot swarm to collaboratively monitor features or a phenomenon of interest not known \textit{a priori}. The problem is challenging in decentralized settings with robots that have limited communication and sensing capabilities. We propose a learnable Perception-Action-Communication (LPAC) architecture for the problem, wherein a convolutional neural network (CNN) processes localized perception; a graph neural network (GNN) facilitates robot communications; finally, a shallow multi-layer perceptron (MLP) computes robot actions. The GNN enables collaboration in the robot swarm by computing \textit{what} information to communicate with nearby robots and \textit{how} to incorporate received information. Evaluations show that the LPAC models---trained using imitation learning---outperform standard decentralized and centralized coverage control algorithms. The learned policy generalizes to environments different from the training dataset, transfers to larger environments with more robots, and is robust to noisy position estimates. The results indicate the suitability of LPAC architectures for decentralized navigation in robot swarms to achieve collaborative behavior.

\end{abstract}
\IEEEpeerreviewmaketitle

\section{Introduction} \label{sc:intro}
\IEEEPARstart{N}{avigating} a swarm of robots through an environment to achieve a common collaborative goal is a challenging problem, especially when the sensing and communication capabilities of the robots are limited.
These problems require systems with high-fidelity algorithms comprising three key capabilities: perception, action, and communication, which are executed in a feedback loop, i.e., the Perception-Action-Communication (PAC) loop.
To seamlessly scale the deployment of such systems across vast environments with large robot swarms, it is imperative to consider a decentralized system wherein each robot autonomously makes decisions, drawing upon its own observations and information received from neighboring robots.

However, designing a navigation algorithm for a decentralized system is challenging.
The robots perform perception and action independently, while the communication module is the only component that can facilitate robot collaboration.
Under limited communication capabilities, the robots must decide \textit{what} information to communicate to their neighbors and \textit{how} to use the received information to take appropriate actions.
This article studies the coverage control problem~\cite{CortesMKB04,Cortes05} as a canonical problem for the decentralized navigation of robot swarms.
We propose a learnable PAC (LPAC) architecture that can learn to process sensor observations, communicate relevant information, and take appropriate actions.

Coverage control is the problem of collaboration in a robot swarm to provide optimal sensor coverage to a set of features or a phenomenon of interest in an environment~\cite{Cortes05}.
The relative importance of the features is modeled as an underlying scalar field known as the \textit{importance density field} (IDF)~\cite{Gosrich22}.
Coverage control with unknown features of interest has applications in various domains, including search and rescue, surveillance~\cite{doitsidis2012optimal}, and target tracking~\cite{pimenta2010simultaneous}.
Consider a critical search and rescue scenario where a robot swarm assists and monitors survivors in a disaster area.
A simple solution could be to distribute the robots in a manner that keeps them further away from each other and covers the entire environment.
However, if one or more survivors are detected, it is desirable to have a robot close to each survivor to provide vigilance, which is not captured by this simple solution.
The fundamental question, then, is how the robots should collaborate to cover the environment efficiently while accounting for the features of interest, i.e., the survivors in this example.
Similarly, coverage control can be used to monitor a continuous phenomenon of interest, such as a wildfire or soil moisture.
\begin{figure*}
  \centering
  \includegraphics[width=0.9\textwidth]{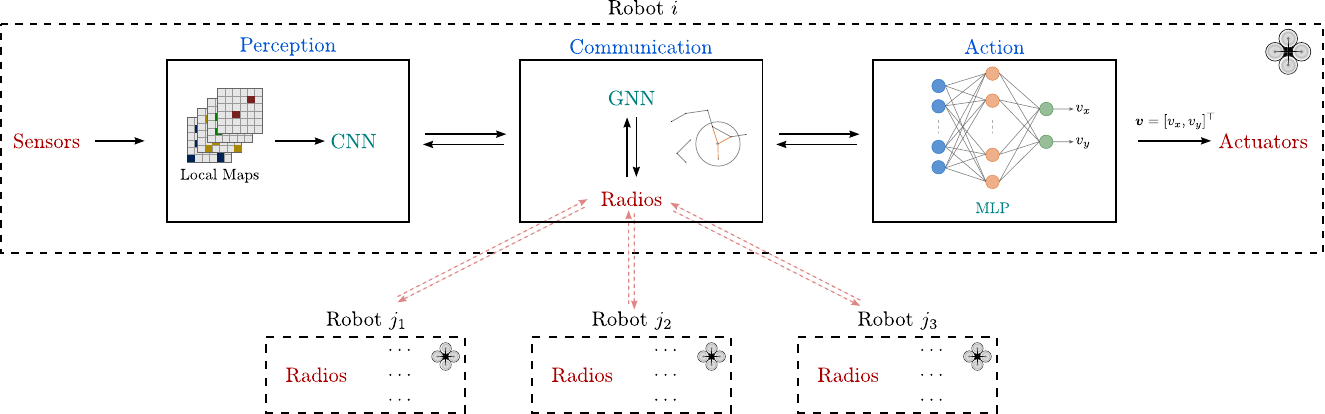}
  \caption{The proposed learnable Perception-Action-Communication (LPAC) architecture for decentralized navigation of robot swarms:
    (1)~In the perception module, a convolutional neural network (CNN) processes maps representing localized observations and generates an abstract representation.
    (2)~In the communication module, a graph neural network (GNN) performs computations on the output of the perception module and the messages received from neighboring robots.
    It generates a fixed-size message to share with nearby robots and aggregates the received information to generate a feature vector for the action module of the robot.
    (3)~In the action module, a shallow multilayer perceptron (MLP) computes the control actions for the robot based on the output generated by the GNN.
  The three modules are executed on each robot independently, with the GNN in the communication module facilitating collaboration between robots. \label{fig:learnable_pac}}
\end{figure*}

This article considers a decentralized multi-robot system with limited communication and sensing capabilities.
Furthermore, the environment is not known \textit{a priori}, and the robots use their sensors to make localized observations of the IDF in the environment.
Decentralized algorithms based on centroidal Voronoi tessellation (CVT), e.g., Lloyd's method, have been developed for robots with limited communication capabilities~\cite{Cortes05, Du99}.
However, in the absence of any prior knowledge of the IDF, such algorithms exhibit poor performance compared to their centralized counterpart.
The primary reason for this is that the robots communicate only the relative positions to their neighbors, which does not directly capture the observations of the robots.
In contrast, communicating the entire observation of a robot to its neighbors scales poorly with team size, observation size, and time.
The LPAC architecture proposed in this article addresses this issue by learning an abstract representation of the observations and communicating the relevant information to nearby robots.

The primary \textit{contribution} of the article is a learnable Perception-Action-Communication (LPAC) architecture for the decentralized coverage control problem (\fgref{fig:learnable_pac}).
The architecture is composed of three different types of neural networks, one for each module of the PAC system.
(1)~In the perception module, a convolutional neural network (CNN) processes localized IDF observations and generates an abstract representation.
(2)~In the communication module, a GNN performs computation on the output of the perception module and the messages received from neighboring robots.
It generates a fixed-size message to communicate with the neighbors and aggregates the received information to generate a feature vector for the action module of the robot.
(3)~In the action module, a shallow multilayer perceptron (MLP) predicts the control actions of the robot based on the feature vector generated by the GNN.

The LPAC architecture is trained using imitation learning with a clairvoyant centralized planner, which has access to the entire IDF and the positions of the robots in the environment.
We extensively evaluate the performance of the architecture and establish the following:
(1)~The LPAC architecture can learn to communicate the relevant information to achieve performance significantly better than both decentralized and centralized CVT-based algorithms.
(2)~The learned policy generalizes to a wide range of features and sizes of the robot swarm.
(3)~The models transfer well to larger environments with bigger robot swarms without any degradation in performance.
(4)~The policy is robust to noisy position estimates.
(5)~The model performs well on a real-world traffic light dataset without further training or fine-tuning.
These results indicate that the LPAC architecture, with a GNN as a collaboration unit, is a promising learning-based approach for the decentralized navigation of robot swarms.

The rest of the article is organized as follows.
\scref{sc:related_work} discusses the related work focusing on the graph neural networks for robot navigation and the coverage control problem.
The problem statement, in \scref{sc:problem_statement}, formalizes the decentralized navigation of robot swarms and the coverage control problem.
The LPAC architecture is presented in \scref{sc:architecture}.
The environment setup, dataset generation, and imitation learning are discussed in \scref{sc:environment}.
Extensive simulation results are presented in \scref{sc:results}.
Finally, \scref{sc:conclusion} provides concluding remarks and discusses future work.

\section{Related Work} \label{sc:related_work}
Algorithms for the navigation of robots, a fundamental problem in multi-robot systems, often need to execute a PAC loop to achieve a collective goal.
The perception module is responsible for acquiring localized sensor observations and processing them to generate a representation of the environment.
Deep learning, in particular convolutional neural networks (CNNs), has been successfully used for processing sensor observations for tasks such as mapping~\cite{tateno2017cnn}, place recognition~\cite{Oertel2020}, and visual odometry~\cite{Yang_2020_CVPR}.
However, designing a navigation algorithm for a decentralized system is challenging as the robots perform perception and action independently, while the communication module is the only component that can facilitate robot collaboration.
Graph neural networks (GNNs) have been shown to be effective in learning decentralized controllers for multi-robot systems.
We focus the related work on
\emph{(i)}~GNNs for navigation of multi-robot systems and
\emph{(ii)}~the coverage control problem as a canonical example of a decentralized navigation problem.

\subsection{Graph Neural Networks for Navigation Problems}
The main challenge in decentralized navigation problems with robots that have limited sensing and communication capabilities is to design a function that computes the information to be communicated to neighboring robots and a policy that can incorporate the received information with the local observations to make decisions.
Graph neural networks (GNNs)~\cite{KipfW17} are particularly suitable for this task as they can operate on a communication graph---the graph imposed by the communication topology of the robots---and can learn to aggregate information from neighboring robots to make decisions~\cite{TolstayaPMLR20, TolstaysPKR21}.
Furthermore, GNNs exhibit several desirable properties for decentralized systems~\cite{RuizGR21}:
\emph{(i)}~\textit{transferability} to new graph topologies not seen in the training set,
\emph{(ii)}~\textit{scalability} to large teams of robots, and
\emph{(iii)}~\textit{stability} to graph deformations.

Tolstaya et al.~\cite{TolstayaPMLR20} established the use of GNNs for learning decentralized controllers for robot swarms and demonstrated their effectiveness in the flocking problem.
They use aggregation GNNs, along with aggregated messages, as the primary architecture for learning a controller that can exploit information from distant robots using only local communications with neighboring robots.
Gama et al.~\cite{GamaDecentralized2022} proposed a framework for synthesizing GNN-based decentralized controllers using imitation learning.
They illustrate the potential of the framework by learning decentralized controllers for the flocking and path planning problems.
Building on these results, we use GNNs in the LPAC architecture as the primary module for collaboration in the coverage control problem.

GNNs have been demonstrated to be effective in several multi-robot tasks including flocking~\cite{TolstayaPMLR20, kortvelesy2021, GamaDecentralized2022,hu2022}, path planning~\cite{li2020, JiGNN21}, target tracking~\cite{zhou2022}, perimeter defense~\cite{lee2023graph}, and information acquisition~\cite{tzes2023graph}.
Gosrich et al.~\cite{Gosrich22} recently demonstrated the use of GNNs for the coverage control problem with robots that have limited sensing but assumed full communication capabilities.
A common theme in these works is the use of imitation learning to train the GNNs.
Most works use carefully hand-crafted features as the input to the GNNs~\cite{Gosrich22,TolstayaPMLR20,GamaDecentralized2022,TolstaysPKR21,JiGNN21}, which, in general, can be challenging to design for complex tasks.

Similar to our work, Li et al.~\cite{li2020} and Hu et al.~\cite{hu2022} use CNNs in conjunction with GNNs for path planning and flocking problems, respectively.
In~\cite{li2020}, CNNs are used to process a local map maintained individually by each robot using the local observations, while in~\cite{hu2022}, the processing is done on raw images in simulated environments.
Using raw images is not always desirable, as a CNN trained in a simulated environment may not generalize to real-world environments.
Furthermore, raw images capture the current state of the environment but not the history of the environment.
Since it is common to maintain a local map in navigation problems, using a CNN to process local maps~\cite{li2020} is more generalizable and can be applied to different hardware platforms and sensors.

Recently, ROS-based frameworks have been developed for testing and deploying decentralized GNN-based policies~\cite{BlumenkampGNN2022, agarwal2023asynchronous}.
While Blumenkamp et al.~\cite{BlumenkampGNN2022} focus on GNNs and different types of network setups, Agarwal et al.~\cite{agarwal2023asynchronous} generalize to asynchronous and decentralized implementation of LPAC architectures.
Despite several important applications of GNNs in multi-robot systems, detailed study of the transferability, generalizability, and robustness properties of GNN-based architectures has been limited, especially when combined with CNN.
In this article, we study these properties of the LPAC architecture in the context of the coverage control problem.

\subsection{Multi-Agent Reinforcement Learning (MARL)}

A joint space representation does not scale well with large number of robots and makes it challenging to train and deploy deep neural network policies.
Foerster et al.~\cite{foerster2016learning} provided an RL framework to explicitly learn communication protocols.
The approach uses Q-learning with the help of lookup tables, and is demonstrated on switch riddle and MNIST games.
G\'omez et al.~\cite{Gomez20} provided an MARL framework for multi-robot systems where they learn to decide on which agents to communicate with.
The framework is integrated with non-linear model predictive control to solve collision avoidance.
Although the robots have independent trajectories, the system assumes that the robots have complete observation of the positions and velocities of all other robots.
MARL has also been used for coverage control~\cite{Meng22} using multi-agent deep deterministic policy gradient.
However, they assume positions of all robots are known to each agent, do not perform explicit communication, and limited to $10\times10$ environments.
In these MARL work, there has been a lack of extensive study on the scalability of the policies to large number of agents.

The theoretical properties of GNNs on transferability, scalability and stability render them more suitable for decentralized systems than MARL framework.
Furthermore, imitation learning often helps in converging to a good policy faster than exploring a large action space in the reinforcement learning framework.
An interesting approach would be to use imitation learning for an initial policy and then fine-tune using RL frameworks to improve performance.

\subsection{Coverage Control in Multi-Robot Systems}
The coverage control problem is a well-studied problem in multi-robot systems and is often used as a benchmark problem for evaluating decentralized controllers.
The problem is closely related to the sensor coverage problem~\cite{wang11} and the facility location problem~\cite{SuzukiDrezner96}.
Cort\'{e}s et al.~\cite{Cortes05} were the first to propose decentralized control algorithms for the problem with robots that have limited sensing and communication capabilities.
The algorithms iteratively use the centroidal Voronoi tessellation (CVT)~\cite{Du99,Edelsbrunner85} of the environment to assign each robot to a region of dominance and perform gradient descent on a cost function to converge to a local minimum.
The algorithms are based on the Lloyd algorithm~\cite{Lloyd82} from quantization theory.

Several standard approaches build upon the work of Cort\'{e}s et al.~\cite{CortesMKB04,Cortes05} and use CVT-based algorithms for the coverage problem.
Different sensor models have been used in the literature, e.g., limited sensor radius identical among all agents~\cite{Cortes05, pratissoli2022limitedrange}, heterogeneous sensing capabilities~\cite{kwok2010distributed}, and heterogeneous disk sector shaped sensor models~\cite{parapari2016, zhong2019}.
This article considers robots with a limited sensor radius identical to all robots.
Sensor models with a footprint in the shape of a disk sector need to include the orientation of the sensor in the state space.
Although we do not consider such a model, the LPAC architecture can be extended to include the sensor orientation.

Similar to most of the work in the literature, we assume that the robots have sensors or processing capabilities to obtain the IDF in the sensor field of view.
However, IDF estimation techniques based on consensus learning~\cite{Schwager08Concensus} and using Gaussian processes~\cite{Wei2021,Santos2021} have been studied.
Including such techniques in the LPAC architecture is an interesting direction for extending the work.

Incorporating explicit exploration strategies can improve the performance of coverage control algorithms at the cost of additional time spent strategically exploring the environment~\cite{Schwager08Ladybug,mirzaei2011cooperative}.
Exploration and coverage are opposing objectives, and the trade-off between the two objectives is a challenging problem.
A dual LPAC architecture that can switch between exploration and coverage modes may be a promising direction for this problem.
Other works have considered time-varying density fields~\cite{Lee2015,santos2019decentralized} for which incorporating an attention-based architecture~\cite{VaswaniSPUJGKP17} in the LPAC architecture may be beneficial.
In this article, we restrict to the standard coverage control problem with no explicit exploration and a static density field.

Coverage control has been used in real-world applications such as monitoring of algae blooms~\cite{couceiro2019}, agricultural fields~\cite{luo2019distributed}, indoor environments~\cite{luo2018adaptive}, surveillance~\cite{doitsidis2012optimal}, and target tracking~\cite{pimenta2010simultaneous}.
In this work, we evaluate the LPAC architecture in coverage control of semantic-derived real-world data; we use traffic signal data from 50 cities in the United States to model the importance density field of the environment.
The wide range of applications of coverage control has been the primary motivation for the development of decentralized controllers for the problem.
Hence, we use the coverage control problem to evaluate the LPAC architecture.

\section{Problem Statement} \label{sc:problem_statement}
The multi-robot coverage control problem belongs to the class of navigation control problems.
A general navigation control problem is defined on a $d$-dimensional environment $\mc{W} \subset \mathbb{R}^d$.
We are given a homogeneous team of $N$ robots $\mc V=\{1,\ldots,N\}$ with $\mv x_i(t) \in \mathbb{R}^{d_x}$ representing the state of robot $i$ at time~$t$.
Along with the position of the robot, the state can also include other information such as velocity, acceleration, and sensor observations.
Similarly, the control input for robot $i$ at time $t$ is denoted by $\mv u_i(t) \in \mathbb{R}^{d_u}$ in a $d_u$-dimensional control space.
The collective state and control for the entire system can be represented as:
\begin{equation*}
  \mv X(t) = \begin{bmatrix}
    \mv x_1(t) \\ \mv x_2(t) \\ \vdots \\ \mv x_N(t)
  \end{bmatrix} \in \mathbb R^{N\times d_x}, \quad
  \mv U(t) = \begin{bmatrix}
    \mv u_1(t) \\ \mv u_2(t) \\ \vdots \\ \mv u_N(t)
  \end{bmatrix} \in \mathbb R^{N\times d_u}.
\end{equation*}
Assuming a static environment, given a state $\mv X(t)$ and a control $\mv U(t)$, the state of the system evolves according to the following Markov model $\mathbb P$ for time step $\Delta t$:
\begin{equation*}
  \mv X(t+\Delta t) = \mathbb P(\mv X \mid \mv X(t), \mv U(t)).
\end{equation*}
In a centralized setting, the navigation control problem is posed as an optimization problem with a cost function $\mc J(\mv X(t), \mv U(t))$.
Its goal is to find a centralized control policy $\Pi_c^*$ that minimizes the expected cost~\cite{GamaDecentralized2022}:
\begin{equation*}
  \Pi_c^* = \argmin_{\Pi_c} \mathbb E \left[ \sum_{t=0}^T\gamma^t \mc J(\mv X(t), \mv U(t)) \right].
\end{equation*}
Here, the control actions are drawn from the policy as $\mv U(t) = \Pi_c(\mv U \mid \mv X(t))$.
The discount factor is given by $\gamma \in [0,1)$, and $T$ is the total time for which the policy is executed.
A centralized policy requires complete knowledge of the state of the system $\mv X(t)$, and all robots need to communicate with the central controller, resulting in a bottleneck in the system.
The computation of the control actions is also centralized, which can be computationally expensive and does not scale well with a large number of robots.
Furthermore, a central controller is not robust because of a single point of failure.
This motivates us to study the decentralized navigation control problem.

\subsection{Decentralized Navigation Control}
\label{sec:decentralized}
In a decentralized setting, each robot computes its own control action based on its own state and the states of nearby robots.
The robots can only communicate with other robots that are within a limited communication radius~$r_c$.
The problem can now be formulated on a \textit{communication graph} $\mc G = (\mc V, \mc E)$, where $\mc V$ is the set of vertices representing the robots and $\mc E$ is the set of edges representing the communication topology.
An edge $(i,j) \in \mc E$ exists if and only if robots $i$ and $j$ can communicate with each other, i.e., when $\|\mv p_i(t) - \mv p_j(t)\| \le r_c$.
The communication graph is assumed to be undirected, i.e., $(i,j) \in \mc E \iff (j,i) \in \mc E$.
The neighbors of robot $i$ are the robots that it can communicate with and are denoted by $\mc N(i) = \{j \mid (i,j) \in \mc E\}$.

Let $\mc X_i(t)$ be the total information acquired by robot $i$ through its own observations and through communication with its neighbors. 
In general, a robot may communicate an abstract representation of its state to its neighbors, i.e., $\mc X_i(t)$ need not be a simple aggregation of the states.
A decentralized policy leverages only this locally available information, $\mc{X}_i(t)$.
Unlike the centralized policy, the decentralized policy is executed by each robot independently and does not require communication with a central controller, thereby mitigating the issues of robustness and scalability.

\subsection{Coverage Control Problem}
The coverage control problem is a navigation control problem where the goal is to provide sensor coverage based on the importance of information at each point in the environment.
An importance density field (IDF) $\Phi:\mc{W} \mapsto	 \mathbb{R}_{\geq 0}$ is defined over a 2-D environment $\mc W\subset \mathbb R^2$.
The IDF represents a non-negative measure of importance at each point in the environment.
With the state of a robot~$i$ given by its position $\mv p_i\in \mc W$ in the environment, the control actions given by the velocity $\dot{\mv p}_i(t)$, and $\Delta t$ as the time step, we use the following model for the state evolution:
\begin{equation}
  \mv p_i(t+\Delta t) = \mv p_i(t) + \dot{\mv p}_i(t) \Delta t.
\end{equation}

The cost function for the coverage control problem is defined as:
\begin{equation}
  \mc J\left(\mv X\right) = \int_{\mv q \in \mc W} \min_{i\in\mc V}  f(\|\mv p_i - \mv q\|) \Phi(\mv q)\,d\mv q.
  \label{eq:coverage_gen}
\end{equation}
Here, $f$ is a non-decreasing function, and a common choice is $f(x) = x^2$ as it is a smooth function and is easy to optimize.
We drop the time index $t$ here and in the rest of the article for convenience.

Assuming that no two robots can occupy the same point in the environment, the Voronoi partition~\cite{deberg} can be used to assign each robot a distinct portion of the environment to cover.
The Voronoi partition $\mc P$ is defined as:
\begin{equation}
  \begin{split}
    \mc P &= \{P_i \mid i \in \mc V\}, \quad \text{where }\\
    P_i &= \{\mv q \in \mc W \mid \|\mv p_i - \mv q\| \le \|\mv p_j - \mv q\|, \forall j \in \mc V\}.
  \end{split}
  \label{eq:voronoi}
\end{equation}
All points in $P_i$ are closer to robot~$i$ than any other robot.
The cost function~\eqref{eq:coverage_gen} can now be expressed in terms of the Voronoi partition as:
\begin{equation}
  \mc J\left(\mv X\right) = \sum_{i=1}^N \int_{\mv q \in P_i} f(\|\mv p_i - \mv q\|) \Phi(\mv q)\,d\mv q.
  \label{eq:coverage_voronoi}
\end{equation}
The cost function~\eqref{eq:coverage_voronoi} is a sum of integrals over disjoint regions and is much easier to compute and optimize than the original function~\eqref{eq:coverage_gen}.
Furthermore, if the Voronoi partition is known, the cost function can be computed in a decentralized manner, as each robot only needs to compute the integral over its own region $P_i$.

We can now define the coverage control problem in the context of the navigation control problem:
Find a decentralized control policy $\Pi$ that minimizes the expected cost $\mc J(\mv X)$~\eqref{eq:coverage_voronoi}.
The policy $\Pi$ is defined over a space of all possible velocities, and each robot independently executes the same policy.

\begin{figure}[t]
  \includegraphics[width=0.9\columnwidth]{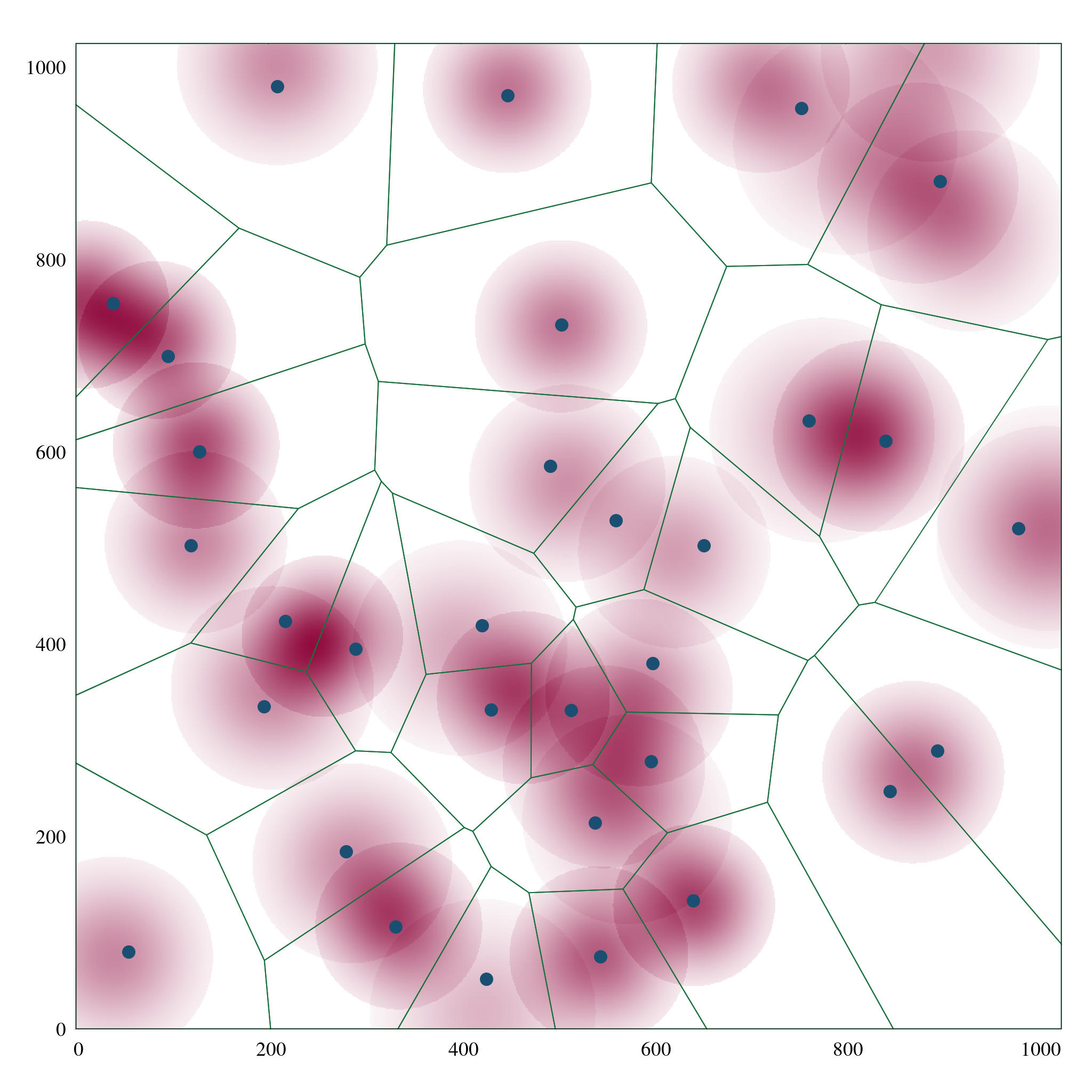}
  \caption{A near-optimal solution to the coverage control problem:
    A team of 32 robots is deployed in an environment of size 1024\SIm{}$\times$1024\SIm{}.
    There are 32 features represented as Gaussians to represent the importance density field (IDF).
    Robots position themselves to provide sensor coverage to the features of interest.
    The green lines represent the Voronoi partition of the environment with respect to the robot positions.
    A robot is closer to all points in its Voronoi region than any other robot.
  \label{fig:coveragecontrol_global}}
\end{figure}

\fgref{fig:coveragecontrol_global} shows a near-optimal solution, along with Voronoi partition, to an instance of the coverage control problem with 32 robots and 32 features in a 1024\SIm{}$\times$1024\SIm{} environment.

In this article, we consider the decentralized coverage control problem with the following restrictions:
(R1)~At any time, each robot can make localized observations of the IDF within a sensor field of view (FoV).
(R2)~The IDF $\Phi$ is static and is not known \textit{a priori}.
(R3)~Each robot can maintain only its own localized observations of the IDF aggregated over time.
(R4)~The robots can only communicate with other robots that are within a limited communication radius.

Additionally, our simulation environment assumes, without loss of generality:
(R5)~The boundary of the environment is convex to allow the use of centroidal Voronoi tessellation (CVT) for imitation learning and evaluation.
This can be relaxed by using coverage control algorithms for non-convex environments~\cite{pimenta2008}.
(R6)~The IDFs are generated using standard random 2D Gaussian functions truncated at 2$\sigma$ (see \scref{sc:dataset_generation}).
This is representative of applications where the importance of information is localized around certain points in the environment and is reduced with distance from these points.
However, the overall approach is general and can potentially be applied to other types of IDFs.

In such a setting, a coverage control algorithm needs to provide the following based on the state of robot $i$ and the information received from its neighbors~$\mc N(i)$:
\begin{enumerate}
  \item A function $\mc I$ that computes the information, in the form of messages, to be communicated by robot $i$ to its neighbors, and 
  \item A common policy $\Pi$ that computes the control action $\mv u_i = \dot{\mv p}_i$ for any robot~$i \in \mc V$.
\end{enumerate}

Designing such decentralized algorithms is challenging and can be intractable for complex systems~\cite{Witsenhausen68}.
This motivates us to use a learning-based approach to design a decentralized coverage control algorithm.
As we will see in \scref{sc:architecture}, the LPAC architecture with GNN addresses the above challenges and provides a scalable and robust solution to the problem.

\begin{remark}
  Let $A_i\subset \mathcal W$ denote the current sensor FoV.
  Then, the cost function can be formulated such that only the current observation is considered, following~\cite{CortesMKB04,Cortes05,Gosrich22,pratissoli2022limitedrange}:
  \begin{equation}
    \mc J\left(\mv X\right) = \sum_{i=1}^N \int_{\mv q \in (P_i\cap A_i)} f(\|\mv p_i - \mv q\|) \Phi(\mv q)\,d\mv q.
    \label{eq:coverage_curr}
  \end{equation}
  Alternatively, under the restrictions of a static IDF (R2) and each robot maintaining its own observations (R3), the cost function can be stated in terms of the observed workspace, $\mathcal W_o\subseteq \mathcal W$:
  \begin{equation}
    \mc J\left(\mv X\right) = \sum_{i=1}^N \int_{\mv q \in (P_i\cap \mathcal W_o)} f(\|\mv p_i - \mv q\|) \Phi(\mv q)\,d\mv q.
    \label{eq:coverage_cumm}
  \end{equation}
  However, the above cost functions ignore the underlying IDF, which hasn't been completely observed.
  Although the robots have access to only the observed IDF, they can still learn to provide more efficient coverage by exploiting its underlying structure.
  Hence, we evaluate the proposed LPAC architecture on the entire IDF, as given by~\eqref{eq:coverage_voronoi}.
  As discussed in \scref{sc:CVT}, we use a clairvoyant algorithm that has knowledge of the entire IDF for imitation learning, and in \scref{sc:results}, the evaluations show that the LPAC architecture performs significantly better than CVT algorithms in terms of the global cost function~\eqref{eq:coverage_voronoi}.
\end{remark}

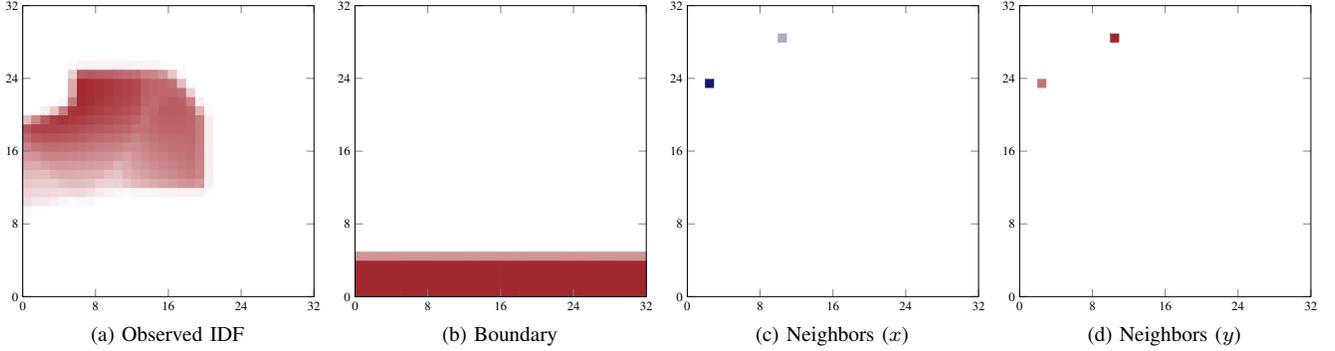
\begin{figure*}[htbp]
  \centering
  \subfloat[Observed IDF]{\begin{tikzpicture}[scale=0.68]
	\pgfplotsset{
		colormap={violet}{color=(white) color=(mDarkRed)}
	}
	\scriptsize
	\begin{axis}[
		xmin=-0.5,
		xmax=31.5,
		ymin=-0.5,
		ymax=31.5,
		xtick={-0.5,7.5,...,31.5},
		ytick={-0.5,7.5,...,31.5},
        xticklabels={0,8,...,32},
        yticklabels={0,8,...,32},
		enlargelimits=false,
		axis on top,
		axis equal image,
		]
		\addplot [matrix plot*,point meta=explicit] table [x=x,y=y,meta=C] {./figures/data/local_maps.csv};
	\end{axis}
\end{tikzpicture}}
  \subfloat[Boundary]{\begin{tikzpicture}[scale=0.68]
	\pgfplotsset{
		colormap={violet}{color=(white) color=(mDarkRed)}
	}
	\scriptsize
	\begin{axis}[
		xmin=-0.5,
		xmax=31.5,
		ymin=-0.5,
		ymax=31.5,
		xtick={-0.5,7.5,...,31.5},
		ytick={-0.5,7.5,...,31.5},
        xticklabels={0,8,...,32},
        yticklabels={0,8,...,32},
		enlargelimits=false,
		axis on top,
		axis equal image,
		]
		\addplot [matrix plot*,point meta=explicit] table [x=x,y=y,meta=C] {./figures/data/obstacle_maps.csv};
	\end{axis}
\end{tikzpicture}}
  \subfloat[Neighbors ($x$)]{\begin{tikzpicture}[scale=0.68]
	\pgfplotsset{
		colormap={violetr}{color=(mDarkBlue) color=(white)}
	}
	\scriptsize
	\begin{axis}[
		xmin=-0.5,
		xmax=31.5,
		ymin=-0.5,
		ymax=31.5,
		xtick={-0.5,7.5,...,31.5},
		ytick={-0.5,7.5,...,31.5},
        xticklabels={0,8,...,32},
        yticklabels={0,8,...,32},
		enlargelimits=false,
		axis on top,
		axis equal image,
		]
		\addplot [matrix plot*,point meta=explicit] table [x=x,y=y,meta=C] {./figures/data/comm_maps_x.csv};
	\end{axis}
\end{tikzpicture}}
  \subfloat[Neighbors ($y$)]{\begin{tikzpicture}[scale=0.68]
	\pgfplotsset{
		colormap={violet}{color=(white) color=(mDarkRed)}
	}
	\scriptsize
	\begin{axis}[
		xmin=-0.5,
		xmax=31.5,
		ymin=-0.5,
		ymax=31.5,
		xtick={-0.5,7.5,...,31.5},
		ytick={-0.5,7.5,...,31.5},
        xticklabels={0,8,...,32},
        yticklabels={0,8,...,32},
		enlargelimits=false,
		axis on top,
		axis equal image,
		]
		\addplot [matrix plot*,point meta=explicit] table [x=x,y=y,meta=C] {./figures/data/comm_maps_y.csv};
	\end{axis}
\end{tikzpicture}}
  \caption{The four channels of the CNN input image in the perception module.
    All channels are ego-centric to the robot and are of size 32$\times$32.
    The first channel (a) represents the IDF observed by the robot in its local vicinity.
    The second channel (b) represents the boundary of the environment.
    They have non-zero values only when the robot is close to the boundary.
    The third (c) and fourth (d) channels represent the positions of the neighbors of the robot.
    For each neighbor, the pixels in the channels corresponding to the relative position of the neighbor have non-zero values.
    The channel (c) represents the $x$-coordinates of the neighbors, and the other channel (d) represents the $y$-coordinates, both normalized by the communication range.
  \label{fig:maps}}
\end{figure*}

\section{Learnable PAC Architecture} \label{sc:architecture}
We propose a learnable Perception-Action-Communication (LPAC) architecture with a Graph Neural Network (GNN) for the coverage control problem.
The three modules of the LPAC architecture are executed on each robot in a decentralized manner.
The input to the architecture is the local IDF observed by the robot, and the output is the velocity control actions for the robot.
The collaboration in the robot swarm is achieved using the communication module, which is responsible for exchanging information between robots.
The following subsections describe the three modules of the LPAC architecture for the coverage control problem in detail.

\subsection{Perception Module} \label{sc:perception}
The perception module is composed of a convolutional neural network (CNN) that takes a four-channel image as input and generates a 32-dimensional feature vector.
The channels are grids of pre-defined size (\sqrdim{32}) and are centered around the current position of the robot.
The first and the second channels are local representations of the importance and boundary maps, respectively.
The other two channels, neighbor maps, encode the relative positions of the neighboring robots.
\fgref{fig:maps} shows an example of the four-channel image, which forms the input to the CNN.

The perception module is responsible for processing data acquired by robot sensors with a limited field of view.
We maintain an \textit{importance map} to represent the IDF sensed by the robot.
Each pixel in the importance map is assigned an importance value if it has been sensed by the robot and is assigned a value of 0 otherwise.
The first channel of the input image is generated by extracting a local map of a pre-defined size (\sqrdim{256}) from the importance map centered around the current position of the robot and, thereafter, downsampling it to the size of the channel (\sqrdim{32}) using bilinear interpolation.

The perception module also maintains a \textit{boundary map} to represent the boundaries of the environment.
Regions outside the environment boundaries are assigned a value of 1, and regions inside the environment boundaries are assigned a value of 0.
The boundary map is used so that the policy can learn to avoid collisions with the environment boundaries.
Similar to the importance map, the boundary map represents a local region of a predefined size (\sqrdim{256}) centered at the current position of the robot.
This map is also downsampled to the size of the channel (\sqrdim{32}) using bilinear interpolation.
In general, an obstacle map, similar to the boundary map, can be maintained to represent the obstacles in the environment.
Such a map can be generated using SLAM or other obstacle detection techniques.

The other two channels, \textit{neighbor maps}, encode the relative positions of the neighboring robots.
These relative positions can be obtained by the sensors on the robot or by exchanging information with the neighboring robots using the communication module.
The resolution of the channel is set to the ratio of the communication range and the size of the channel.
A pixel, subject to the resolution constraint, is assigned a value if a neighboring robot is present in the corresponding region and is assigned a value of 0 otherwise.
A large communication range relative to the channel size makes the resolution very coarse, which, in turn, results in a loss of accuracy of the relative positions of neighboring robots.
To mitigate this issue, we have two channels for the neighbor maps; one channel encodes the relative $x$ coordinate, while the other is for the relative $y$ coordinate.
We normalize the relative coordinates by the communication range.
If two or more neighboring robots are present in the same region of a pixel, the pixel is assigned the sum of the normalized relative coordinates of the neighboring robots.
The neighbor maps are designed in this way to ensure that they are \textit{permutation invariant}, i.e., the relative positions of the neighboring robots are independent of the order in which they are encoded in the map.

These four channels are concatenated to form the input to the CNN.
The CNN is composed of three sequences of a convolutional layer followed by batch normalization~\cite{IoffeS15BatchNorm} and a leaky ReLU pointwise nonlinearity.
The convolution layers have a kernel size of \sqrdim{3} and a stride of one with zero padding and generate 32 output channels.
The output of the last sequence is flattened and passed through a linear layer with a leaky ReLU activation to generate a 32-dimensional feature vector.
The feature vector is then sent to the communication module for further processing.

\subsection{Communication with Graph Neural Networks} \label{sec:gnn}
The communication module is responsible for exchanging information between robots and enabling collaboration in the robot swarm.
The main component of the module is a GNN, which is a layered information processing architecture that operates on graphs and makes inferences by diffusing information across the graph.
In the LPAC architecture, the graph for GNN is defined by the communication graph of the robot swarm.
The vertices $\mc V$ of the graph correspond to the robots, and the edges $\mc E$ represent the communication links between the robots, as discussed in \scref{sec:decentralized}.

\begin{figure}[htbp]
  \centering
  \usetikzlibrary{backgrounds}
\tikzstyle{Arc} = [thick,-{Latex[length=2mm,width=1.8mm]}]
\begin{tikzpicture}[scale=0.3]
	\footnotesize
	\def\horlen{0.8}
	\def\vertlen{1.6}
	\def\halfvert{0.9}
	\tikzstyle{inblock}=[draw=none, fill=none, align=center, inner sep=0.1cm]
	\tikzstyle{gnnblock}=[draw=mDarkRed, ultra thick, inner sep=0.2cm, fill=none, node distance=1.5cm, minimum width=1.5cm, text width=3.3cm, align=center,minimum height=1.2cm]
	\tikzstyle{layer}=[draw=none, fill opacity=0.1, inner sep=0.30cm, fill=mSteelGray]

	\node (in) [inblock] {$\mv X=\mv X_0$};

	\node (z1) [gnnblock, below right=0.6 cm and -1.0 cm of in] {$\mv Z_1 = \sum\limits_{k=0}^K (\mv S)^k \mv X_0 \mv H_{1k}$};
	\node (x1) [gnnblock, draw=mBlue, right=\horlen cm of z1] {$\mv X_1 = \sigma(\mv Z_1)$};
	\begin{scope}[on background layer]
		\node[layer,fit=(x1) (z1)] (l1) {};
		\path (l1.south east) ++ (0.0,-0.45) node [above left] {\scriptsize Layer 1};
	\end{scope}
	\draw[Arc] (in.east) -| (z1);
	\draw[Arc] (z1) -- (x1);

	\node (z2) [gnnblock, below=\vertlen cm of z1] {$\mv Z_2 = \sum\limits_{k=0}^K (\mv S)^k \mv X_1 \mv H_{2k}$};
	\node (x2) [gnnblock, draw=mBlue, right=\horlen cm of z2] {$\mv X_2 = \sigma(\mv Z_2)$};
	\begin{scope}[on background layer]
		\node[layer,fit=(x2) (z2)] (l2) {};
		\path (l2.south east) ++ (0.0,-0.45) node [above left] {\scriptsize Layer 2};
	\end{scope}
	\path (x1.south) -- (z2) coordinate[midway] (b);
	\draw[Arc] (x1.south) |- (b) -| (z2);
	\draw[Arc] (z2) -- (x2);


	\node (z3) [gnnblock, below=\vertlen cm of z2] {$\mv Z_3 = \sum\limits_{k=0}^K (\mv S)^k \mv X_{2} \mv H_{3k}$};
	\node (x3) [gnnblock, draw=mBlue, right=\horlen cm of z3] {$\mv X_3 = \sigma(\mv Z_3)$};
	\begin{scope}[on background layer]
		\node[layer,fit=(x3) (z3)] (l3) {};
		\path (l3.south east) ++ (0.0,-0.45) node [above left] {\scriptsize Layer $3$};
	\end{scope}
    \path (x2.south) -- (z3) coordinate[midway] (b);
	\draw[Arc] (x2) |- (b) -| (z3);
	\draw[Arc] (z3) -- (x3);
	\node (out) [inblock,below=0.8cm of x3] {$\mv X_3=\Psi (\mv X; \mv S, \mc H)$};
	\draw[Arc] (x3) -- (out);
\end{tikzpicture}
  \caption{An example of the GNN architecture used in the communication module.
  The architecture is composed of $L=3$ layers of graph convolution filters (red boxes) followed by pointwise nonlinearities (blue boxes).}
  \label{fig:gnn}
\end{figure}
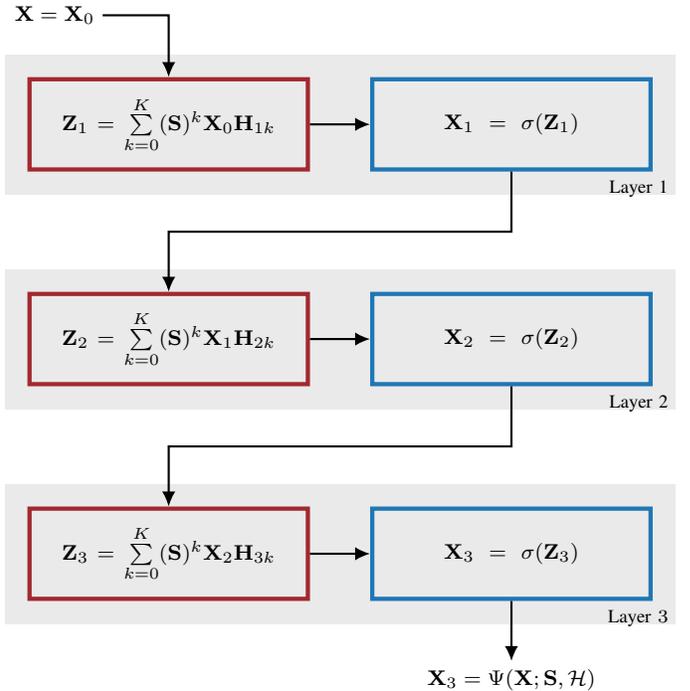

Our GNN architecture is a layered composition of $L$ \textit{graph convolution filters}~\cite{RuizGR21} with ReLU as the pointwise nonlinearity, as shown in \fgref{fig:gnn}.
Each graph convolution filter is parameterized by $K$ hops of message diffusion and is a polynomial function of the \textit{shift operator} $\mv S\in \mathbb R^{N\times N}$, which is a matrix representation of the communication graph for $N$ robots.
The elements $[\mv S]_{ij}$ can be non-zero only if $(i, j)\in \mc E$.
There are many ways to define the shift operator~$\mv S$, such as the adjacency matrix, the Laplacian matrix, or the normalized Laplacian matrix of the communication graph.
In our models, we use the normalized adjacency matrix as the shift operator, which is defined as:
\begin{equation}
  \mv S = \mv D^{-1/2} \mv A \mv D^{-1/2}.
\end{equation}
Here, $\mv A$ is the adjacency matrix, and $\mv D$ is the diagonal degree matrix.

The input to the GNN is a collection of features $\mv X_0\in \mathbb R^{N\times d_0}$, where each row $\mv x_i$ is the output of the perception module, augmented with the position of the robot normalized by the environment size, for robot $i\in \mc V$ and $d_0$ is the dimension of the feature vector.
The learning weight parameters of the GNN are given by $\mv H_{lk}\in\mathbb R^{d_{(l-1)}\times d_{l}},\,\forall l\in\{1,\cdots,L\},\,\forall k\in\{1,\cdots,K\}$, where $d_l$ is the dimension of layer $l$.
The output $\mv Z_l$  of the convolution layer is a polynomial function of the shift operator $\mv S$ and is processed by the pointwise nonlinearity to generate the input to the next layer:
\begin{equation}
  \mv Z_l = \sum_{k=0}^K (\mv S)^k \mv X_{l-1} \mv H_{lk}, \quad \mv X_l = \sigma(\mv Z_l).
\end{equation}
Here, we set $(\mv S)^0$ as the identity matrix of size $N\times N$.
The final output of the GNN is $\mv X_L$, and the architecture is denoted by $\Psi(\mv X; \mv S,\mc H)$.
In our models, we use $L=5$ graph convolution layers with $d_0=34$ and $d_l=256,\,\forall l\in\{1,\cdots,L\}$.

\begin{figure*}[htbp]
  \centering
  \subfloat[Communication graph]{\begin{tikzpicture}[scale=0.24]
	\small
	\tikzstyle{scMid}=[midway, fill=none, black]
	\tikzstyle{scSt}=[near start, fill=none, black]
	\tikzstyle{circb}=[circle, draw=mDarkRed, very thick, fill=none, inner sep=0.4pt, minimum size=0.5cm]
	\node (ri) [circb, draw=mDarkRed] at (0,0)  {$i$};
	\coordinate (ric) at (ri);
	\node (r1) [circb, draw=teal] at ($(ric) + (-0.0, -11.0)$) {$1$};
	\node (r4) [circb, draw=gray] at ($(ric) + (-11.0, -20.0)$) {$4$};
	\node (r3) [circb, draw=teal] at ($(ric) + (-10.0, -1.0)$) {$3$};
	\node (r2) [circb, draw=teal] at ($(ric) + (-6.0, -8.0)$) {$2$};
	\node (r5) [circb, draw=gray] at ($(ric) + (-6.0, -16.0)$) {$5$};
	\node (r6) [circb, draw=gray] at ($(ric) + (-15.0, --1.0)$) {$6$};
	\draw [Arc,draw=teal,thick] (r1) -- node[scMid,above right] {$\mv Y_1$} (ri);
	\draw [Arc,draw=teal,thick] (r2) -- node[scMid,near end,left] {$\mv Y_2$} (ri);
	\draw [Arc,draw=teal,thick] (r3) -- node[scMid,above right] {$\mv Y_3$} (ri);
	\draw  (r1) -- (r5);
	\draw  (r2) -- (r1);
	\draw  (r2) -- (r3);
	\draw  (r4) -- (r5);
	\draw  (r3) -- (r6);
	\draw  (r2) -- (r5);
\end{tikzpicture}}\hspace{1cm}%
  \subfloat[Graph convolution filter (red box) with pointwise nonlinearity (blue box)]{\tikzstyle{ArcThin} = [thick,-{Latex[length=1.4mm,width=1.2mm]}]
\begin{tikzpicture}[scale=0.2]
	\small
	\def\horlen{0.8}
	\def\vertlen{1.6}
	\def\halfvert{0.9}
	\tikzstyle{inblock}=[draw=none, fill=none, align=center, inner sep=0.1cm]
	\tikzstyle{gnnblock}=[draw=mDarkRed, ultra thick, inner sep=0.2cm, node distance=1.5cm, minimum width=1.5cm, align=center,minimum height=1.0cm]

	\tikzstyle{circ}=[circle, draw=teal, very thick, fill=none, inner sep=1.5pt, minimum size=0.7cm]

	\node (xk) [inblock] {$(\mv y_i)_{l0} = (\mv x_i)_{l-1}$};
	\node (yk) [gnnblock, fill=none, right=\horlen cm of xk] {$\begin{aligned}
	(\mv z_i)_l &= \sum\limits_{k=0}^K(\mv y_i)_{lk} \mv H_{lk}\\
	(\mv y_i)_{lk} &= \smashoperator{\sum\limits_{j\in \{1,2,3\}}} s_{ij} (\mv y_j)_{l(k-1)}
\end{aligned}$};
	\node (zk) [gnnblock,draw=mBlue, right=\horlen cm of yk] {$\sigma\Big((\mv z_i)_l\Big)$};
	\node (xka) [right=\horlen cm of zk] {$(\mv x_i)_l$};
	\draw[Arc] (xk) -- (yk);
	\draw[Arc] (yk) -- (zk);
	\draw[Arc] (zk) -- (xka);
	\node (j2) [circ, above=0.6 cm of yk] {$\mv Y_2$};
	\coordinate (twofive) at ($(yk.north west)!0.25!(yk.north east)$);
	\coordinate (sevenfive) at ($(yk.north west)!0.75!(yk.north east)$);
	\node (j1) [circ, above=0.6 cm of twofive] {{\footnotesize $\mv Y_1$}};
	\node (j3) [circ, above=0.6 cm of sevenfive] {{\footnotesize $\mv Y_3$}};
	\node (yi) [circ, below=0.6 cm of yk, draw=mDarkRed] {{\footnotesize $\mv Y_i$}};
	\node (yitext) [inblock,right=0.2 cm of yi] {{\footnotesize $=\left\{(\mv y_i)_{lk}\mid k\in\{0,\ldots,K-1\},\ l\in\{1,\ldots,L\}\right\}$}};
	\draw[ArcThin, draw=teal] (j2) -- (yk);
	\draw[ArcThin, draw=teal] (j1) -- (twofive);
	\draw[ArcThin, draw=teal] (j3) -- (sevenfive);
	\draw[ArcThin, draw=mDarkRed] (yk) -- (yi);
\end{tikzpicture}}%
  \caption{Distributed implementation of the GNN architecture for a robot $i$.
    (a)~The communication graph highlights the neighboring robots, i.e., $\mc N(i) = \{1,2,3\}$.
    The robot $i$ receives aggregated messages $\mv Y_j=\{(\mv y_j)_{lk}\}, \forall j\in\{1,2,3\}$ from the neighboring robots.
    (b)~For a layer $l$ of the GNN, the output of the previous layer $(\mv x_i)_{l-1}$ and the aggregated messages are processed by the graph convolution filter to generate the output $(\mv z_i)_l$, which is then processed by the pointwise nonlinearity to generate the output $(\mv x_i)_l$.
    The convolution filter also generates the aggregated message $\mv Y_i$ to be sent to the neighboring robots.
  \label{fig:gnn_dist}}
\end{figure*}
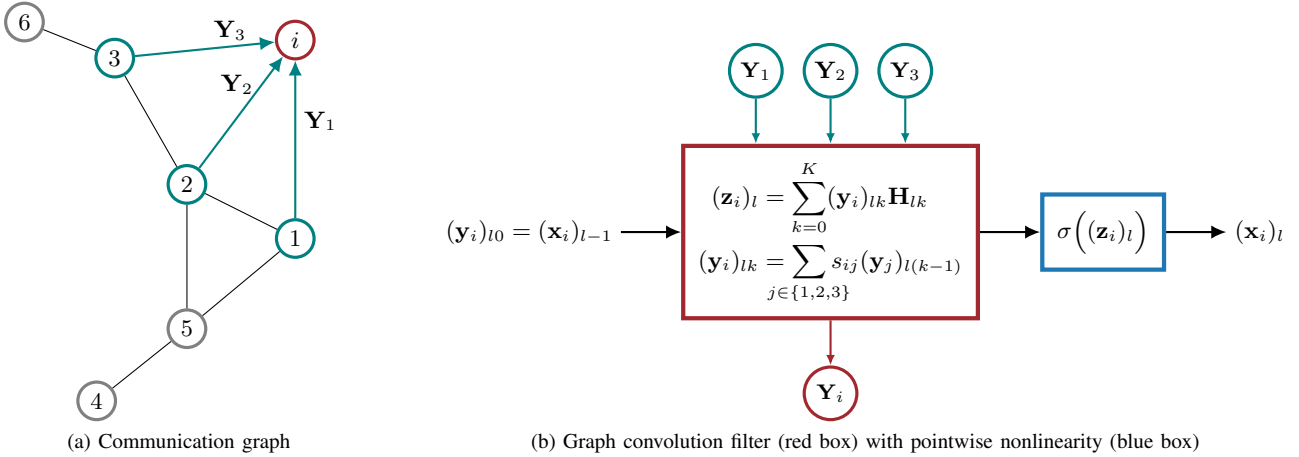

\textit{Distributed Implementation:}
The GNN architecture inherently allows a distributed implementation of the communication module.
To see this, consider the computation $\mv Y_{lk}= (\mv S)^k \mv X_{l-1}$, where $\mv X_{l-1}$ is the input to the $l$-th layer of the GNN.
The computation can be written recursively as $\mv Y_{lk} = \mv S \mv Y_{l(k-1)}$, with $\mv Y_{l0} = \mv X_{l-1}$.
For a robot $i$, the corresponding vector in $\mv Y_{lk}$ is given by:
\begin{equation}
  (\mv y_i)_{lk} = [\mv Y_{lk}]_{i} = \sum_{j\in \mc N(i)} s_{ij} (\mv y_j)_{l(k-1)},
  \label{eqn:yilk}
\end{equation}
where $\mc N(i)$ is the set of neighbors of robot $i$.
The above equation uses the fact that the shift operator $\mv S$ is a matrix representation of the communication graph, and hence, $s_{ij} = [\mv S]_{ij}$ is non-zero only if $(i, j)\in \mc E$ or equivalently, $j\in \mc N(i)$.
The equation also precisely defines the information to be exchanged between neighboring robots.
Note that the computation of $(\mv y_i)_{lk}$ requires $(\mv y_j)_{l(k-1)}$ for all $j\in \mc N(i)$.
The information sent by the robot $i$, also known as \textit{aggregated message}~\cite{TolstayaPMLR20}, is a then a collection of vectors $\mv Y_i$:
\begin{equation}
  \mv Y_i = \{(\mv y_i)_{lk}\},\quad \forall k\in\{0,\ldots,K-1\},\ l\in\{1,\ldots,L\}.
  \label{eqn:agg_msg}
\end{equation}
The output of the graph convolution filter for robot $i$ is then given by:
\begin{equation}
  (\mv z_i)_l = \sum_{k=0}^K (\mv y_i)_{lk} \mv H_{lk}.
\end{equation}
Finally, pointwise nonlinearity is applied to generate the output $(\mv x_i)_l$ of the $l$-th layer for robot $i$.
The distributed implementation of the GNN architecture is illustrated in \fgref{fig:gnn_dist}.

In the actual deployment of the architecture, the module additionally maintains buffers to send and receive information from the neighboring robots using radio communication.
In general, the communication is asynchronous, i.e., the robots may receive messages from the neighboring robots at different time steps.
Furthermore, the communication module may also run concurrently with the perception module, such that the robots may receive messages from the neighboring robots while processing the sensor data.
We refer the reader to~\cite{agarwal2023asynchronous} for a detailed discussion on asynchronous and distributed implementations of the GNN architecture.

\subsection{Action Module}
The action module is responsible for generating velocity control actions for the robot.
It uses a shallow multi-layer perceptron (MLP) with ReLU activations to process the feature vector generated by the communication module.
The MLP has two layers with 32 output channels each, and the final output of the MLP is processed by a linear layer to generate the velocity control actions for the robot.
The linear layer has two output channels corresponding to the $x$ and $y$ components of the velocity control action; it is used as the final layer of the MLP to scale the output to the range of the velocity control action.
The velocity control actions are sent to a low-level controller to generate the actual control commands for the robot actuators.

The action module may, in general, be composed of additional functionalities that modify the velocity actions before sending them to the low-level controller.
For example, depending on the type of robot, such as a ground robot or a quadrotor, the velocity commands may need to be modified to account for the dynamics of the robot.
The velocity commands may also need to be modified to ensure that the robots do not collide with each other and the boundaries of the environment.

\section{Environment and Imitation Learning} \label{sc:environment}
In this section, we discuss the clairvoyant algorithm, based on the centroidal Voronoi tessellation, used to generate the dataset for imitation learning of the LPAC architecture.
We also detail baseline algorithms that operate with limited sensing and communication capabilities used to evaluate the performance of the LPAC architecture.
Finally, we discuss the simulation environment and the dataset generation process for imitation learning.

\subsection{Centroidal Voronoi Tessellation} \label{sc:CVT}
We now discuss an iterative gradient descent algorithm that uses the centroidal Voronoi tessellation (CVT) to generate a robot configuration that provides good coverage of the environment.
The algorithm, also referred to as Lloyd's algorithm~\cite{Lloyd82,CortesMKB04}, is widely used to solve the coverage control problem.
It relies on computing the Voronoi partition (or tessellation) of the region with respect to the locations of the robots.
For $N$ robots, the Voronoi partition can be computed in $\mc O(N\log N)$ time using sweep line algorithm~\cite{deberg}, and efficient implementations~\cite{cgalvoronoi} are available.
We restate the definition of the Voronoi partition~$\mc P$ (\scref{sc:problem_statement}):
\begin{equation*}
	\begin{split}
		\mc P &= \{P_i \mid i \in \mc V\}, \quad \text{where }\\
		P_i &= \{\mv q \in \mc W \mid \|\mv p_i - \mv q\| \le \|\mv p_j - \mv q\|, \forall j \in \mc V\}.
	\end{split}
\end{equation*}

Given $N$ robots operating in a workspace $\mc W$ with a convex boundary, an observed subset of the workspace $\mc W_o\subseteq \mc W$, an importance density field (IDF)  $\Phi: \mc W \mapsto \mathbb{R}$, and a Voronoi cell $P_i$ for robot $i$, we can compute the generalized mass, centroid, and the polar moment of inertia as:
\begin{equation}
	\begin{split}
		m_i &= \int_{\mv q \in (P_i\cap \mc W_o)} \Phi(\mv q) \, d\mv q,\\
		\mv c_i &= \frac{1}{m_i} \int_{\mv q \in (P_i \cap \mc W_o)} \mv q \Phi(\mv q) \, d\mv q, \quad \text{and}\\
		I_i &= \int_{\mv q \in (P_i\cap \mc W_o)} \|\mv q - \mv c_i\|^2 \Phi(\mv q) \, d\mv q.
	\end{split}
	\label{eq:mass_centroid}
\end{equation}

The objective function~\eqref{eq:coverage_voronoi} for the coverage control problem, with $f(\|\mv p_i - \mv q\|) = \|\mv p_i - \mv q\|^2$, can be rewritten as:
\begin{equation}
	\begin{split}
		\mc J(\mc P) &= \sum_{i \in \mc V} \int_{\mv q \in (P_i\cap \mc W_o)} \|\mv p_i - \mv q\|^2 \Phi(\mv q) \, d\mv q\\
		&= \sum_{i \in \mc V} I_i + \sum_{i \in \mc V} m_i \|\mv p_i - \mv c_i\|^2.
	\end{split}
	\label{eq:voronoi_objective}
\end{equation}

Taking the partial derivative of the objective function~\eqref{eq:voronoi_objective} with respect to the location of the robot $i$, we get:
\begin{equation}
		\frac{\partial \mc J(\mc P)}{\partial \mv p_i} = 2m_i (\mv p_i - \mv c_i)
	\label{eq:voronoi_gradient}
\end{equation}
The partial derivates vanish at the centroid of the Voronoi cell, i.e., $\mv p_i = \mv c_i$; thus, the centroid of the Voronoi cell is the local minimum of the objective function~\eqref{eq:voronoi_objective}.
Hence, we can write a control law~\cite{CortesMKB04} that drives the robot towards the centroid of the Voronoi cell as:
\begin{equation}
	\mv u_i = \dot{\mv p}_i = -k (\mv p_i - \mv c_i).
	\label{eq:voronoi_control}
\end{equation}
Here, $k$ is a positive gain for the control law.
The control law in \eqref{eq:voronoi_control} has nice convergence properties; it is guaranteed to converge to a local minimum of the objective function~\cite{CortesMKB04}.

The algorithm can now be expressed as an iteration of the following steps until convergence or for a maximum number of iterations:
\begin{enumerate}
	\item Compute the Voronoi partition $\mc P$ of the region with respect to the locations of the robots.
	\item Compute the mass centroids $\mv c_i$ for each Voronoi cell $P_i$.
	\item Move each robot towards the centroid of the Voronoi cell using the control law~\eqref{eq:voronoi_control}.
\end{enumerate}

We can define different variants of the above algorithm with the same control law~\eqref{eq:voronoi_control}, depending on the observed workspace $\mc W_o$ and information available to the robots for computing the Voronoi partition and the centroids.
In this article, we refer to the algorithms as variants of CVT.

{\bf Clairvoyant:} The clairvoyant is a centralized algorithm based on CVT.
It has perfect knowledge of the positions of all the robots at all times.
Thus, it can compute the exact Voronoi partition.
It also has complete knowledge of the IDF for the centroid computation for each Voronoi cell, i.e., $\mc W_o = \mc W$.
Although the algorithm is not optimal, it generally computes solutions that are very close to the optimal solution.

{\bf Centralized CVT (C-CVT):} The C-CVT is also a centralized algorithm based on CVT.
Similar to the clairvoyant algorithm, it knows the positions of all the robots and computes the exact Voronoi partition.
However, unlike the clairvoyant algorithm, the C-CVT can access a limited IDF.
It operates on the cumulative knowledge of the IDF of all the robots, sensed up to the current time step, i.e.,
  \begin{equation*}
    \mathcal W_o=\bigcup_{i\in \mc V}\mc W_o^{(i)}\subseteq \mathcal W,
  \end{equation*}
  where $\mc W_o^{(i)}\subseteq \mc W$ is the workspace observed by robot $i$ along its entire trajectory.

Thus, the amount of information available to the C-CVT is dependent on the sensor radius of the robots and the trajectory taken by the robots.

{\bf Decentralized CVT (D-CVT):} The D-CVT is the decentralized version of the C-CVT algorithm.
Here, each robot uses the positions of neighboring robots, i.e., the robots within its communication range, to compute the Voronoi partition.
Furthermore, each robot has restricted access to the IDF sensed by itself up to the current time, i.e., $\mc W_o = \mc W_{o}^{(i)}$.
Thus, the D-CVT algorithm uses only the local information available to each robot to take control actions.

As one can expect from the amount of information available to each algorithm, the clairvoyant algorithm is the best-performing algorithm, followed by the C-CVT and the D-CVT algorithms.
The clairvoyant algorithm is used to generate the dataset for training, and the C-CVT and D-CVT baseline algorithms are used to evaluate the performance of the LPAC architecture.

\subsection{Coverage Control Environment} \label{sc:dataset_generation}
The environment is a 2D grid world of size \sqrdim{1024} cells with a resolution of \SImSqrDim{1} per cell.
The environment is populated with $M$ features of interest that are randomly placed in the environment.
The IDF is generated by defining a 2-D Gaussian distribution as a probability density function centered at the feature of interest, with a randomly generated standard deviation $\sigma\in[40, 60]$ and a randomly generated scale factor $\alpha\in[6, 10]$.
The Gaussian distribution is set to 0 beyond a distance of $2\sigma$ from the center of the feature of interest to avoid any detection of features from a faraway location.
The value of a cell in the IDF is set to the sum of the integrals of the Gaussian distributions over the area of the cell for all features of interest.
Finally, the values are normalized to have a maximum value of one.

The environment is populated with $N$ robots that are randomly placed in the environment according to a uniform probability distribution.
The robots have a square sensor field of view with a side length of \SIm{64}.
The sensor field of view is centered at the location of the robot.
A disk-shaped sensor field-of-view can be approximated by a square sensor field-of-view with a side length of the diameter of the disk and setting the values outside the disk to zero.
The robots have a communication range of \SIm{128}, i.e., they are able to communicate with each other if within this range.
The maximum speed of the robots is set to \SIvel{5}.

\subsection{Data Generation and Imitation Learning}
  The learning pipeline comprises data generation, imitation learning, and evaluation.
  Imitation learning is used to train the LPAC architecture to mimic the behavior of the clairvoyant algorithm, which serves two main purposes:
  (i)~drive the simulation by taking actions during data generation, and
  (ii)~provide near-optimal velocity actions for the LPAC architecture to learn from.
  Although the clairvoyant algorithm has knowledge of the entire IDF, the input to the LPAC architecture for both training and evaluation is always limited to the observed IDF with the limited sensing capabilities of the robots.
  Each robot~$i$ independently runs the same LPAC policy using only the observed workspace $\mc W_o^{(i)}$ made by the robot~$i$ and GNN-based abstract information communicated by the neighboring robots.
  There are two primary advantages to this approach:
  Firstly, designing highly specialized algorithms with limited sensing capabilities is challenging, which is circumvented by using the clairvoyant algorithm.
  Second, as the results show in \scref{sc:results}, the LPAC policy is able to outperform the centralized algorithm (C-CVT) with limited observations, which would not have been possible if the LPAC architecture had been trained using the C-CVT algorithm.

Using the clairvoyant algorithm, the simulation environment evolves in discrete time steps of \SIs{0.2}, i.e., with a frequency of \SIHz{5}, and the maximum distance that the robots are able to move is \SIm{1} in each time step.
The algorithm is run until convergence or for a maximum of $1000$ iterations.
At each iteration, for each robot, the four-channel input to the CNN of the perception module (\scref{sc:perception}), along with the control actions and positions of robots, are stored as a state-action pair.
Such state-action pairs are stored every five iterations of the algorithm.
We additionally generate configurations at the converged state of the CVT algorithm to help the imitation learning algorithm learn the converged state.
In total, 100,000 data points are generated, each containing state-action pairs for 32 robots.

The training is performed using the Adam optimizer~\cite{KingBa15} in Python using PyTorch~\cite{PyTorch} and PyTorch Geometric~\cite{PyTorchGeometric}.
We use a batch size of 750 and train the network for 100 epochs.
The learning rate is set to \num{e-4}, and the weight decay is set to \num{e-3}.
The mean squared error (MSE) loss is used as the loss function, where the target is the output of the clairvoyant algorithm, and the prediction is the output of the LPAC architecture.
The training and validation loss curves are shown in \fgref{fig:loss}.
The model with the lowest validation loss is selected as the final model.

\begin{figure}[ht]
  \centering
	\begin{tikzpicture}
	\small
	\footnotesize
  \begin{axis}[
    width=0.95\columnwidth,
    height=0.6\columnwidth,
    xlabel={Epoch},
    ylabel={MSE Loss},
		xlabel style={font=\small},
		ylabel style={font=\small},
		legend style={font=\scriptsize, at={(0.50,1.15)},anchor=north},
		legend columns=-1,
    grid=major,
    grid style={gray!30}
    ]

    \addplot+[mark=none,mBlue,thick]
    table[x=epoch,y=train_loss,col sep=space,header=true] {figures/data/train_loss.dat};
    \addlegendentry{Training Loss\;\;}

    \addplot+[mark=none,mDarkRed,thick,dashed]
    table[x=epoch,y=validation_loss,col sep=space,header=true] {figures/data/val_loss.dat};
    \addlegendentry{Validation Loss}

    \pgfmathsetmacro{\minvalx}{16}
    \pgfmathsetmacro{\minvaly}{0.3239021375775337}

    \draw[fill=mDarkRed] (axis cs:\minvalx,\minvaly) circle[radius=2pt];
  \end{axis}
\end{tikzpicture}
  \caption{Evolution of the training and validation Mean Squared Error (MSE) loss over epochs for the primary LPAC-K3 model trained on \sqrdim{1024} grid world with 32 robots.
  The training loss steadily decreases, indicating the model is fitting the data more closely, while the validation loss remains higher and fluctuates, indicating the model is not overfitting.
The highlighted point marks the epoch (\#16) at which the validation loss is minimized, which is used to select the final model.\label{fig:loss}}
\end{figure}

\section{Results} \label{sc:results}
This section presents empirical evaluations of the proposed LPAC architecture for the coverage control problem.
Our experiments establish the following:
\begin{enumerate}
  \item The model \textbf{outperforms the baseline} centralized and decentralized CVT algorithms. It learns to share relevant abstract information, which is generally challenging to design for decentralized algorithms.
  \item The \textbf{ablation study} shows: (a)~The model with $K=3$ in the GNN architecture performs better than the ones with $K=1$ and $K=2$.
    Thus, the architecture is able to exploit the propagation of information over a larger portion of the communication graph.
    (b)~The neighbor maps and the normalized robot positions are essential for improving the performance of the model.
  \item The model \textbf{generalizes} to environments with varying numbers of robots and features, especially when they are larger than the training distribution.
  \item The model \textbf{transfers} well to larger environments with a larger number of robots while keeping the ratio of robots to environment size the same.
  \item The model is \textbf{robust to noisy position estimates}, indicating that the model can be deployed on real-world robots with noisy sensors.
  \item Models trained with different communication ranges perform well for up to half the environment size, and the performance is better than the CVT algorithms for all ranges.
  \item Even though the model is trained on synthetic datasets with randomly generated features, it performs well on \textbf{real-world datasets} without further training.
\end{enumerate}

We implemented an open-source platform\footnote{\url{https://github.com/KumarRobotics/CoverageControl}} for simulating the coverage control problem with the LPAC architecture.
The platform is implemented in \texttt{C++}, \texttt{CUDA}, and \texttt{Python} using the \texttt{PyTorch} library.
The \texttt{Python} interface is used for training and rapid prototyping, while the evaluation and deployment are done using the \texttt{C++} interface.

\begin{figure*}[htbp]
  \centering
  \subfloat{\begin{tikzpicture}\node [draw=none, rotate=90, minimum width=0.24\textwidth, inner sep=0cm] {\footnotesize D-CVT};\end{tikzpicture}}
  \subfloat{\includegraphics[width=0.24\textwidth]{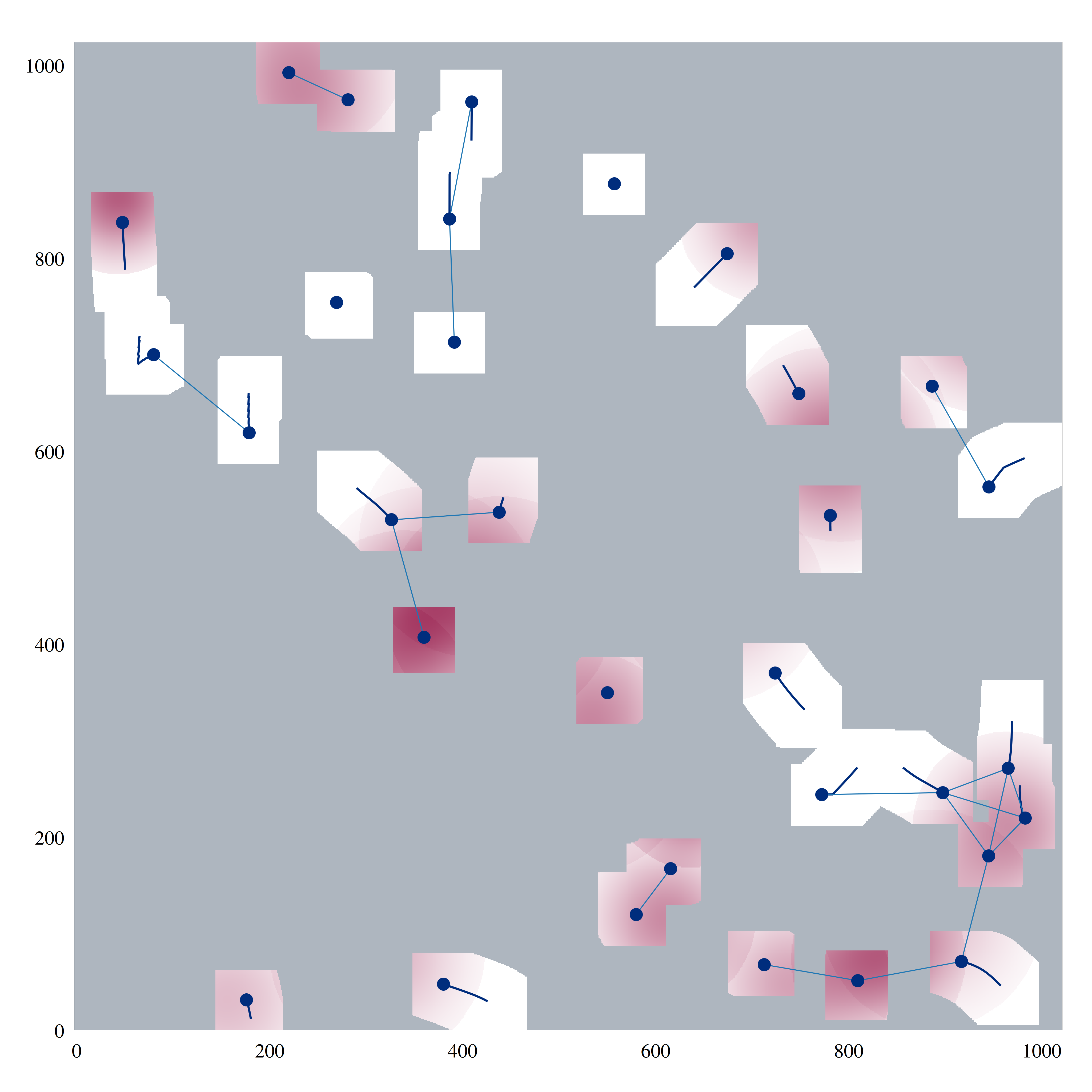}}
  \subfloat{\includegraphics[width=0.24\textwidth]{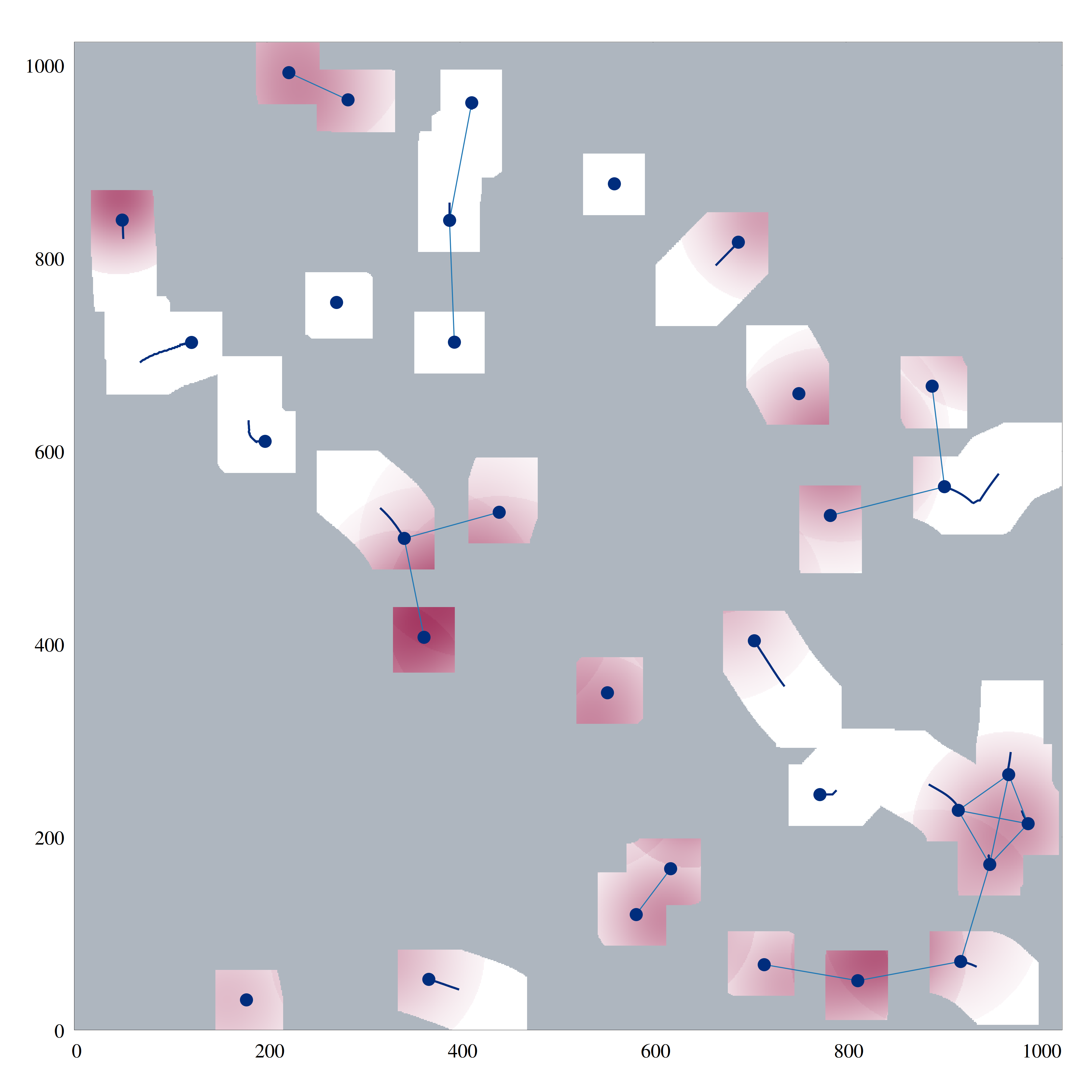}}
  \subfloat{\includegraphics[width=0.24\textwidth]{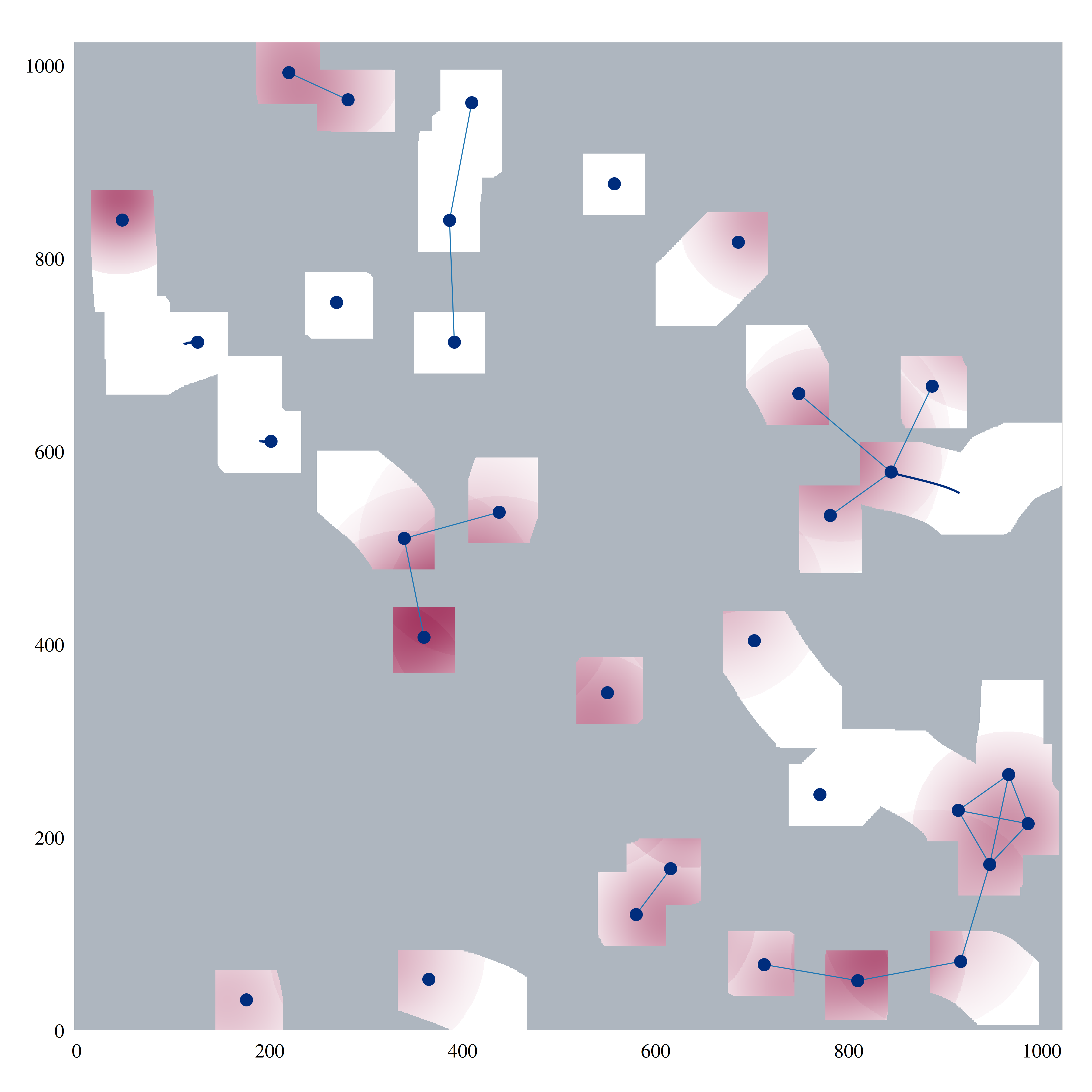}}
  \subfloat{\includegraphics[width=0.24\textwidth]{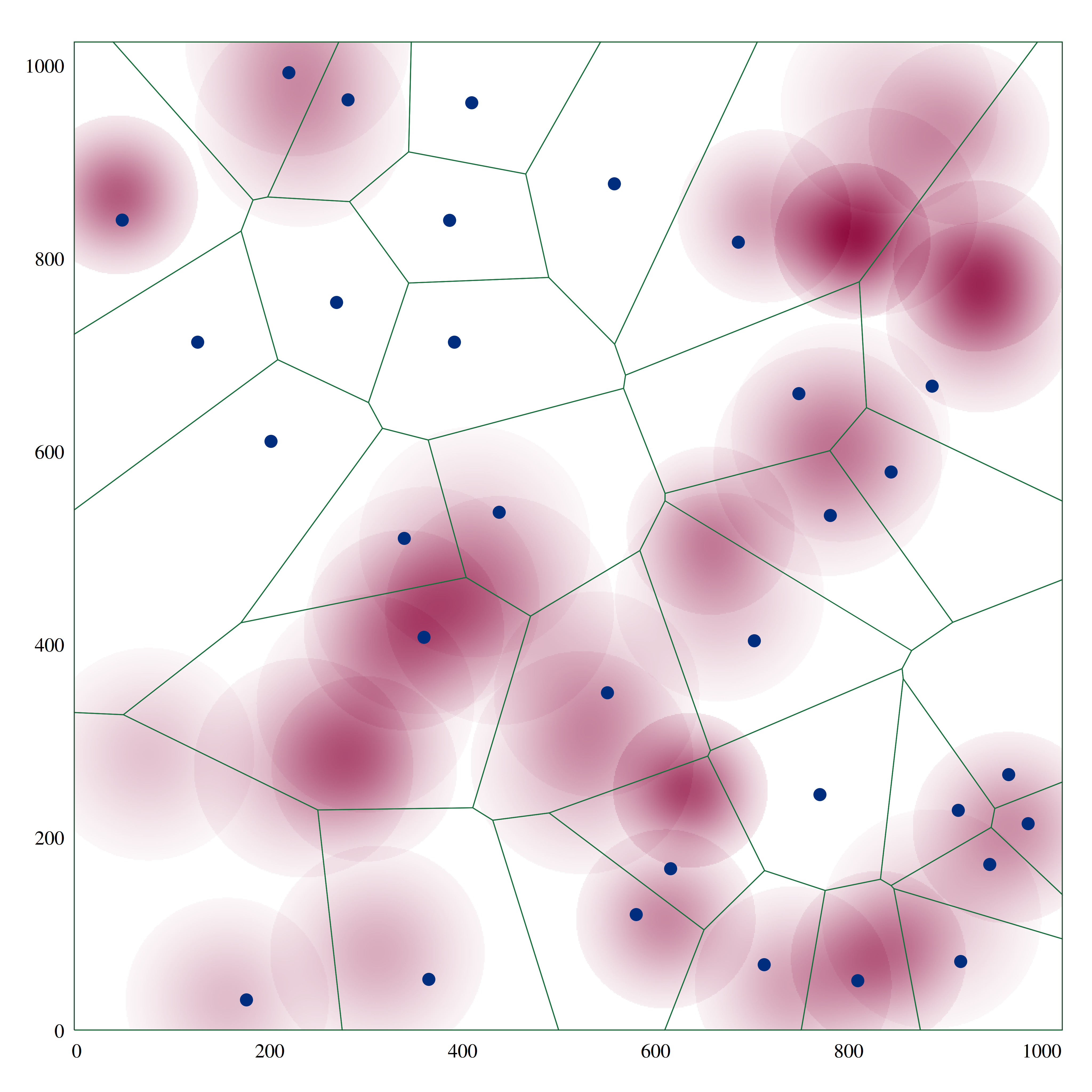}}\\
  \subfloat{\begin{tikzpicture}\node [draw=none, rotate=90, minimum width=0.24\textwidth, inner sep=0cm] {\footnotesize C-CVT};\end{tikzpicture}}
  \subfloat{\includegraphics[width=0.24\textwidth]{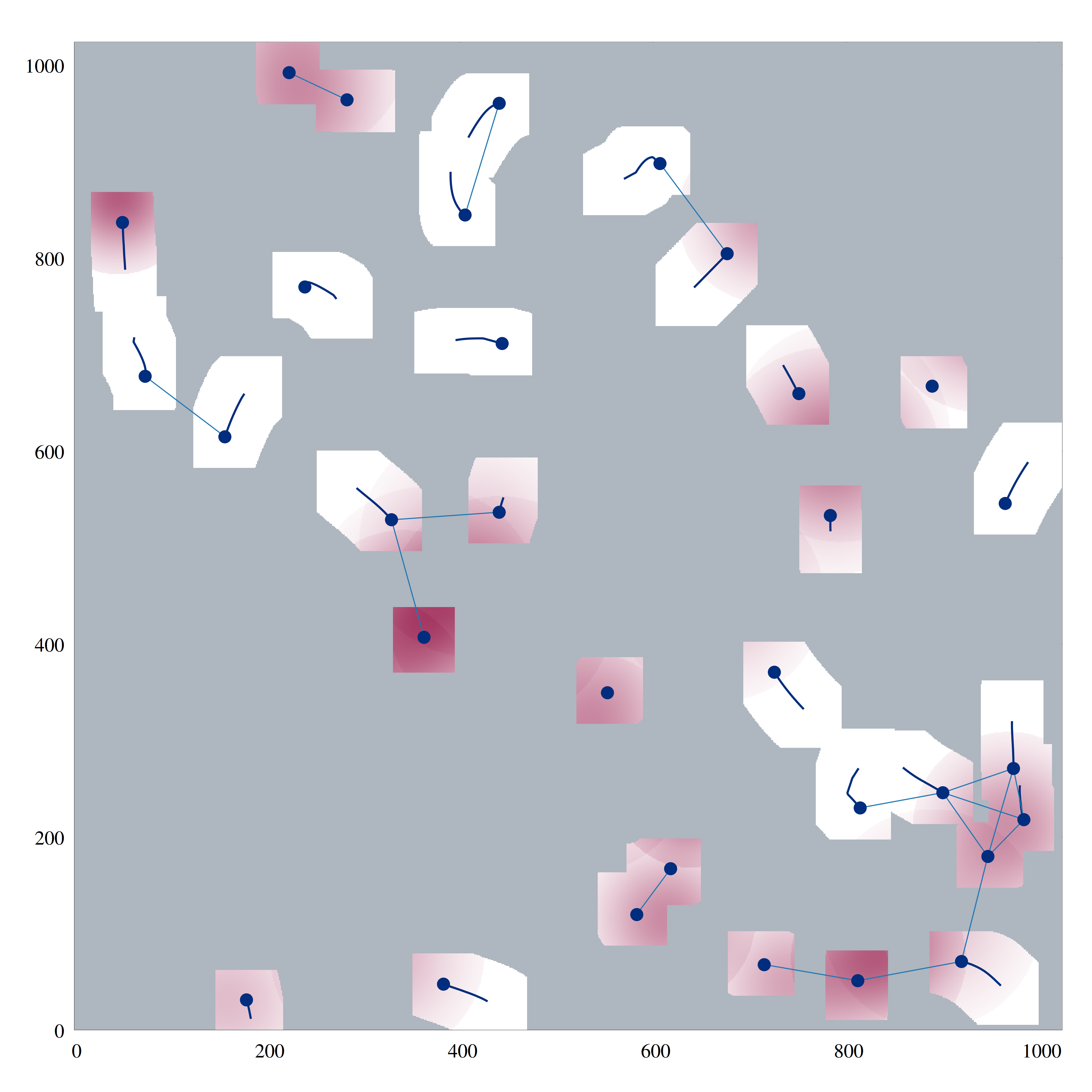}}
  \subfloat{\includegraphics[width=0.24\textwidth]{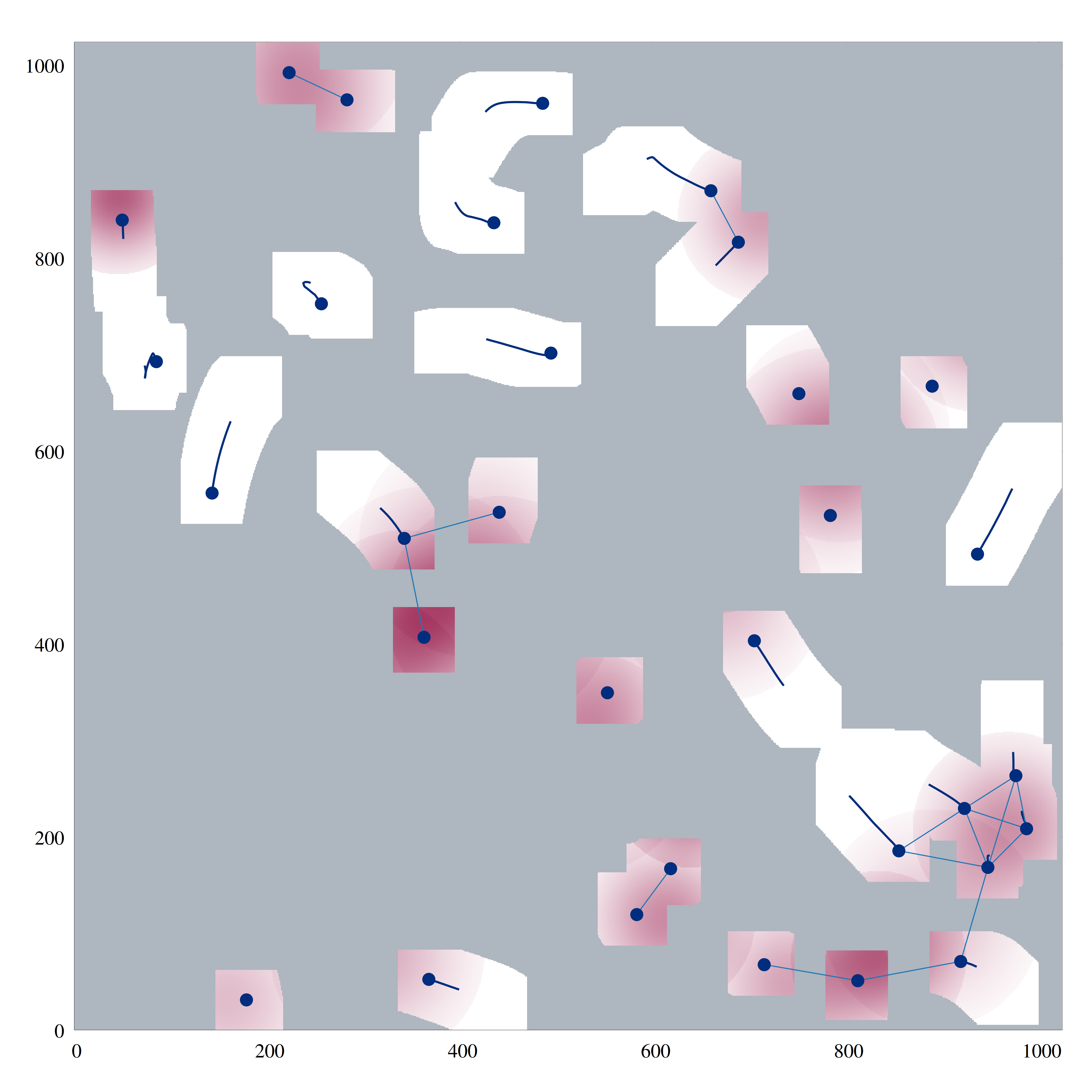}}
  \subfloat{\includegraphics[width=0.24\textwidth]{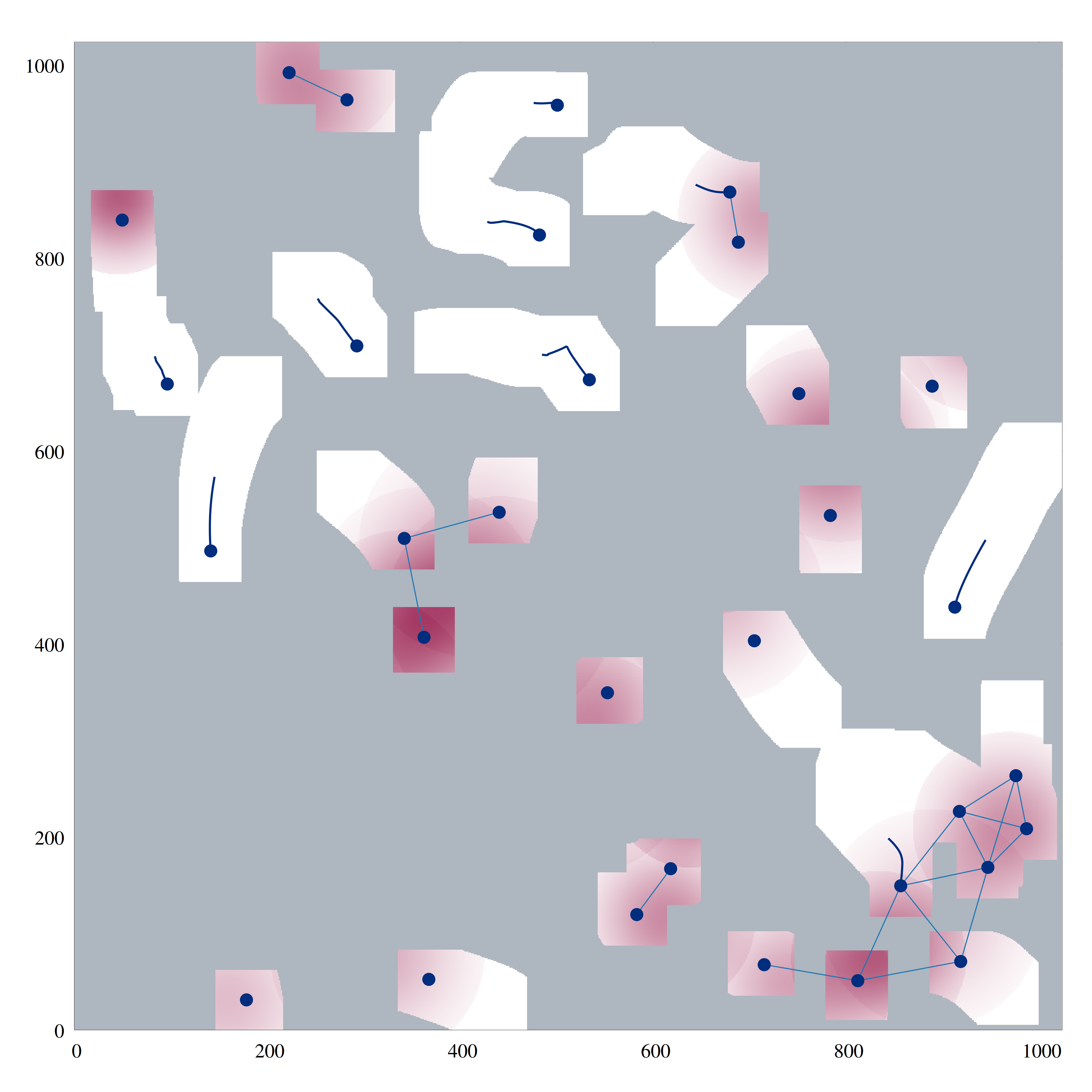}}
  \subfloat{\includegraphics[width=0.24\textwidth]{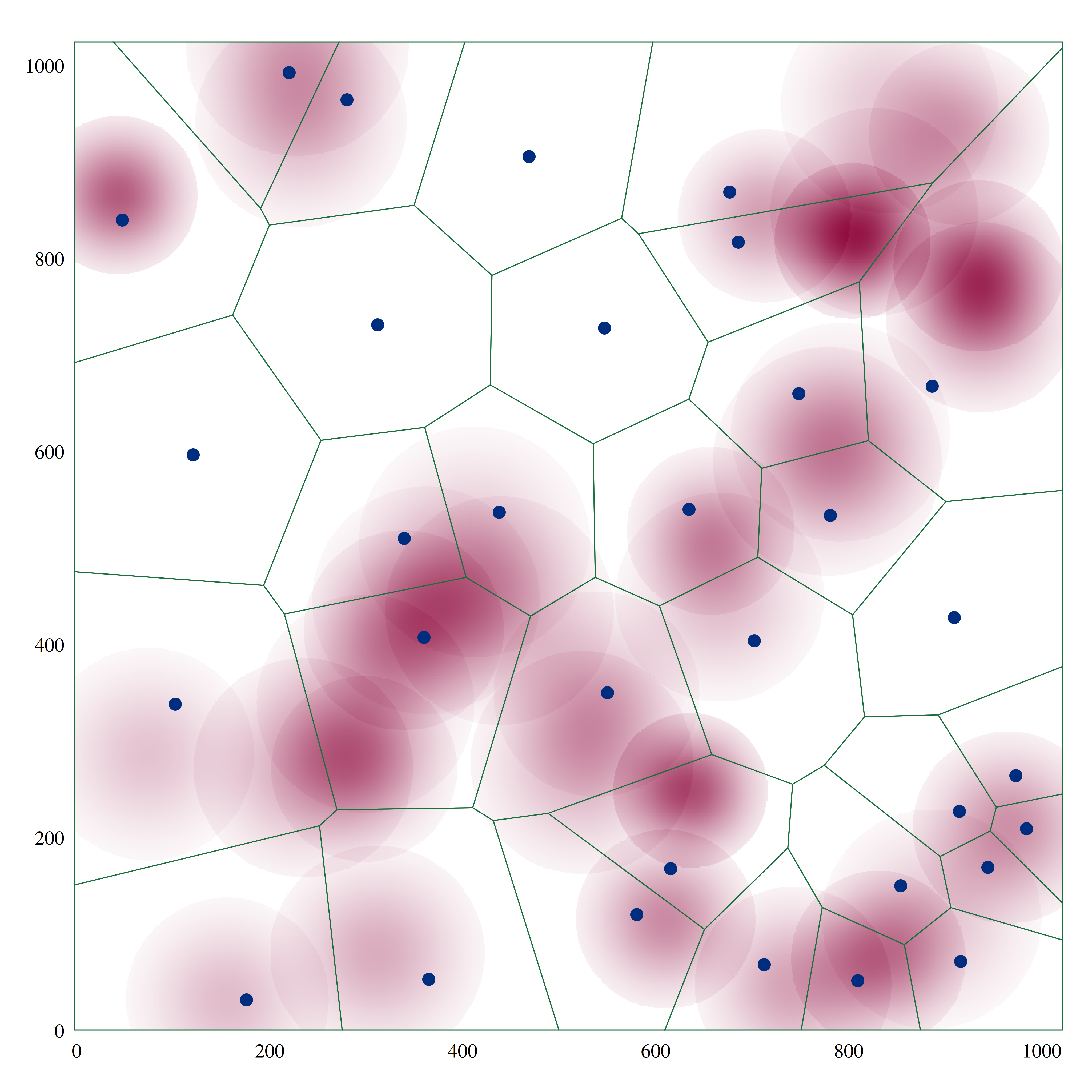}}\\
  \subfloat{\begin{tikzpicture}\node [draw=none, rotate=90, minimum width=0.24\textwidth, inner sep=0cm] {\footnotesize LPAC-K3};\end{tikzpicture}}
  \setcounter{subfigure}{0}
  \subfloat[Time Step = 60]{\includegraphics[width=0.24\textwidth]{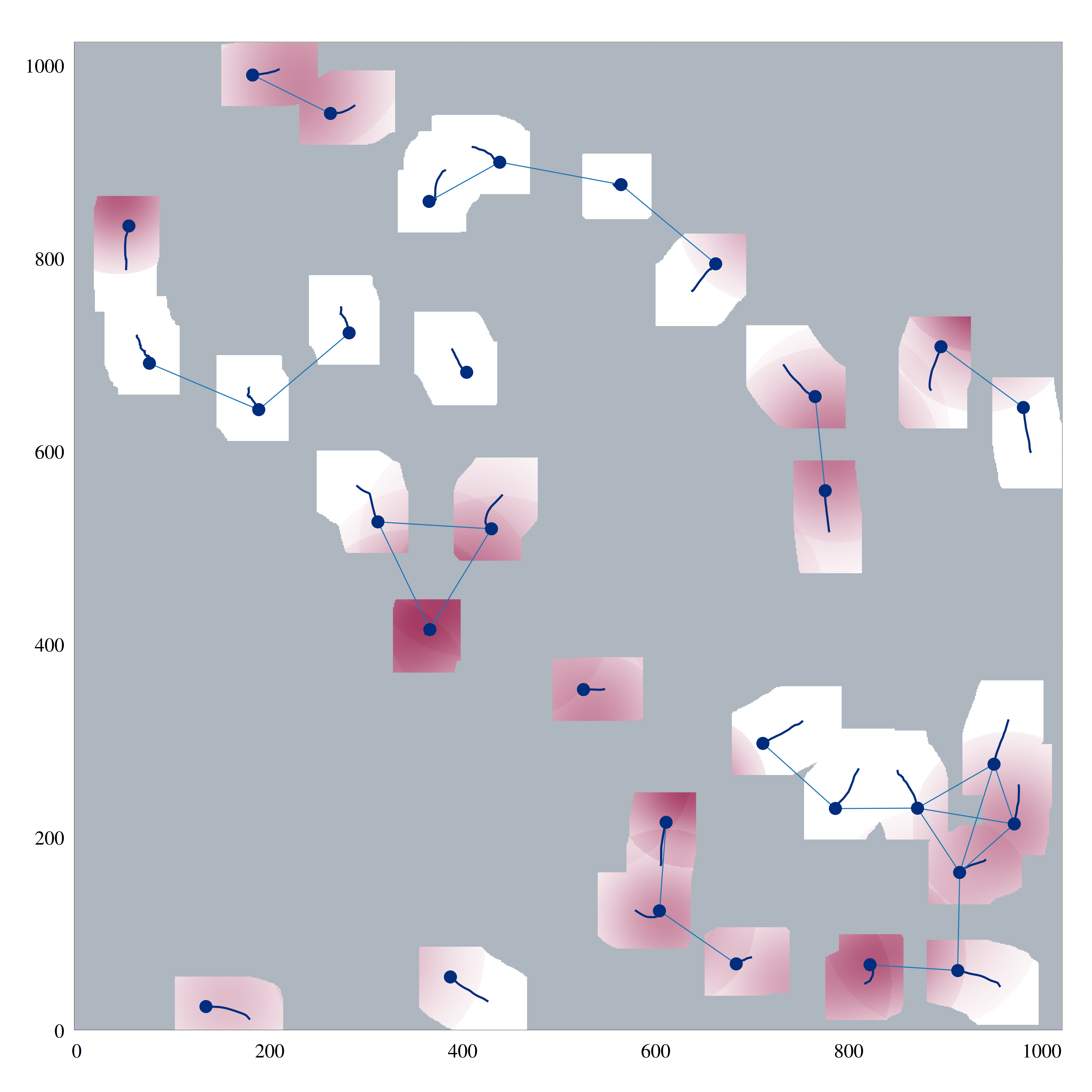}}
  \subfloat[Time Step = 120]{\includegraphics[width=0.24\textwidth]{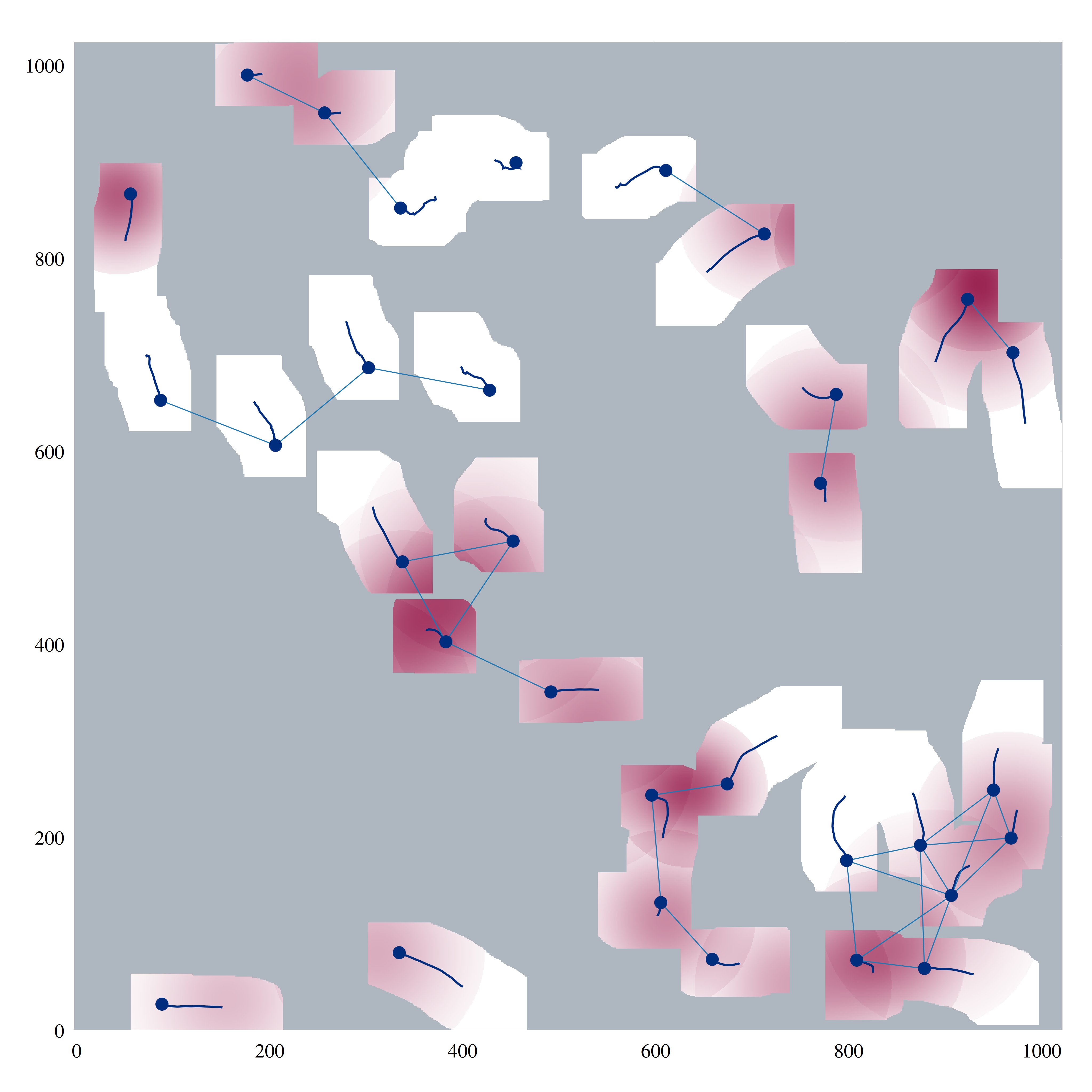}}
  \subfloat[Time Step = 180]{\includegraphics[width=0.24\textwidth]{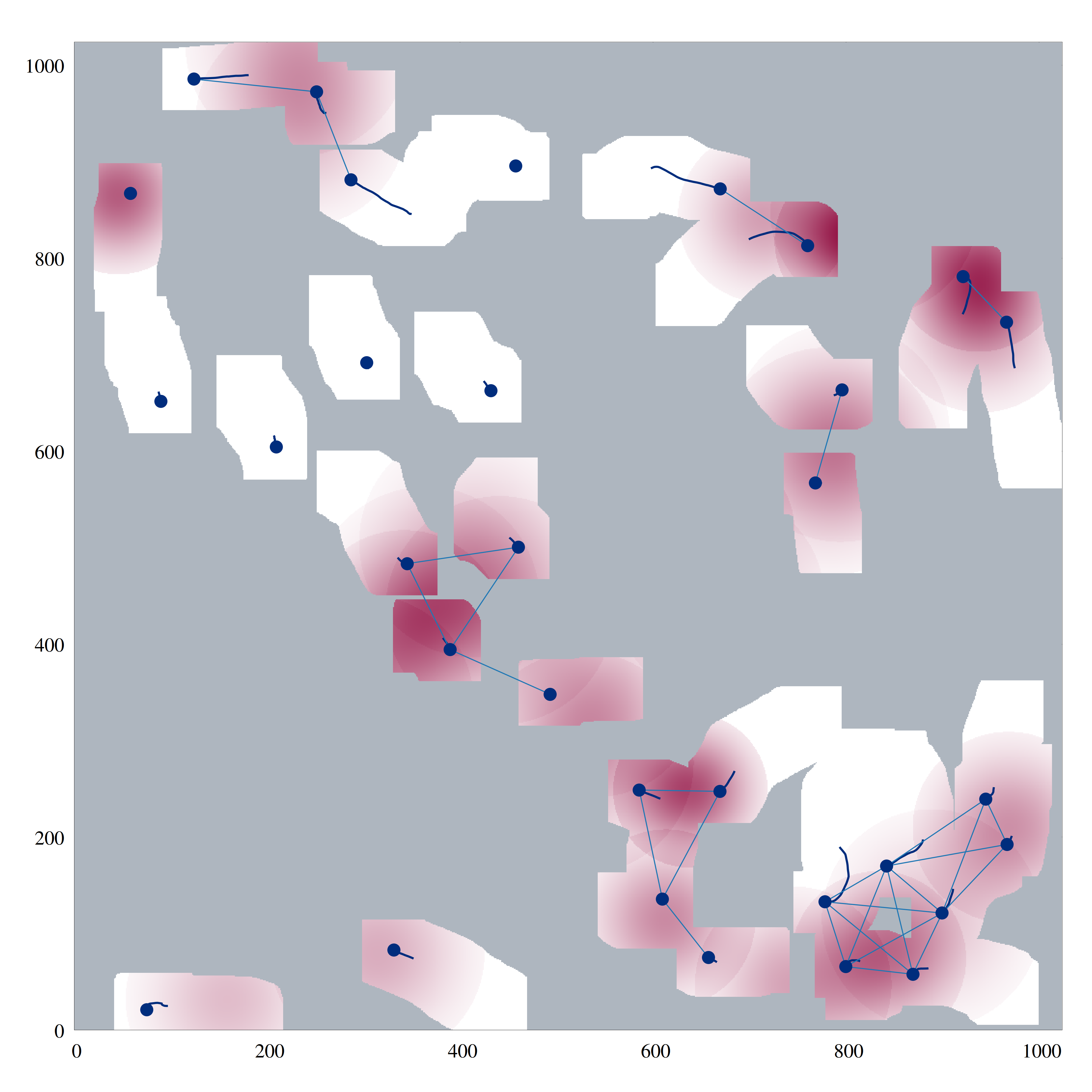}}
  \subfloat[Time Step = 900]{\includegraphics[width=0.24\textwidth]{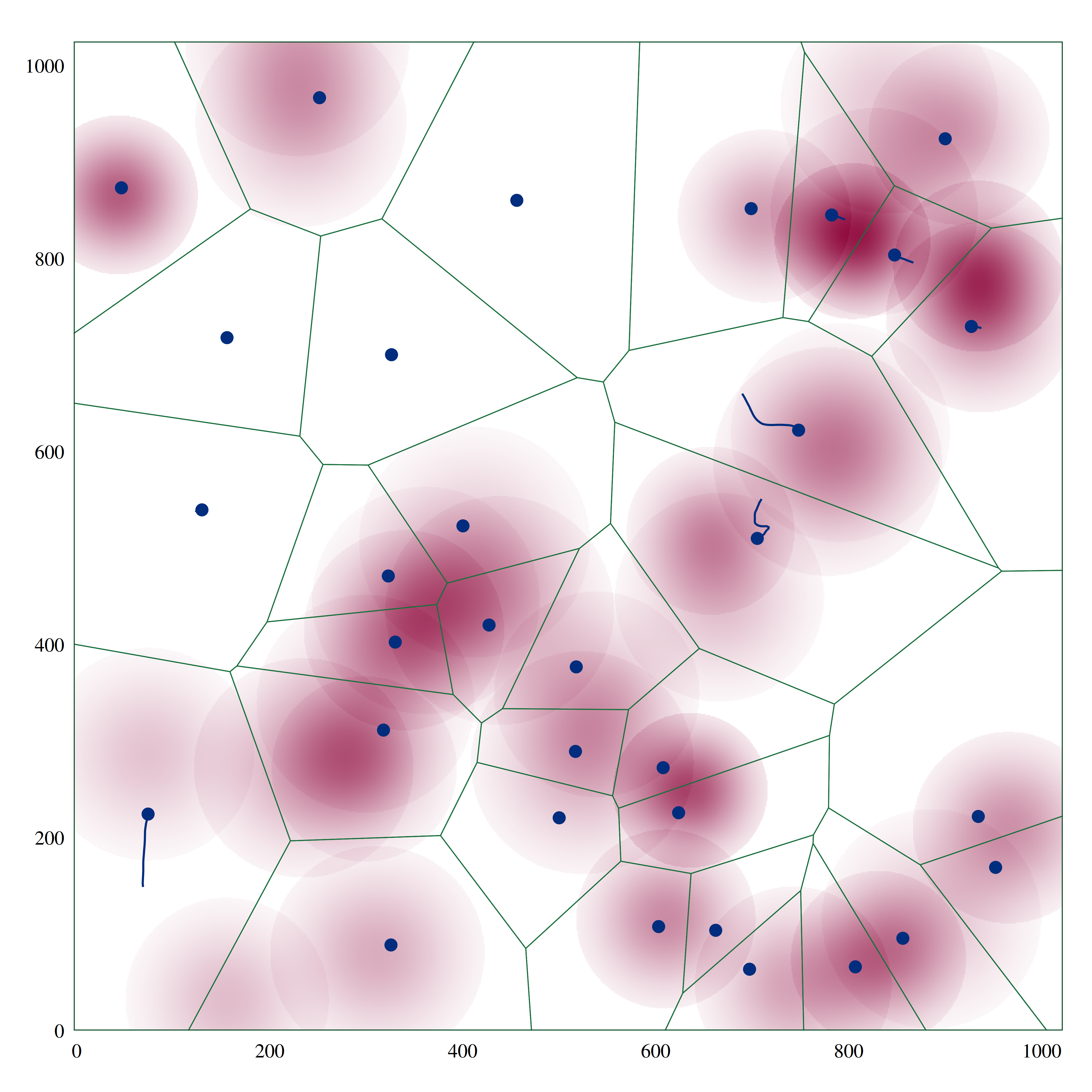}}
  \caption{Progression of the decentralized and centralized CVT algorithms (D-CVT and C-CVT) and the LPAC-K3 model for a sample environment.
    The features and the initial positions of the robots are sampled uniformly at random.
    The first three columns show the cumulative observations of all robots up to time steps 60, 120, and 180, respectively.
    The robots are shown as blue discs with blue lines showing the trajectory for the past 40 time steps.
    The light blue lines show communication links between robots based on a range of \SIm{128}.
    The last column shows the positions of the robots at time step 900 with the Voronoi cells of the robots.
    The entire IDF is shown in the final time step.
  \label{fig:timeline}}
\end{figure*}

\subsection{Comparison to Baseline Algorithms} \label{sc:baseline-comparison}
We evaluate our learned LPAC models against the decentralized and centralized centroidal Voronoi tessellation (CVT) algorithms, as discussed in \scref{sc:CVT}.
We also compare the models against the clairvoyant algorithm, which has knowledge of the entire IDF and the positions of all robots, and it can compute near-optimal actions for each robot.

We evaluated the performance of the LPAC models and the baseline algorithms on 100 environments, each of size \SImSqrDim{1024} with 32 robots.
The initial positions of the robots are sampled uniformly at random in each environment.
The IDF in each environment is generated by 32 Gaussian distributions at random locations in the environment.
\fgref{fig:timeline} shows the progression of the controllers for various time steps in a random environment.
Since the CVT-based algorithms converge very fast, we show the state of the system for the beginning of the episode, i.e., time steps \numlist{60;120;180}.
For this environment, the LPAC-K3 controller performs better than decentralized CVT at all timesteps and better than centralized CVT after 12 time steps.

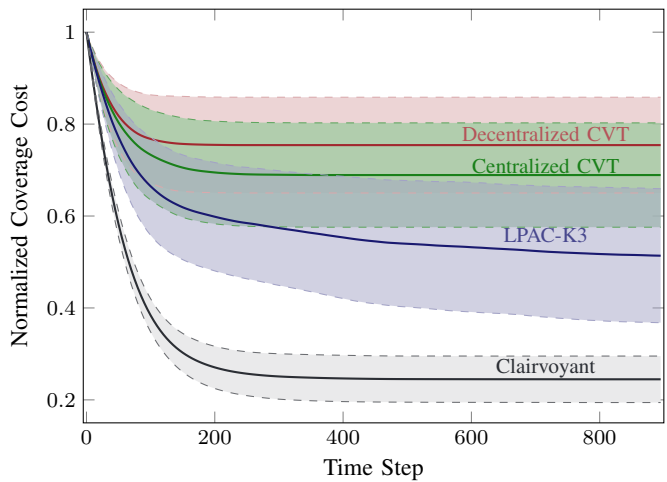
\begin{figure}[ht]
  \begin{tikzpicture}
	\footnotesize
	\tikzset{
		mylabel/.style={pos=0.80, above, yshift=-2.5pt}
	}
	\pgfplotstableread[col sep=comma]{./figures/data/normalized_cost_time_1024.csv}{\costdata}
	\begin{axis}[
		width=1.05\columnwidth,
		height=0.8\columnwidth,
		xlabel={Time Step},
		ylabel={Normalized Coverage Cost},
		xlabel style={font=\small},
		ylabel style={font=\small},
		legend style={font=\scriptsize, at={(0.50,1.15)},anchor=north},
		legend columns=-1,
		xmin=-5,
		xmax=900,
		ymin=0.15,
		ymax=1.05,
		]
		\def\controller{LloydLocalVoronoi}
		\addplot[forget plot, mark=none, mark size=5pt,draw=mDarkRed, thick]  table[x=step, y=\controller, col sep=comma] {\costdata} node[mylabel, yshift=+1.0pt] {\textcolor{mDarkRed!80}{\footnotesize Decentralized CVT}};
		\addplot[forget plot, name path=upper,draw=mDarkRed!40, thin, dashed] table[x=step, y expr=\thisrow{\controller}+\thisrow{\controller_std}, col sep=comma] {\costdata};
		\addplot[forget plot, name path=lower,draw=mDarkRed!40, thin, dashed] table[x=step,y expr=\thisrow{\controller}-\thisrow{\controller_std}, col sep=comma] {\costdata};
		\addplot[fill=mDarkRed!40, fill opacity=0.5] fill between[of=upper and lower];

		\def\controller{LloydLocalSensorGlobalComm}
		\addplot[forget plot, mark=none, mark size=5pt,draw=mGreen, thick]  table[x=step, y=\controller, col sep=comma] {\costdata} node[mylabel] {\textcolor{mGreen}{\footnotesize Centralized CVT}};
		\addplot[forget plot, name path=upper,draw=mGreen!70, thin, dashed] table[x=step, y expr=\thisrow{\controller}+\thisrow{\controller_std}, col sep=comma] {\costdata};
		\addplot[forget plot, name path=lower,draw=mGreen!70, thin, dashed] table[x=step,y expr=\thisrow{\controller}-\thisrow{\controller_std}, col sep=comma] {\costdata};
		\addplot[fill=mGreen!40, fill opacity=0.8] fill between[of=upper and lower];

		\def\controller{CNNGNN_k3}
		\addplot[forget plot, mark=none, mark size=5pt,draw=mDarkBlue, thick]  table[x=step, y=\controller, col sep=comma] {\costdata} node[mylabel, yshift=+3.0pt] {\textcolor{mDarkBlue!80}{\footnotesize LPAC-K3}};
		\addplot[forget plot, name path=upper,draw=mDarkBlue!40, dashed, thin] table[x=step, y expr=\thisrow{\controller}+\thisrow{\controller_std}, col sep=comma] {\costdata};
		\addplot[forget plot, name path=lower,draw=mDarkBlue!40, dashed, thin] table[x=step,y expr=\thisrow{\controller}-\thisrow{\controller_std}, col sep=comma] {\costdata};
		\addplot[fill=mDarkBlue!40, fill opacity=0.5] fill between[of=upper and lower];

		\def\controller{LloydGlobalOnline}
		\addplot[forget plot, mark=none, mark size=5pt,draw=mSteelGray, thick]  table[x=step, y=\controller, col sep=comma] {\costdata} node[mylabel] {\textcolor{mSteelGray}{\footnotesize Clairvoyant}};
		\addplot[forget plot, name path=upper,draw=mSteelGray!70, dashed, thin] table[x=step, y expr=\thisrow{\controller}+\thisrow{\controller_std}, col sep=comma] {\costdata};
		\addplot[forget plot, name path=lower,draw=mSteelGray!70, dashed, thin] table[x=step,y expr=\thisrow{\controller}-\thisrow{\controller_std}, col sep=comma] {\costdata};
		\addplot[fill=mSteelGray!10] fill between[of=upper and lower];
	\end{axis}
\end{tikzpicture}
  \caption{Evaluation of the LPAC model with respect to CVT-based algorithms:
    The coverage cost is normalized by the cost at the initialization of the environment. Hence, the cost is 1 at time step 0.
    The controllers are run for 100 environments for 900 time steps each.
    For each controller, the mean performance over all environments is shown as a solid line, and the standard deviation is shown as a shaded region.
  The LPAC model outperforms both the centralized and the decentralized CVT algorithms.\label{fig:cost_std_time}}
\end{figure}
\begin{figure}[ht]
  \begin{tikzpicture}
	\footnotesize
	\tikzset{
		mylabel/.style={pos=0.50, above, yshift=-2.5pt}
	}
  \pgfplotstableread[col sep=comma]{./figures/data/area_coverage_1024.csv}{\costdata}
	\begin{axis}[
		width=1.05\columnwidth,
		height=0.8\columnwidth,
		xlabel={Time Step},
    ylabel={Observed Area (\% of the environment)},
		xlabel style={font=\small},
		ylabel style={font=\small},
		legend style={font=\scriptsize, at={(0.50,1.15)},anchor=north},
		legend columns=-1,
		xmin=-5,
		xmax=900,
		]
		\def\controller{LloydLocalVoronoi}
		\addplot[forget plot, mark=none, mark size=5pt,draw=mDarkRed, thick]  table[x=step, y=\controller, col sep=comma] {\costdata} node[mylabel, yshift=+0.1pt] {\textcolor{mDarkRed!80}{\footnotesize Decentralized CVT}};
		\addplot[forget plot, name path=upper,draw=mDarkRed!40, thin, dashed] table[x=step, y expr=\thisrow{\controller}+\thisrow{\controller_std}, col sep=comma] {\costdata};
		\addplot[forget plot, name path=lower,draw=mDarkRed!40, thin, dashed] table[x=step,y expr=\thisrow{\controller}-\thisrow{\controller_std}, col sep=comma] {\costdata};
		\addplot[fill=mDarkRed!40, fill opacity=0.5] fill between[of=upper and lower];
		\def\controller{LloydLocalSensorGlobalComm}
		\addplot[forget plot, mark=none, mark size=5pt,draw=mGreen, thick]  table[x=step, y=\controller, col sep=comma] {\costdata} node[mylabel] {\textcolor{mGreen}{\footnotesize Centralized CVT}};
		\addplot[forget plot, name path=upper,draw=mGreen!70, thin, dashed] table[x=step, y expr=\thisrow{\controller}+\thisrow{\controller_std}, col sep=comma] {\costdata};
		\addplot[forget plot, name path=lower,draw=mGreen!70, thin, dashed] table[x=step,y expr=\thisrow{\controller}-\thisrow{\controller_std}, col sep=comma] {\costdata};
		\addplot[fill=mGreen!40, fill opacity=0.8] fill between[of=upper and lower];
		\def\controller{CNNGNN_k3}
		\addplot[forget plot, mark=none, mark size=5pt,draw=mDarkBlue, thick]  table[x=step, y=\controller, col sep=comma] {\costdata} node[mylabel, yshift=+3.0pt] {\textcolor{mDarkBlue!80}{\footnotesize LPAC-K3}};
		\addplot[forget plot, name path=upper,draw=mDarkBlue!40, dashed, thin] table[x=step, y expr=\thisrow{\controller}+\thisrow{\controller_std}, col sep=comma] {\costdata};
		\addplot[forget plot, name path=lower,draw=mDarkBlue!40, dashed, thin] table[x=step,y expr=\thisrow{\controller}-\thisrow{\controller_std}, col sep=comma] {\costdata};
		\addplot[fill=mDarkBlue!40, fill opacity=0.5] fill between[of=upper and lower];
		\def\controller{LloydGlobalOnline}
		\addplot[forget plot, mark=none, mark size=5pt,draw=mSteelGray, thick]  table[x=step, y=\controller, col sep=comma] {\costdata} node[mylabel] {\textcolor{mSteelGray}{\footnotesize Clairvoyant}};
		\addplot[forget plot, name path=upper,draw=mSteelGray!70, dashed, thin] table[x=step, y expr=\thisrow{\controller}+\thisrow{\controller_std}, col sep=comma] {\costdata};
		\addplot[forget plot, name path=lower,draw=mSteelGray!70, dashed, thin] table[x=step,y expr=\thisrow{\controller}-\thisrow{\controller_std}, col sep=comma] {\costdata};
		\addplot[fill=mSteelGray!10] fill between[of=upper and lower];
	\end{axis}
\end{tikzpicture}
  \caption{Percentage of the environment observed over time: The decentralized and centralized CVT algorithms observe a small portion of the environment as they quickly converge to local minima based on local observations.
    The clairvoyant algorithm, leveraging complete knowledge of the importance density function (IDF), spreads the robots effectively across the environment, resulting in a higher observed area.
  The LPAC-K3 policy, though initially observing less than the clairvoyant algorithm, gradually increases the observed area over time by distributing the robots more effectively, ultimately achieving a higher observed area compared to the CVT algorithms.\label{fig:areacoverage}}
\end{figure}
\fgref{fig:cost_std_time} shows the performance of the LPAC models and the baseline algorithms over time for 100 environments.
The features used for the IDF and the initial positions of the robots are uniformly sampled at random.
For performance evaluation, we measure the coverage cost relative to the initial state of the environment, i.e., we normalize the coverage cost at each timestep by the initial coverage cost.
This provides a consistent baseline that highlights how effectively each method reduces the coverage cost from its starting conditions, thereby reducing the impact of the initial conditions on the comparison.

The solid lines in the figure show the normalized coverage cost averaged over all environments, and the shaded regions show the standard deviation of the coverage cost.
The average performance of the LPAC model, with $K=3$ hops in the GNN architecture, is significantly better than both the decentralized and centralized CVT algorithms.
The standard deviation of the LPAC model is slightly higher than the CVT algorithms, but the spread of the standard deviation is mostly below that of the decentralized CVT algorithm.

\fgref{fig:areacoverage} shows the percentage of the environment observed by the robots over time.
There is a correlation between the observed area and the coverage cost, shown in \fgref{fig:cost_std_time}.
The decentralized and centralized CVT algorithms observe only a small portion of the environment as they quickly converge to local minima based on their immediate local observations.
In contrast, the clairvoyant algorithm, which has complete knowledge of the IDF, strategically spreads the robots across the environment to reach near-optimal locations, resulting in a higher observed area.
Note that the observed area for the clairvoyant algorithm is calculated based on what the robots would have observed during their motion, as the algorithm itself has complete knowledge of the IDF.
Initially, the observed area of the LPAC model is lower than that of the clairvoyant algorithm, but it increases over time since the LPAC model does not have an explicit convergence criterion.
Similar to the clairvoyant algorithm, the LPAC-K3 policy effectively distributes the robots and explores more of the environment, achieving a greater observed area compared to the CVT algorithms.

\begin{figure}[ht]
  \begin{tikzpicture}
	\small
	\tikzset{
		mylabel/.style={pos=0.85, above, yshift=-2.0pt}
	}
	\footnotesize
	\pgfplotstableread[col sep=comma]{./figures/data/stats_best_performance_1024.csv}{\costdata}
	\begin{axis}[
		width=1.0\columnwidth,
		height=0.7\columnwidth,
		xlabel={Time Step},
		ylabel={{Number of Environments\\ with Best Performance}},
		xlabel style={font=\small},
		ylabel style={font=\small, align=center},
		xmin=-5,
		xmax=900,
		ymin=-5,
		ymax=100
		]
		\def\controller{LloydLocalVoronoi}
		\addplot[forget plot, mark=none, mark size=5pt,draw=mDarkRed, thick]  table[x=step, y=\controller, col sep=comma] {\costdata} node[mylabel, yshift=1pt] {\textcolor{mDarkRed!80}{\scriptsize Decentralized CVT}};

		\def\controller{LloydLocalSensorGlobalComm}
		\addplot[forget plot, mark=none, mark size=5pt,draw=mGreen, thick]  table[x=step, y=\controller, col sep=comma] {\costdata} node[mylabel, yshift=3.0pt] {\textcolor{mGreen}{\scriptsize Centralized CVT}};

		\def\controller{CNNGNN_k3}
		\addplot[forget plot, mark=none, mark size=5pt,draw=mDarkBlue, thick]  table[x=step, y=\controller, col sep=comma] {\costdata} node[mylabel, yshift=+2.0pt] {\textcolor{mDarkBlue!70}{\scriptsize LPAC-K3}};

	\end{axis}
\end{tikzpicture}
  \caption{Number of environments for which each controller performs the best (after the first time step):
    The controllers are run for 100 environments for 900 time steps each.
    The environments are randomly generated---the center of the IDF features and the initial positions of the robots are sampled uniformly at random.
  The LPAC-K3 model performs best for about 50\% of the environments within the first few time steps and reaches above 85\% after 500 time steps.\label{fig:env_performance}}
\end{figure}
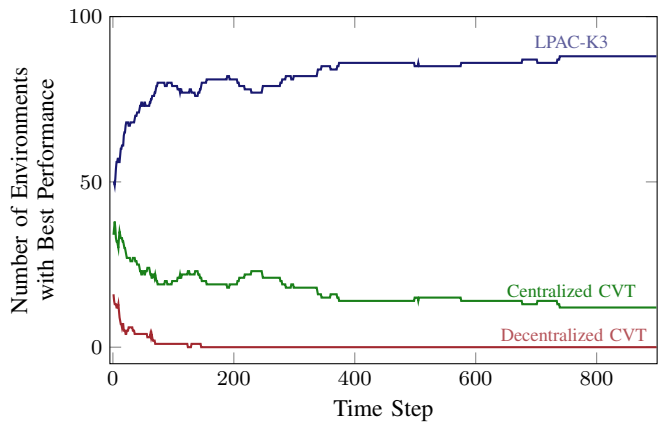

\fgref{fig:env_performance} shows the number of environments, out of the 100 environments, for which each controller performs the best.
The LPAC-K3 model performs the best for about 50\% of the environments within the first few time steps and reaches above 85\% after 500 time steps.
This indicates that the LPAC model can learn a policy that performs well in most environments over time.

These results establish that the LPAC architecture is suitable for the coverage control problem and outperforms the decentralized CVT algorithm.
Even though the centralized CVT algorithm has knowledge of all the observations of the robots, the LPAC model still outperforms the centralized CVT algorithm.
This indicates that the LPAC model learns to share abstract information about the environment with other robots and propagates this information over the communication graph, resulting in improved performance.

\subsection{Ablation Study}
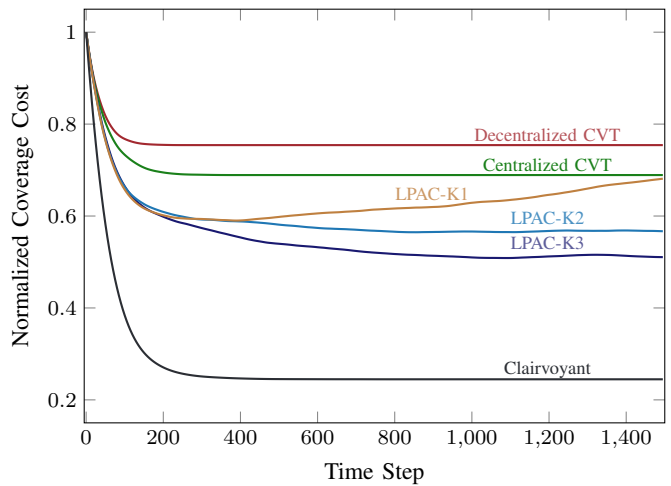
\begin{figure}[htp]
  \begin{tikzpicture}
	\small
	\tikzset{
		mylabel/.style={pos=0.80, above, yshift=-2.5pt}
	}
	\footnotesize
	\pgfplotstableread[col sep=comma]{./figures/data/normalized_cost_time_all_1024.csv}{\costdata}
	\begin{axis}[
		width=1.05\columnwidth,
		height=0.8\columnwidth,
		xlabel={Time Step},
		ylabel={Normalized Coverage Cost},
		xlabel style={font=\small},
		ylabel style={font=\small},
		legend style={font=\scriptsize, at={(0.50,1.15)},anchor=north},
		legend columns=-1,
		xmin=-5,
		xmax=1500,
		ymin=0.15,
		ymax=1.05,
		]
		\def\controller{LloydLocalVoronoi}
		\addplot[forget plot, mark=none, mark size=5pt,draw=mDarkRed, thick]  table[x=step, y=\controller, col sep=comma] {\costdata} node[mylabel, yshift=1pt] {\textcolor{mDarkRed!80}{\scriptsize Decentralized CVT}};

		\def\controller{LloydLocalSensorGlobalComm}
		\addplot[forget plot, mark=none, mark size=5pt,draw=mGreen, thick]  table[x=step, y=\controller, col sep=comma] {\costdata} node[mylabel, yshift=1pt] {\textcolor{mGreen}{\scriptsize Centralized CVT}};

		\def\controller{CNNGNN_k3}
		\addplot[forget plot, mark=none, mark size=5pt,draw=mDarkBlue, thick]  table[x=step, y=\controller, col sep=comma] {\costdata} node[mylabel, yshift=+2.0pt] {\textcolor{mDarkBlue!70}{\scriptsize LPAC-K3}};

		\def\controller{CNNGNN_k2}
		\addplot[forget plot, mark=none, mark size=5pt,draw=mBlue, thick]  table[x=step, y=\controller, col sep=comma] {\costdata} node[mylabel, yshift=+2.0pt] {\textcolor{mBlue!80}{\scriptsize LPAC-K2}};

		\def\controller{CNNGNN_k1}
		\addplot[forget plot, mark=none, mark size=5pt,draw=brown, thick]  table[x=step, y=\controller, col sep=comma] {\costdata} node[pos=0.6, yshift=+5.0pt] {\textcolor{brown!80}{\scriptsize LPAC-K1}};

		\def\controller{LloydGlobalOnline}
		\addplot[forget plot, mark=none, mark size=5pt,draw=mSteelGray, thick]  table[x=step, y=\controller, col sep=comma] {\costdata} node[mylabel] {\textcolor{mSteelGray}{\scriptsize Clairvoyant}};
	\end{axis}
\end{tikzpicture}
  \caption{Performance of LPAC architecture with different number of hops $K$ in the GNN layer:
    The controllers are run for 100 environments for 1500 time steps each.
    All the LPAC models perform well in the first 200 time steps, but the LPAC-K3 model with $3$ hops in the GNN layer outperforms the other LPAC models in the long run.
    The LPAC-K1 model with a single hop in the GNN layer is particularly unstable, as it is not able to learn to collaborate with the robots that are not in its communication range.
    The LPAC-K2 and LPAC-K3 models outperform both the centralized and decentralized CVT algorithms. \label{fig:cost_time}}
\end{figure}

We performed an ablation study of the LPAC architecture to understand the importance of the different components of the architecture.
For the first ablation study, we trained LPAC models with different numbers of hops in the GNN architecture, ranging from 1 to 3.
A $K=3$ hop model enables the diffusion of information over a larger portion of the communication graph, whereas a $K=1$ hop model limits diffusion to the immediate neighbors of a robot.
\fgref{fig:cost_time} shows the performance of the LPAC models with different numbers of hops in the GNN architecture and compares them with the decentralized and centralized CVT algorithms.
The LPAC-K1 model performs worse than the other LPAC models and is unstable over time.
This is expected, as the model is not able to collaborate with robots that are not within its communication range.
The LPAC-K2 and LPAC-K3 models, on the other hand, significantly outperform the CVT algorithms.
A large number of hops in the GNN architecture is not necessarily advantageous, as it requires more computation and memory and may become unsuitable for real-time applications.

\begin{figure}[ht]
  \begin{tikzpicture}
	\small
	\tikzset{
		mylabel/.style={pos=0.80, above, yshift=-2.5pt}
	}
	\footnotesize
	\pgfplotstableread[col sep=comma]{./figures/data/normalized_cost_time_ablation_1024.csv}{\costdata}
	\begin{axis}[
		width=1.05\columnwidth,
		height=0.8\columnwidth,
		xlabel={Time Step},
		ylabel={Normalized Coverage Cost},
		xlabel style={font=\small},
		ylabel style={font=\small},
		legend style={font=\scriptsize, at={(0.50,1.15)},anchor=north},
		legend columns=-1,
		xmin=-5,
		xmax=900,
		ymin=0.15,
		ymax=1.05,
		]
		\def\controller{LloydLocalVoronoi}
		\addplot[forget plot, mark=none, mark size=5pt,draw=mDarkRed, thick]  table[x=step, y=\controller, col sep=comma] {\costdata} node[mylabel, yshift=1pt] {\textcolor{mDarkRed!80}{\scriptsize Decentralized CVT}};

		\def\controller{LloydLocalSensorGlobalComm}
		\addplot[forget plot, mark=none, mark size=5pt,draw=mGreen, thick]  table[x=step, y=\controller, col sep=comma] {\costdata} node[mylabel, yshift=1pt] {\textcolor{mGreen}{\scriptsize Centralized CVT}};

		\def\controller{CNNGNN_k3}
		\addplot[forget plot, mark=none, mark size=5pt,draw=mDarkBlue, thick]  table[x=step, y=\controller, col sep=comma] {\costdata} node[mylabel, yshift=+2.0pt] {\textcolor{mDarkBlue!70}{\scriptsize LPAC-K3}};

		\def\controller{CNNGNN_nocomm_k3}
		\addplot[forget plot, mark=none, mark size=5pt,draw=mBlue, thick]  table[x=step, y=\controller, col sep=comma] {\costdata} node[mylabel, yshift=+0.2pt] {\textcolor{mBlue!80}{\scriptsize LPAC-NoNeighborMap}};

		\def\controller{CNNGNN_nopos_k3}
		\addplot[forget plot, mark=none, mark size=5pt,draw=brown, thick]  table[x=step, y=\controller, col sep=comma] {\costdata} node[mylabel, yshift=+1.0pt] {\textcolor{brown!80}{\scriptsize LPAC-NoPos}};

		\def\controller{LloydGlobalOnline}
		\addplot[forget plot, mark=none, mark size=5pt,draw=mSteelGray, thick]  table[x=step, y=\controller, col sep=comma] {\costdata} node[mylabel] {\textcolor{mSteelGray}{\scriptsize Clairvoyant}};
	\end{axis}
\end{tikzpicture}
  \caption{Ablation study of the LPAC model:
    The LPAC architecture without the neighbor maps (LPAC-NoNeighborMaps) as inputs to the CNN layer performs significantly worse than the full LPAC model, even though the average performance is better than both the centralized and decentralized CVT algorithms.
    The LPAC architecture (LPAC-NoPos), without the normalized robot positions as additional features to the GNN layer, performs slightly worse than the full LPAC model.
  This indicates that even when a global position estimate is not available, the LPAC model can still perform well. \label{fig:cost_time_ablation}}
\end{figure}
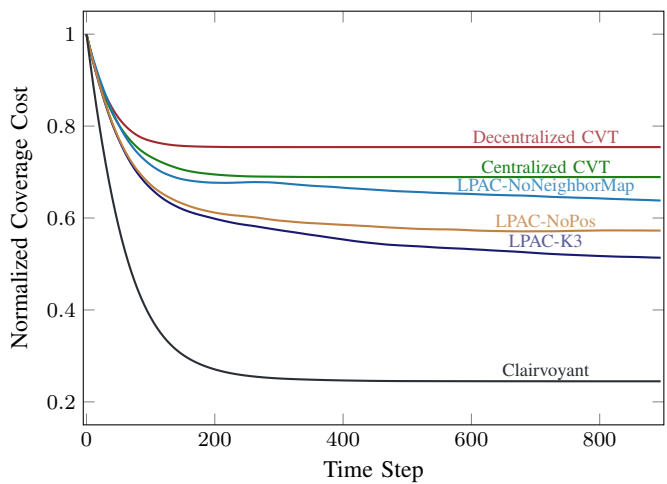

For the second ablation study, we trained LPAC models without the neighbor maps as inputs to the CNN layer (LPAC-NoNeighborMaps), and without the normalized robot positions as additional features to the GNN layer (LPAC-NoPos).
\fgref{fig:cost_time_ablation} shows the performance of these ablated LPAC models with respect to the primary LPAC model (LPAC-K3) and the CVT algorithms.
The LPAC-NoNeighborMaps model performs significantly worse than the LPAC-K3 model.
This indicates that the neighbor maps are essential for the LPAC model to perform well.
The LPAC-NoPos model performs slightly worse than the LPAC-K3 model, but the performance is still better than the CVT algorithms.
The normalized position of the robot as an additional feature to the GNN layer is, therefore, helpful in improving the performance of the LPAC model.
However, the LPAC model can still perform well even when the position of the robot is not available, as in the case of a robot without a GPS or good state estimation.

The ablation study shows that the proposed LPAC architecture is suitable for the coverage control problem, and the different components of the architecture are essential for the performance of the model.
In the following subsections, we evaluate the performance of the LPAC architecture, focusing on $K=3$ hops (LPAC-K3) in the GNN architecture as the primary model.

\subsection{Generalization to Varying Number of Robots and Features} \label{sc:generalization}
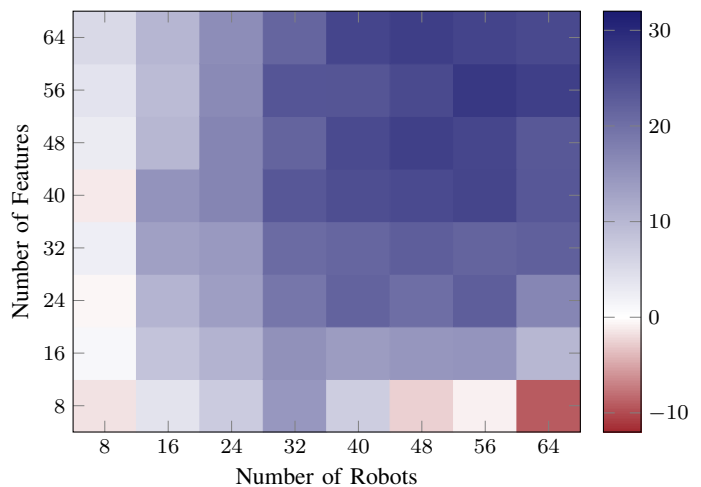
\begin{figure}[t]
  \begin{tikzpicture}
	\pgfplotsset{
		colormap={violet}{color(-10)=(mDarkRed) color(0)=(white) color(27)=(mDarkBlue)}
	}
	\footnotesize
	\begin{axis}[
		view={0}{90},   
		xlabel={Number of Robots},
		ylabel={Number of Features},
		xlabel style={font=\small},
		ylabel style={font=\small},
		width=0.94\columnwidth,
		xmin=4,
		xmax=68,
		ymin=4,
		ymax=68,
		xtick={8,16,24,32,40,48,56,64},
		ytick={8,16,24,32,40,48,56,64},
		colorbar,
		colorbar style={
			yticklabel style={
				/pgf/number format/.cd,
				fixed,
				precision=0,
				fixed zerofill,
			},
			colormap name=violet,
		},
		enlargelimits=false,
		axis on top,
		point meta min=-12,
		point meta max=32,
		]
		\addplot [matrix plot*,point meta=explicit] table [x=x,y=y,meta=C] {./figures/data/rf_avg_diff.dat};
	\end{axis}
\end{tikzpicture}
  \caption{Evaluation of the LPAC-K3 model on environments with varying numbers of features and robots.
    The controllers are evaluated in 100 environments for each combination of the number of robots and features.
    The performance is computed as the percentage improvement of the average cost over the decentralized CVT algorithm.
    The model performs worse than the decentralized CVT algorithm when the number of robots or features is very small (8 for each).
    However, the model performs significantly better when the number of robots and features increases.
  \label{fig:rf}}
\end{figure}

The primary LPAC model (LPAC-K3) was trained on environments with 32 robots and 32 features in a \SImSqrDim{1024} environment.
We evaluate the performance of the model on environments with varying numbers of robots and features from 8 to 64, in step of 8, in the same size environment.
For each combination of a number of robots and features, we generated 100 environments with the locations of the features and the initial positions of the robots sampled uniformly at random.
The controllers are run for 1500 time steps for each environment.
\fgref{fig:rf} shows the performance of the LPAC model, with $K=3$ hops in the GNN architecture, with respect to the decentralized CVT algorithm.
The performance is computed as the percentage improvement of the average cost over the decentralized CVT (D-CVT) algorithm:
\begin{equation}
  \frac{(\text{Avg. Cost D-CVT}) - (\text{Avg. Cost LPAC-K3})}{(\text{Avg. Cost \text{D-CVT})}} \times 100
\end{equation}

The LPAC model performs worse than the decentralized CVT algorithm when the number of robots or features is very small (8 for each).
The sensor field-of-view of the robots is \numproduct{64x64}, whereas the environment size is \numproduct{1024x1024}.
This represents a very small fraction of the environment that the robots can observe, with a ratio of $1:256$.
When the number of features is very small, large empty regions are problematic as the robots may not see any features, especially in the first few steps, and thus, the LPAC policy fails to take appropriate actions.
The LPAC policy does not often see such scenarios during training, and hence, the policy may not perform well in such cases.
For example, the LPAC model performs 9\% worse than the decentralized CVT algorithm with 8 features, even with 64 robots.

A similar case arises when the number of robots is very low, resulting in a sparse disconnected communication graph.
The GNN-based LPAC policy is not effective in such scenarios as the robots are not able to communicate with each other.

However, the model performs significantly better for a larger number of robots and features.
Interestingly, the performance of the LPAC model improves for cases beyond the training distribution, i.e., when the number of robots and features is larger than~32.
With 32~robots and 32~features and beyond, the LPAC model performs at least 20\% better than the decentralized CVT algorithm.

These results demonstrate that the LPAC model can generalize effectively to scenarios featuring substantially more robots and features than those used during training.
This capability is significant, as it allows the model that was trained on a sufficiently large, but finite set of conditions, to be reliably applied in environments where the number of robots and features is not fixed in advance and may far surpass the levels encountered during training.
This empirical results correlate with the theoretical results on transferability~\cite{RuizGR21}, i.e., policies trained on sufficiently large number of robots can be transferred to even larger teams.

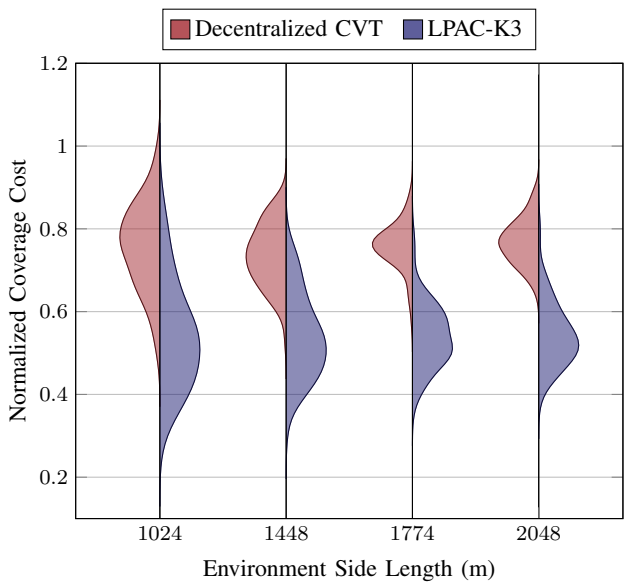
\begin{figure}[tbp]
  \centering
  \pgfplotsset{width=\linewidth}
\begin{tikzpicture}
	\footnotesize
\violinsetoptions[
no mirror,
scaled,
]{%
	xmin=0,xmax=13,
	ymin=0.1,ymax=1.2,
	xlabel style={
		yshift = {-3*width("a")},
	},
	ymajorgrids=true,
	legend pos=north west,
}
\violinplot[%
index=LloydLocalVoronoi,
relative position=2,
color=mDarkRed,
invert=true,
label={$1024$},
]{./figures/data/normalized_cost_env_1024.csv}
\violinplot[%
index=CNNGNN_k3,
relative position=2,
color=mDarkBlue,
label={}
]{./figures/data/normalized_cost_env_1024.csv}
\violinplot[%
index=LloydLocalVoronoi,
relative position=5,
color=mDarkRed,
invert=true,
label={$1448$}
]{./figures/data/normalized_cost_env_1448.csv}
\violinplot[%
index=CNNGNN_k3,
relative position=5,
color=mDarkBlue,
label={}
]{./figures/data/normalized_cost_env_1448.csv}
\violinplot[%
index=LloydLocalVoronoi,
relative position=8,
color=mDarkRed,
invert=true,
label={$1774$}
]{./figures/data/normalized_cost_env_1774.csv}
\violinplot[%
index=CNNGNN_k3,
relative position=8,
color=mDarkBlue,
label={}
]{./figures/data/normalized_cost_env_1774.csv}
\violinplot[%
index=LloydLocalVoronoi,
relative position=11,
color=mDarkRed,
invert=true,
label={$2048$}
]{./figures/data/normalized_cost_env_2048.csv}
\violinplot[%
index=CNNGNN_k3,
relative position=11,
color=mDarkBlue,
label={}
]{./figures/data/normalized_cost_env_2048.csv}
\begin{axis}[
	xmin=0,xmax=13,
	ymin=0.1,ymax=1.2,
    width=\columnwidth,
	legend style={/tikz/every even column/.append style={column sep=0.2cm}},
	legend style={font=\small, at={(0.50,1.12)},anchor=north},
	legend columns=-1,
	legend entries={Decentralized CVT, LPAC-K3},
    xlabel={Environment Side Length (m)},
	ylabel={Normalized Coverage Cost},
 	xlabel style={
		yshift = {-2*width("a")},
        font = \small,
	},
  	ylabel style={
		yshift = {4*width("a")},
        font = \small,
	},
	]
	\addplot[only marks, mark=square*, mark options={scale=1.8, fill=mDarkRed!80}] coordinates{(NaN,NaN)};
	\addlegendimage{only marks, mark=square*, mark options={scale=1.8, fill=mDarkRed!80}}
	\addplot[only marks, mark=square*, mark options={scale=1.8, fill=mDarkBlue}] coordinates{(NaN,NaN)};
	\addlegendimage{only marks, mark=square*, mark options={scale=1.8, fill=mDarkBlue!70}}
\end{axis}
\end{tikzpicture}
\pgfplotsset{width=\columnwidth}
  \caption{Scalability of LPAC models: The LPAC-K3 model, trained on environments of size \SImSqrDim{1024} with 32 robots, is executed on larger environments with a larger number of robots while keeping the ratio of robots to environment size the same.
    In the violin plots, the LPAC-K3 model (blue) consistently outperforms the decentralized CVT algorithm (red) in all cases.
    The width of the violin plots is proportional to the number of environments for the average coverage cost.
  \label{fig:violin}}
\end{figure}

\subsection{Transferability to Larger Environments}
One of the advantages of using a GNN-based architecture is that it scales well to larger environments and a larger number of robots.
We evaluate the primary LPAC model (LPAC-K3), which was trained on environments with 32 robots and 32 features in a \SImSqrDim{1024} environment, on larger environments while keeping the number of robots and features per unit of environment area the same.
The model is not retrained or fine-tuned on the larger environments, and the robot parameters, such as the communication range (\SIm{128}) and the sensor field of view (\SImSqrDim{64}), are kept the same.
\fgref{fig:violin} shows the performance of the LPAC model with respect to the decentralized CVT algorithm on environments ranging from \SImSqrDim{1024} to \SImSqrDim{2048} with 32 to 128 robots.
The shaded regions in the figure show the distribution of the normalized coverage cost over 100 environments.
Note that the peaks of the distributions for the LPAC model are at a significantly lower coverage cost than those of the decentralized CVT algorithm.
Interestingly, the spread of the distributions for both controllers decreases as the size of the environment increases, even though the ratio of robots to environment size is kept the same.
This could be attributed to the fact that the ratio of environment size and the number of robots does not exactly match with the expected degree of a robot in the communication graph---the expected degree of a robot is 1.41 in the \SImSqrDim{1024} environment, and 1.49 in the \SImSqrDim{2048} environment, as computed using a Monte Carlo simulation.
The discrepancy is caused because of the boundary limits of the environment, which changes the probability of a robot being connected to other robots from a linear relationship to a more complex one.

\begin{figure}[t]
  \begin{tikzpicture}
	\pgfplotsset{
		colormap={violet}{color=(mDarkBlue) color(2.5)=(white) color=(mDarkRed)}
	}
	\footnotesize
	\begin{axis}[
		view={0}{90},   
		xlabel={\small Environment Side Length (m)},
		width=0.96\columnwidth,
		xmin=-0.5,
		xmax=3.5,
		ymin=-0.5,
		ymax=3.5,
		xtick={0, 1, 2, 3},
		ytick={0, 1, 2, 3},
		xticklabels={1024, 1448, 1774, 2048},
		yticklabels={D-CVT, C-CVT, LPAC-K2, LPAC-K3},
		y dir=reverse,
		yticklabel style={rotate=90},
		colorbar,
		colorbar style={
			yticklabel style={
				/pgf/number format/.cd,
				fixed,
				precision=2,
				fixed zerofill,
			},
			colormap name=violet,
		},
		enlargelimits=false,
		axis on top,
		point meta min=1.95,
		point meta max=3.05,
		]
		\addplot [matrix plot*,point meta=explicit] table [x=x,y=y,meta=C] {./figures/data/controller_env_heatmap.dat};
	\end{axis}
\end{tikzpicture}
  \caption{Performance of coverage control algorithms as a ratio of average costs with respect to the clairvoyant algorithm:
    The decentralized CVT (D-CVT) and centralized CVT (C-CVT) algorithms and the LPAC-K2 and LPAC-K3 models are evaluated on increasing environment size while keeping the ratio of robots to environment size the same.
    The LPAC models outperform both the decentralized and centralized CVT algorithms in all cases.
    The average costs for the D-CVT and the C-CVT algorithms are around 2.8 and 2.5 times that of the clairvoyant algorithm, respectively.
  The average costs for the LPAC-K3 model are about twice the clairvoyant algorithm, even though the model is decentralized and the IDF is not known \textit{a priori} to the model.\label{fig:controller_env_heatmap}}
\end{figure}
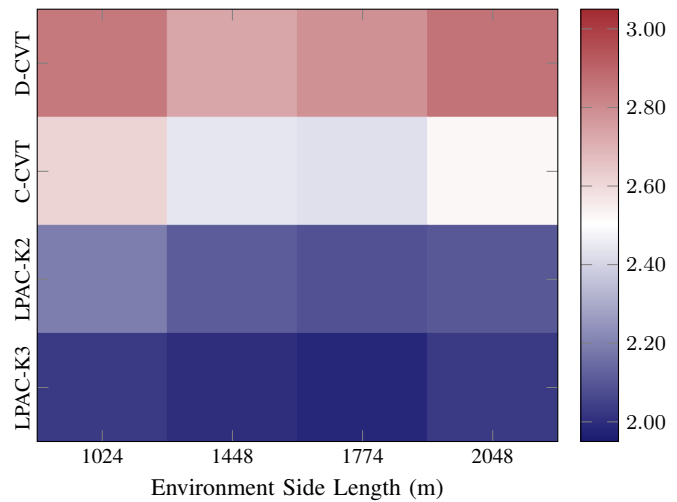

We also evaluated the centralized CVT algorithm and the LPAC model with $K=2$ hops in the GNN architecture in the same environments.
Similar to LPAC-K3, the LPAC-K2 model was trained on environments with 32 robots and 32 features in a \SImSqrDim{1024} environment.
\fgref{fig:controller_env_heatmap} shows the performance of the controllers in the larger environments.
The performance is measured as the ratio of the average coverage cost with respect to the clairvoyant algorithm, i.e., the average cost of the clairvoyant algorithm is~1.
Since the clairvoyant algorithm has the knowledge of the entire IDF, the coverage costs of other algorithms are always worse than the clairvoyant algorithm, and the ratio is always greater than~1.
The results show that both LPAC-K2 and LPAC-K3 models outperform the decentralized and centralized CVT algorithms for all environment sizes, and the models transfer well to larger environments.
The performance of the LPAC-K3 model is the best of all the controllers, whereas the performance of the LPAC-K2 model is slightly worse than that of the LPAC-K3 model.

These results establish that the LPAC architecture can be trained in a small environment and can be deployed in larger environments with a larger number of robots.
The results support the theoretical analysis on the transferability of GNNs~\cite{RuizGR21}.
The LPAC architecture utilizes CNN and GNN modules, and the inputs to these modules are designed such that both rely primarily on local, compositional structures rather than fixed global representations.
As the number of robots and the environment size grows, each robot's decision-making process remains focused on its immediate neighbors, ensuring that the learning process is independent of global scale.
Since the LPAC architecture, particularly GNN, is invariant to the number of vertices (robots), the same learned rules apply seamlessly to larger teams.
Moreover, as we scale the number of robots and maintain a constant density, the local neighborhood structure encountered by each robot remains essentially the same, creating a form of regularity.
This regularity in the local graph structure ensures that the learned local policies generalize effectively as the system expands.

\subsection{Robustness to Noisy Positions} \label{sc:robustness}
\begin{figure*}[ht]
  \centering
  \subfloat[$\epsilon \sim\mc N(0, 5^2)$]{\begin{tikzpicture}[scale=0.50]
	\small
	\tikzset{
		mylabel/.style={pos=0.80, above, yshift=-2.5pt}
	}
	\large
	\pgfplotstableread[col sep=comma]{./figures/data/noisy_cost_time_5.csv}{\costdata}
	\begin{axis}[
		width=0.5\textwidth,
		height=0.35\textwidth,
		xlabel={Time Step},
		ylabel={Normalized Coverage Cost},
		xlabel style={font=\large},
		ylabel style={font=\large},
		xmin=0,
		xmax=900,
		ymin=0.15,
		ymax=1.05,
		xmajorgrids=true,
		ymajorgrids=true,
		]
		\def\controller{LloydLocalVoronoi}
		\addplot[forget plot, mark=none, mark size=5pt,draw=mDarkRed, thick]  table[x=step, y=\controller, col sep=comma] {\costdata} node[mylabel, yshift=1pt] {\textcolor{mDarkRed!80}{\normalsize Decentralized CVT}};

		\def\controller{LloydLocalSensorGlobalComm}
		\addplot[forget plot, mark=none, mark size=5pt,draw=mGreen, thick]  table[x=step, y=\controller, col sep=comma] {\costdata} node[mylabel, yshift=1pt] {\textcolor{mGreen}{\normalsize Centralized CVT}};

		\def\controller{CNNGNN_k3}
		\addplot[forget plot, mark=none, mark size=5pt,draw=mDarkBlue, thick]  table[x=step, y=\controller, col sep=comma] {\costdata} node[mylabel, yshift=+2.0pt] {\textcolor{mDarkBlue!70}{\normalsize LPAC-K3}};

		\def\controller{LloydGlobalOnline}
		\addplot[forget plot, mark=none, mark size=5pt,draw=mSteelGray, thick]  table[x=step, y=\controller, col sep=comma] {\costdata} node[mylabel] {\textcolor{mSteelGray}{\normalsize Clairvoyant}};
	\end{axis}
\end{tikzpicture}}\hspace{0.1cm}
  \subfloat[$\epsilon \sim\mc N(0, 10^2)$]{\begin{tikzpicture}[scale=0.5]
	\small
	\tikzset{
		mylabel/.style={pos=0.80, above, yshift=-2.5pt}
	}
	\large
	\pgfplotstableread[col sep=comma]{./figures/data/noisy_cost_time_10.csv}{\costdata}
	\begin{axis}[
		width=0.5\textwidth,
		height=0.35\textwidth,
		xlabel={Time Step},
		xlabel style={font=\large},
		ylabel style={font=\large},
		xmin=0,
		xmax=900,
		ymin=0.15,
		ymax=1.05,
		xmajorgrids=true,
		ymajorgrids=true,
		]
		\def\controller{LloydLocalVoronoi}
		\addplot[forget plot, mark=none, mark size=5pt,draw=mDarkRed, thick]  table[x=step, y=\controller, col sep=comma] {\costdata} node[mylabel, yshift=1pt] {\textcolor{mDarkRed!80}{\normalsize Decentralized CVT}};

		\def\controller{LloydLocalSensorGlobalComm}
		\addplot[forget plot, mark=none, mark size=5pt,draw=mGreen, thick]  table[x=step, y=\controller, col sep=comma] {\costdata} node[mylabel, yshift=1pt] {\textcolor{mGreen}{\normalsize Centralized CVT}};

		\def\controller{CNNGNN_k3}
		\addplot[forget plot, mark=none, mark size=5pt,draw=mDarkBlue, thick]  table[x=step, y=\controller, col sep=comma] {\costdata} node[mylabel, yshift=+2.0pt] {\textcolor{mDarkBlue!70}{\normalsize LPAC-K3}};

		\def\controller{LloydGlobalOnline}
		\addplot[forget plot, mark=none, mark size=5pt,draw=mSteelGray, thick]  table[x=step, y=\controller, col sep=comma] {\costdata} node[mylabel] {\textcolor{mSteelGray}{\normalsize Clairvoyant}};
	\end{axis}
\end{tikzpicture}}\hspace{0.1cm}
  \subfloat[$\epsilon \sim\mc N(0, 15^2)$]{\begin{tikzpicture}[scale=0.5]
	\small
	\tikzset{
		mylabel/.style={pos=0.80, above, yshift=-2.5pt}
	}
	\large
	\pgfplotstableread[col sep=comma]{./figures/data/noisy_cost_time_15.csv}{\costdata}
	\begin{axis}[
		width=0.5\textwidth,
		height=0.35\textwidth,
		xlabel={Time Step},
		xlabel style={font=\large},
		ylabel style={font=\large},
		xmin=0,
		xmax=900,
		ymin=0.15,
		ymax=1.05,
		xmajorgrids=true,
		ymajorgrids=true,
		]
		\def\controller{LloydLocalVoronoi}
		\addplot[forget plot, mark=none, mark size=5pt,draw=mDarkRed, thick]  table[x=step, y=\controller, col sep=comma] {\costdata} node[mylabel, yshift=1pt] {\textcolor{mDarkRed!80}{\normalsize Decentralized CVT}};

		\def\controller{LloydLocalSensorGlobalComm}
		\addplot[forget plot, mark=none, mark size=5pt,draw=mGreen, thick]  table[x=step, y=\controller, col sep=comma] {\costdata} node[mylabel, yshift=1pt] {\textcolor{mGreen}{\normalsize Centralized CVT}};

		\def\controller{CNNGNN_k3}
		\addplot[forget plot, mark=none, mark size=5pt,draw=mDarkBlue, thick]  table[x=step, y=\controller, col sep=comma] {\costdata} node[mylabel, yshift=+2.0pt] {\textcolor{mDarkBlue!70}{\normalsize LPAC-K3}};

		\def\controller{LloydGlobalOnline}
		\addplot[forget plot, mark=none, mark size=5pt,draw=mSteelGray, thick]  table[x=step, y=\controller, col sep=comma] {\costdata} node[mylabel] {\textcolor{mSteelGray}{\normalsize Clairvoyant}};
	\end{axis}
\end{tikzpicture}}\hspace{0.1cm}
  \subfloat[$\epsilon \sim\mc N(0, 20^2)$]{\begin{tikzpicture}[scale=0.5]
	\small
	\tikzset{
		mylabel/.style={pos=0.80, above, yshift=-2.5pt}
	}
	\large
	\pgfplotstableread[col sep=comma]{./figures/data/noisy_cost_time_20.csv}{\costdata}
	\begin{axis}[
		width=0.5\textwidth,
		height=0.35\textwidth,
		xlabel={Time Step},
		xlabel style={font=\large},
		ylabel style={font=\large},
		xmin=0,
		xmax=900,
		ymin=0.15,
		ymax=1.05,
		xmajorgrids=true,
		ymajorgrids=true,
		]
		\def\controller{LloydLocalVoronoi}
		\addplot[forget plot, mark=none, mark size=5pt,draw=mDarkRed, thick]  table[x=step, y=\controller, col sep=comma] {\costdata} node[mylabel, yshift=2pt] {\textcolor{mDarkRed!80}{\normalsize Decentralized CVT}};

		\def\controller{LloydLocalSensorGlobalComm}
		\addplot[forget plot, mark=none, mark size=5pt,draw=mGreen, thick]  table[x=step, y=\controller, col sep=comma] {\costdata} node[mylabel, below, yshift=+4.2pt] {\textcolor{mGreen}{\normalsize Centralized CVT}};

		\def\controller{CNNGNN_k3}
		\addplot[forget plot, mark=none, mark size=5pt,draw=mDarkBlue, thick]  table[x=step, y=\controller, col sep=comma] {\costdata} node[mylabel, yshift=-9.0pt] {\textcolor{mDarkBlue!70}{\normalsize LPAC-K3}};

		\def\controller{LloydGlobalOnline}
		\addplot[forget plot, mark=none, mark size=5pt,draw=mSteelGray, thick]  table[x=step, y=\controller, col sep=comma] {\costdata} node[mylabel] {\textcolor{mSteelGray}{\normalsize Clairvoyant}};
	\end{axis}
\end{tikzpicture}}
  \caption{Evaluation of coverage algorithms with a Gaussian noise $\epsilon$ added to the position of each robot, i.e., the sensed position $\mv{\bar p}_i = \mv p_i + \epsilon$.
    The performance of the algorithms, measured as normalized coverage cost in the $y$-axis, is not significantly affected by the noise, except for the very large noise with a standard deviation of \SIm{20}.
  \label{fig:noisy}}
\end{figure*}
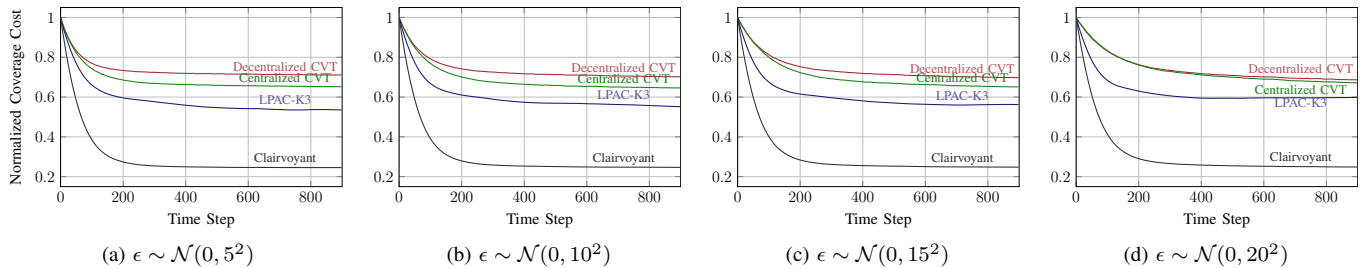

Real-world robots are subject to noise in their position estimates.
This is particularly true for robots that use GPS for localization, as the GPS signal can be affected by the environment, such as tall buildings and the weather.
Thus, we evaluate the performance of the LPAC-K3 model with respect to the CVT algorithms with simulated noise in the position of the robots.
A Gaussian noise is added to the position of each robot in the simulator.
The standard deviation of the Gaussian noise is varied from 5 to 20 meters in steps of 5 meters.
\fgref{fig:noisy} shows the performance of the controllers for different noise levels.
The controllers, including LPAC-K3, are not significantly affected by the noise.
However, there is still a degradation in the performance of the controllers with increasing noise, especially for the noise with a standard deviation of \SIm{20}.
These results indicate that LPAC models, as well as the CVT algorithms, are robust and can be deployed on real-world robots with noisy position estimates.

\subsection{Performance on Real-World Datasets} \label{sc:real-world}
In addition to testing on uniform, randomly generated feature distributions, we also evaluate our LPAC architecture using traffic light datasets derived from real-world traffic light locations from 50 cities in the US.
An area of \SImSqrDim{1024} was selected in each city, and the locations of the traffic lights were extracted from the OpenStreetMap database.
The traffic light locations form the locations of the features in the IDF.
The core idea behind using the traffic light dataset is to assess the generalizability of the LPAC policy under non-uniform and realistically structured feature distributions.
In our training and baseline evaluations, features (represented by Gaussian peaks in the IDF) are placed uniformly at random.
This setup, while controlled, does not reflect spatial patterns or correlations that often arise in real-world scenarios.

In contrast, the traffic light dataset introduces features located at fixed, non-uniform points that mirror real-world urban layouts.
Traffic lights are typically clustered along road networks, follow certain urban planning principles, and occur at predictable intervals rather than scattered randomly.
This is more than a trivial variation in feature placement as it forces the policy to address structured complexity rather than the uniform randomness it was exposed to during training.

\begin{figure*}[htbp]
  \centering
  \def\figwidth{0.24\textwidth}
  \subfloat{\begin{tikzpicture}\node [draw=none, rotate=90, minimum width=0.22\textwidth, inner sep=0cm] {\footnotesize Portland};\end{tikzpicture}}\hspace{0.03cm}
  \subfloat{\includegraphics[width=\figwidth]{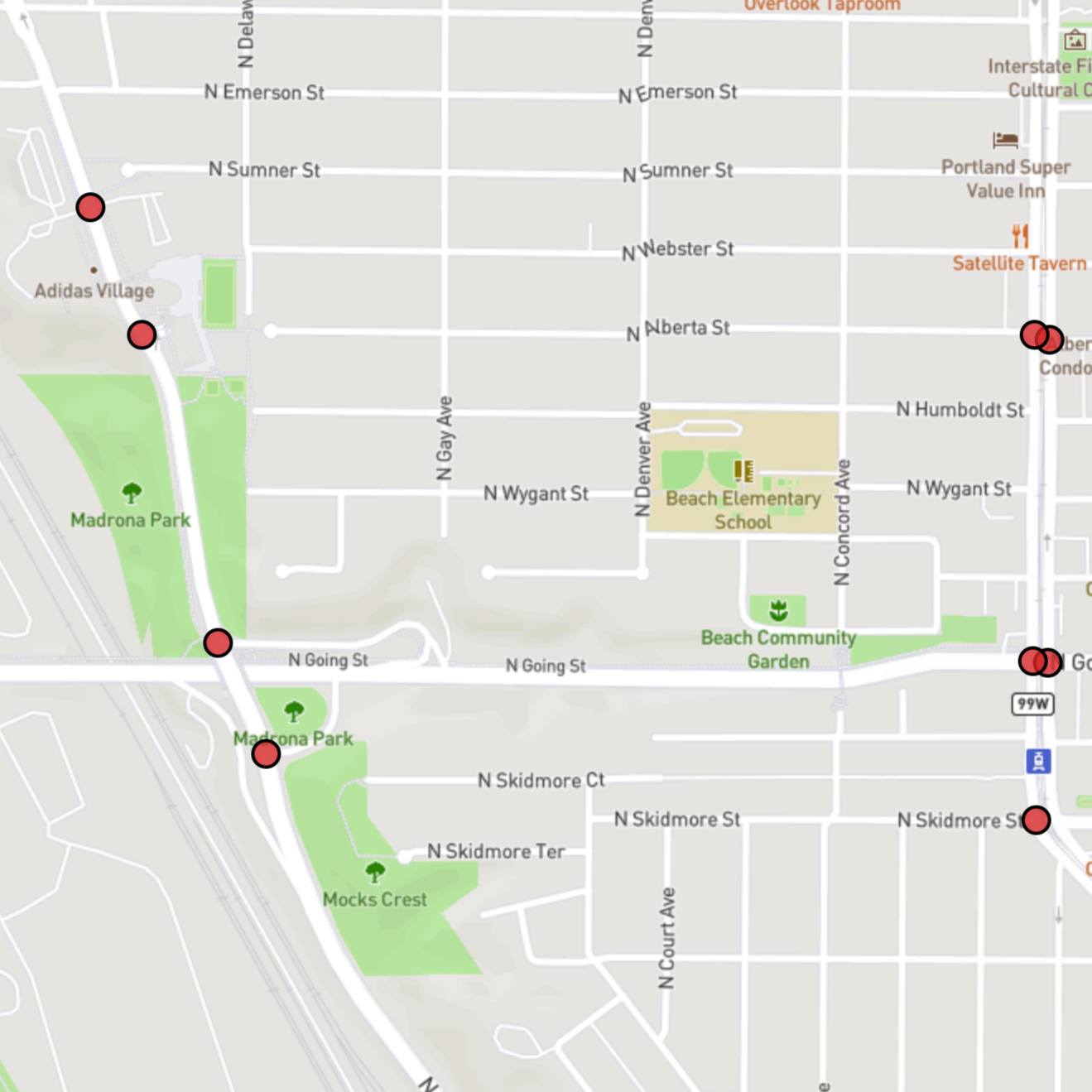}}\hspace{0.02cm}
  \subfloat{\includegraphics[width=\figwidth,trim={4.9cm 4cm 1.9cm 2.0cm},clip]{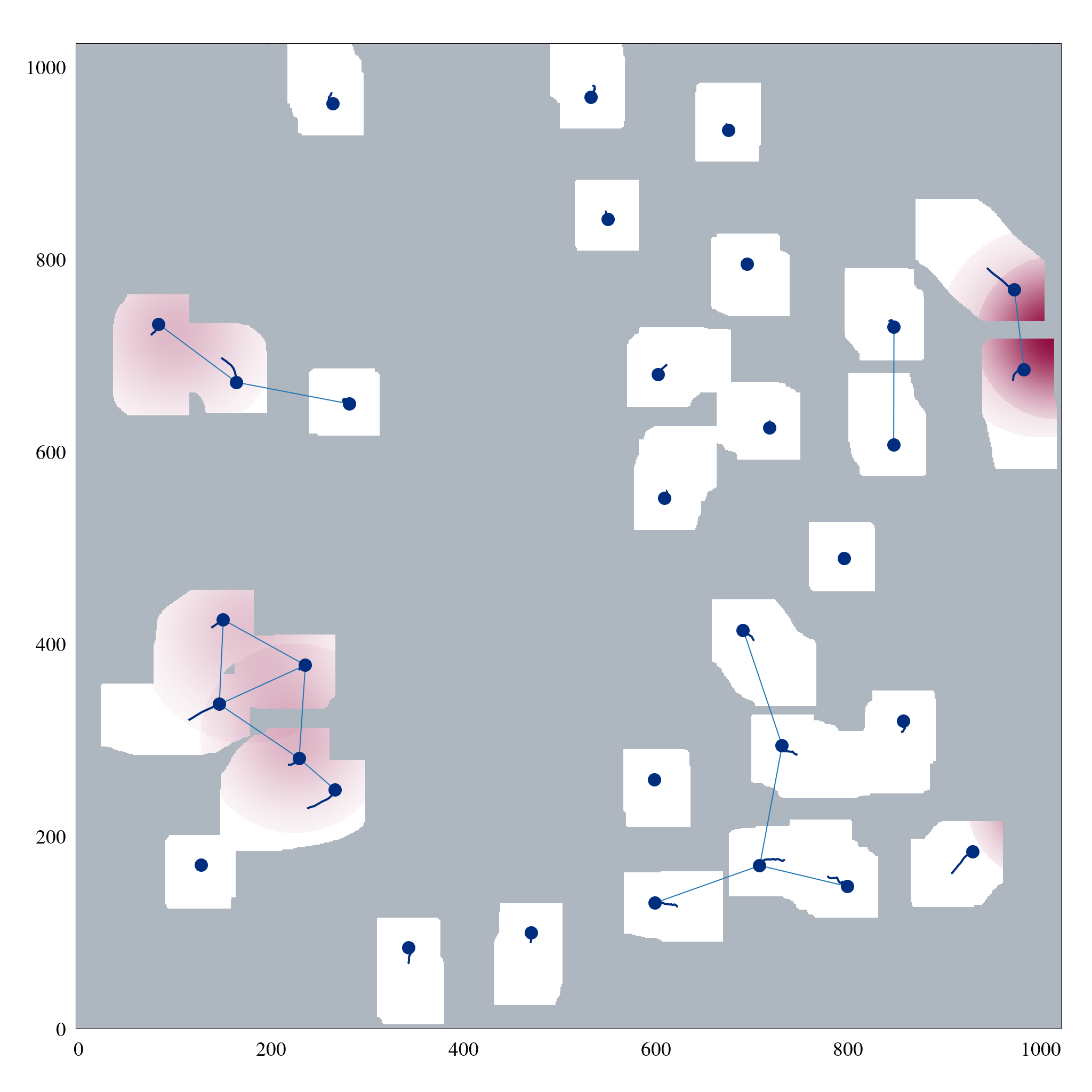}}\hspace{0.02cm}
  \subfloat{\includegraphics[width=\figwidth,trim={4.9cm 4cm 1.9cm 2.0cm},clip]{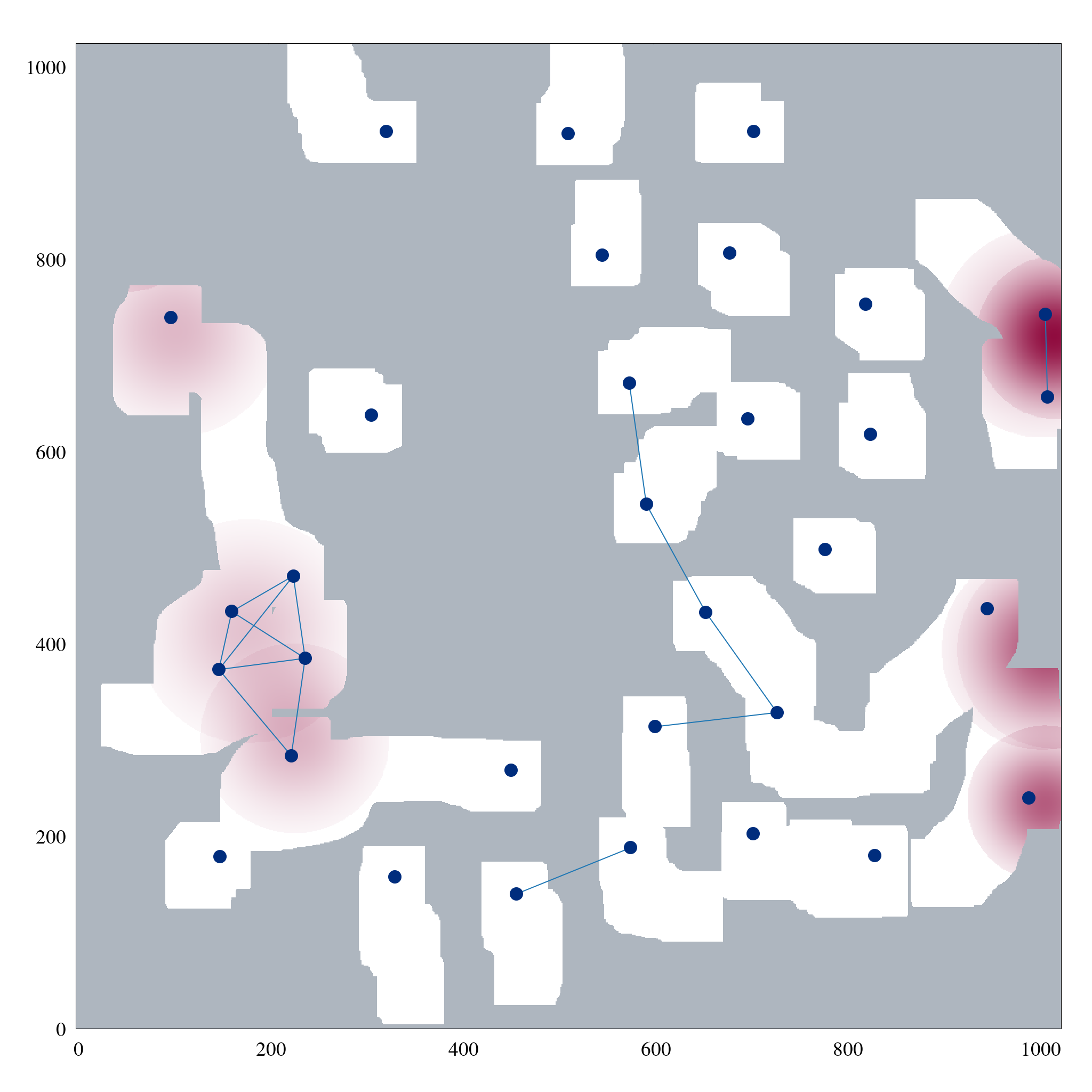}}\hspace{0.02cm}
  \subfloat{\includegraphics[width=\figwidth,trim={4.9cm 4cm 1.9cm 2.0cm},clip]{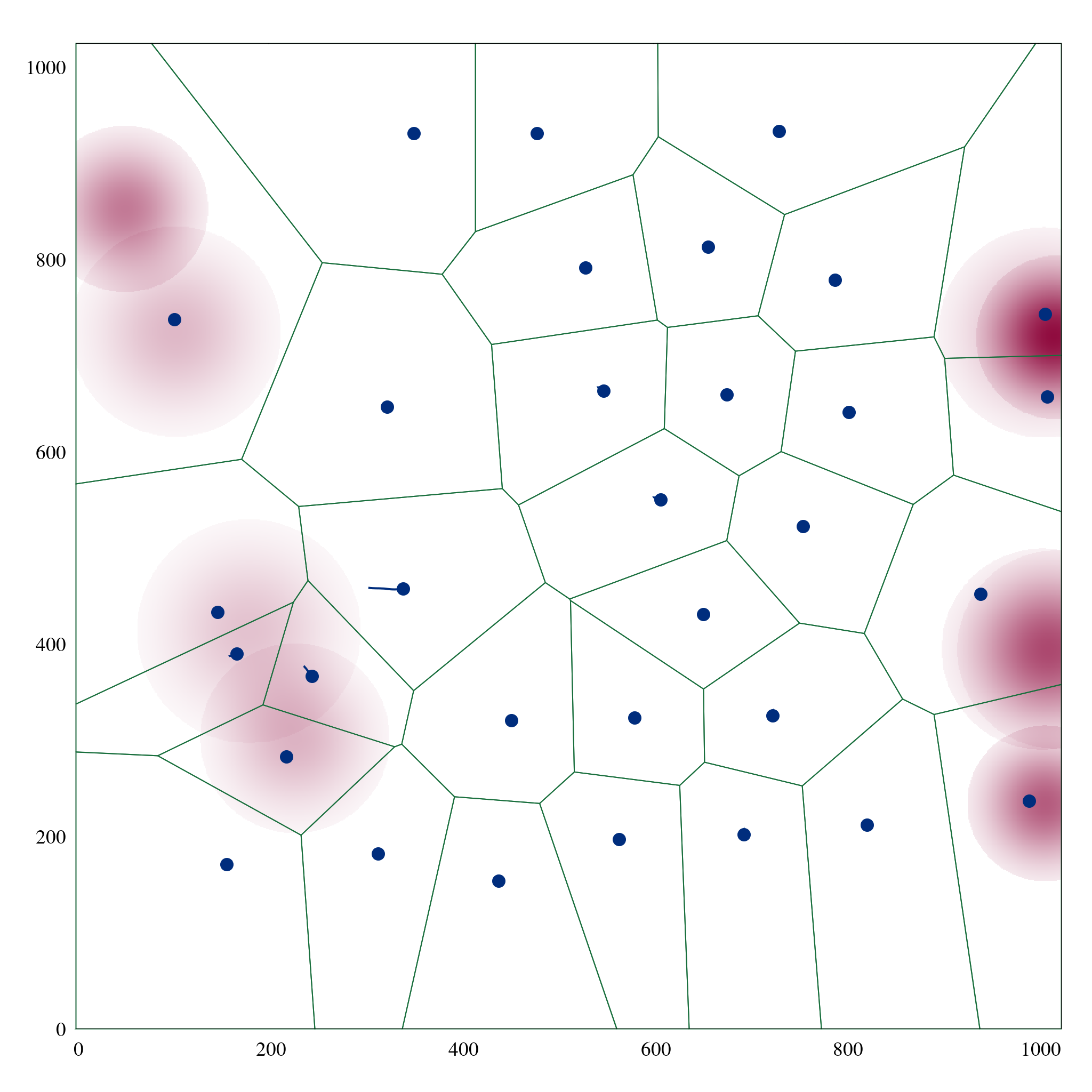}}\\
  \subfloat{\begin{tikzpicture}\node [draw=none, rotate=90, minimum width=0.22\textwidth, inner sep=0cm] {\footnotesize San Francisco};\end{tikzpicture}}\hspace{0.03cm}
  \subfloat{\includegraphics[width=\figwidth]{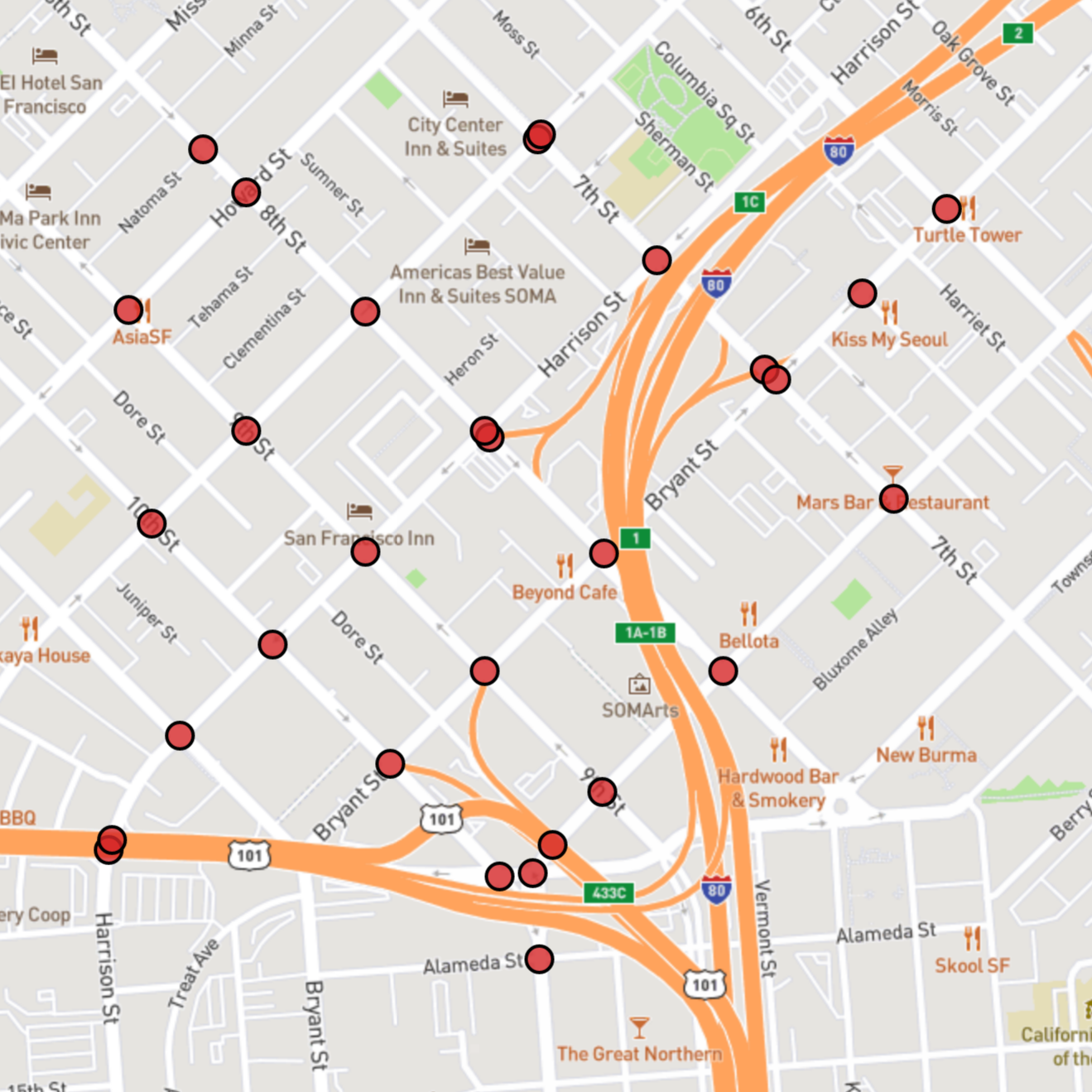}}\hspace{0.02cm}
  \subfloat{\includegraphics[width=\figwidth,trim={4.9cm 4cm 1.9cm 2.0cm},clip]{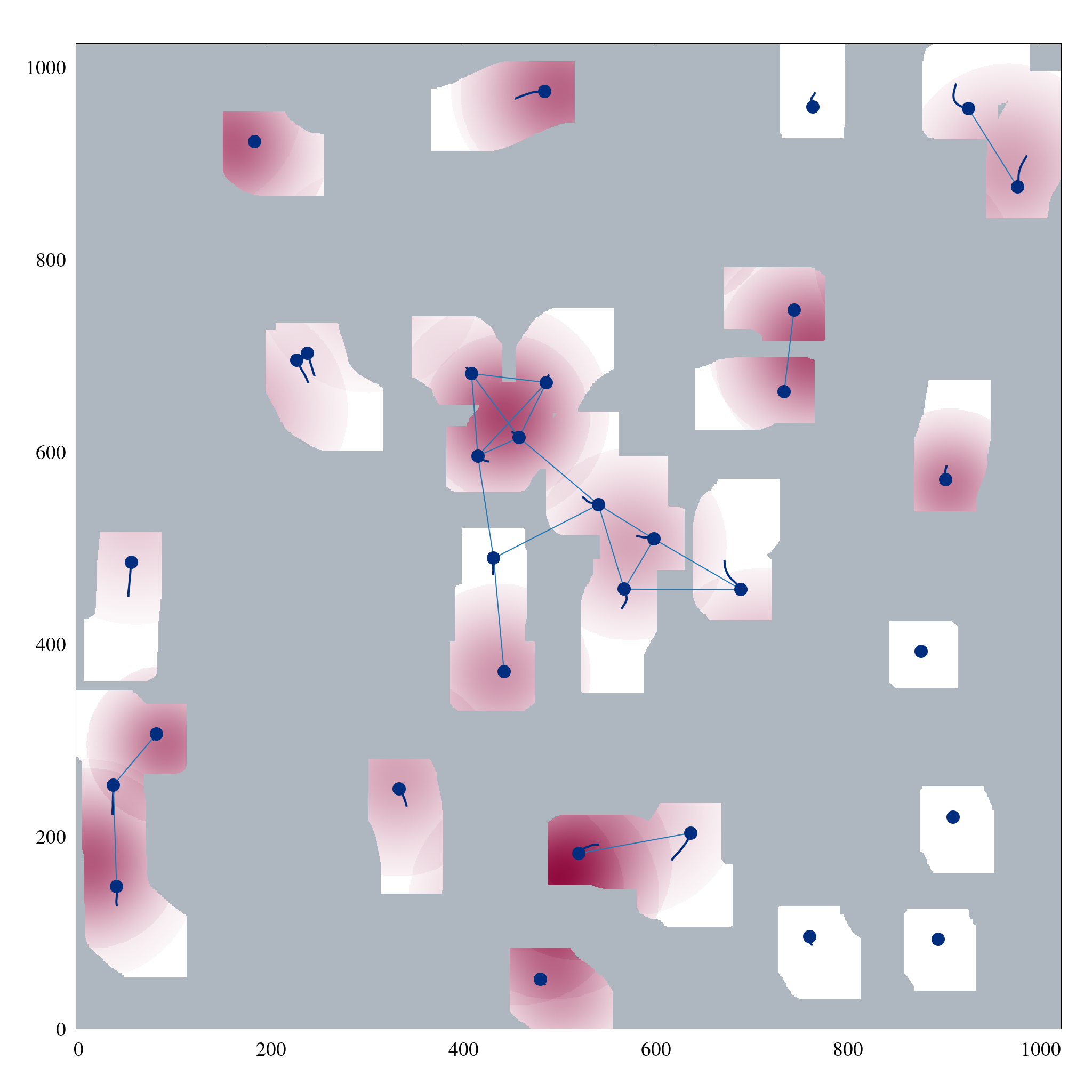}}\hspace{0.02cm}
  \subfloat{\includegraphics[width=\figwidth,trim={4.9cm 4cm 1.9cm 2.0cm},clip]{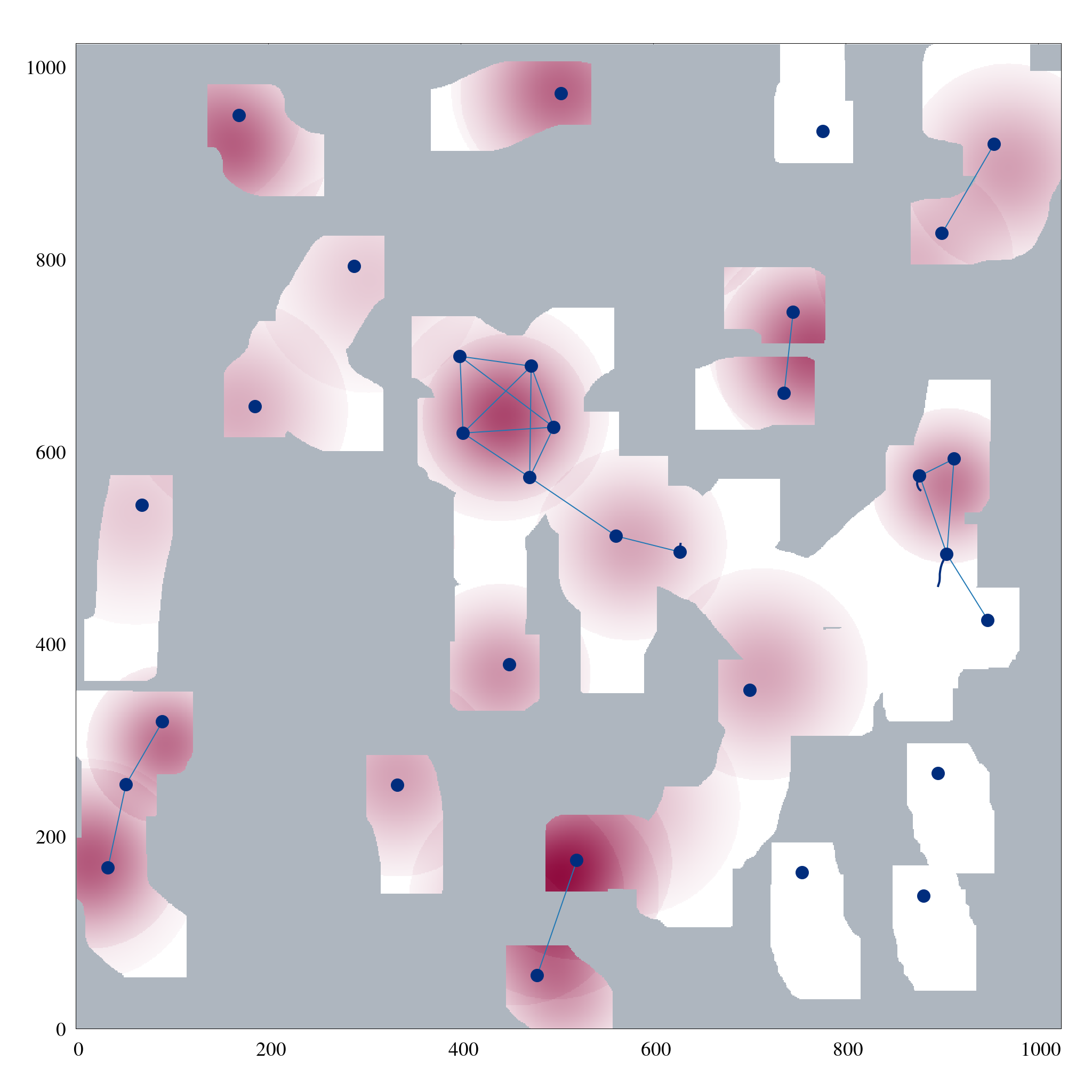}}\hspace{0.02cm}
  \subfloat{\includegraphics[width=\figwidth,trim={4.9cm 4cm 1.9cm 2.0cm},clip]{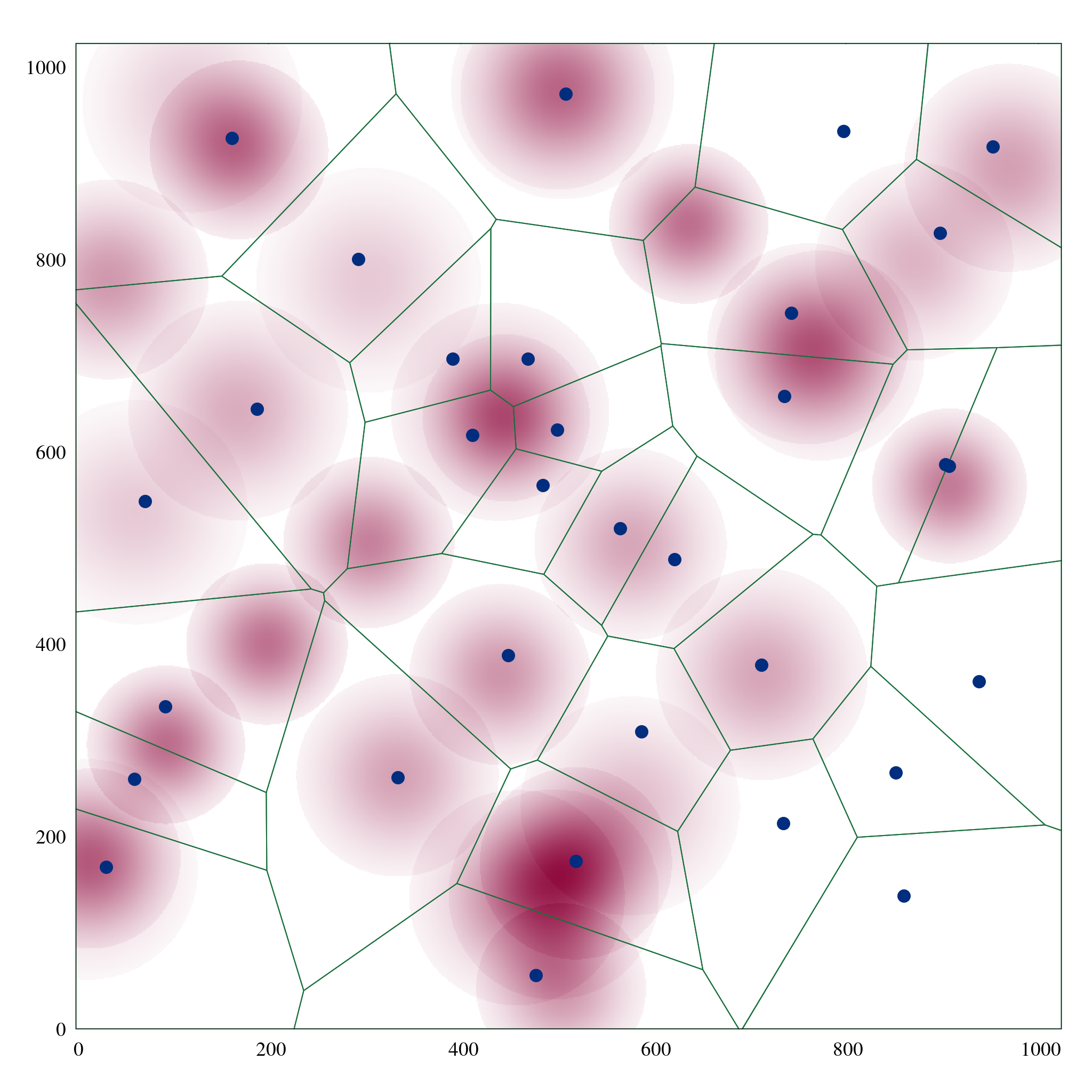}}\\
  \subfloat{\begin{tikzpicture}\node [draw=none, rotate=90, minimum width=0.22\textwidth, inner sep=0cm] {\footnotesize Boston};\end{tikzpicture}}\hspace{0.03cm} \setcounter{subfigure}{0}%
  \subfloat[Map]{\includegraphics[width=\figwidth]{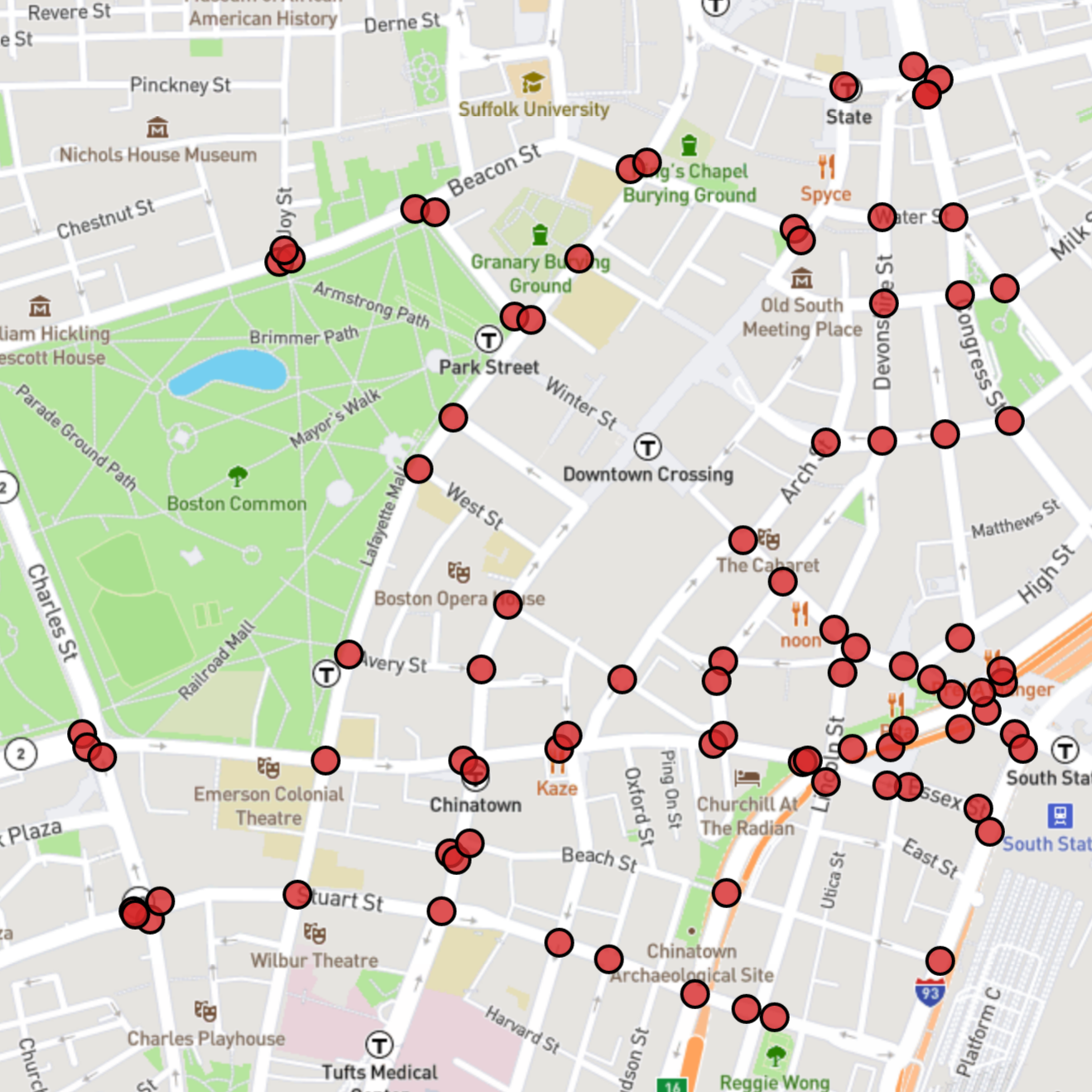}}\hspace{0.02cm}
  \subfloat[Time Step = 100]{\includegraphics[width=\figwidth,trim={4.9cm 4cm 1.9cm 2.0cm},clip]{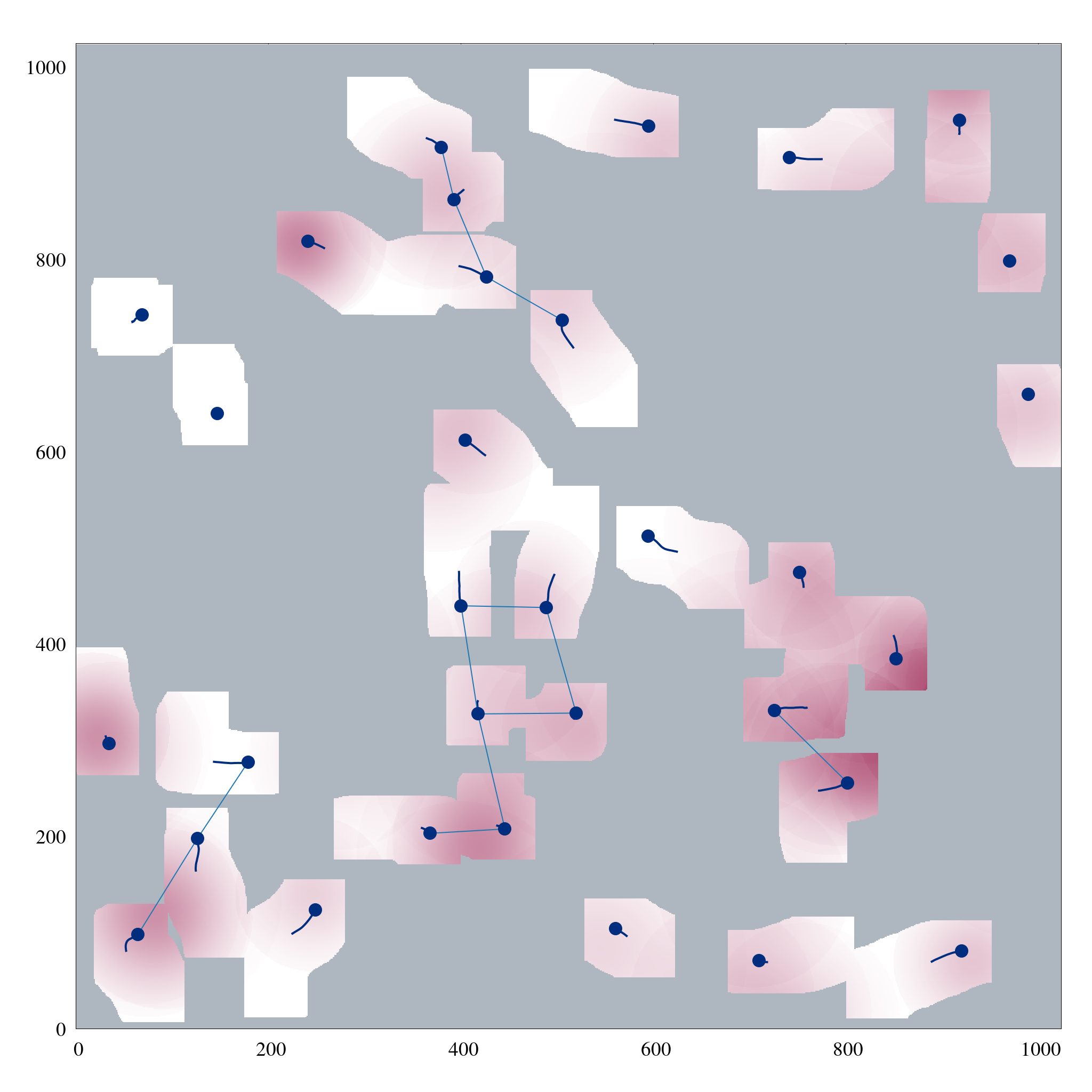}}\hspace{0.02cm}
  \subfloat[Time Step = 600]{\includegraphics[width=\figwidth,trim={4.9cm 4cm 1.9cm 2.0cm},clip]{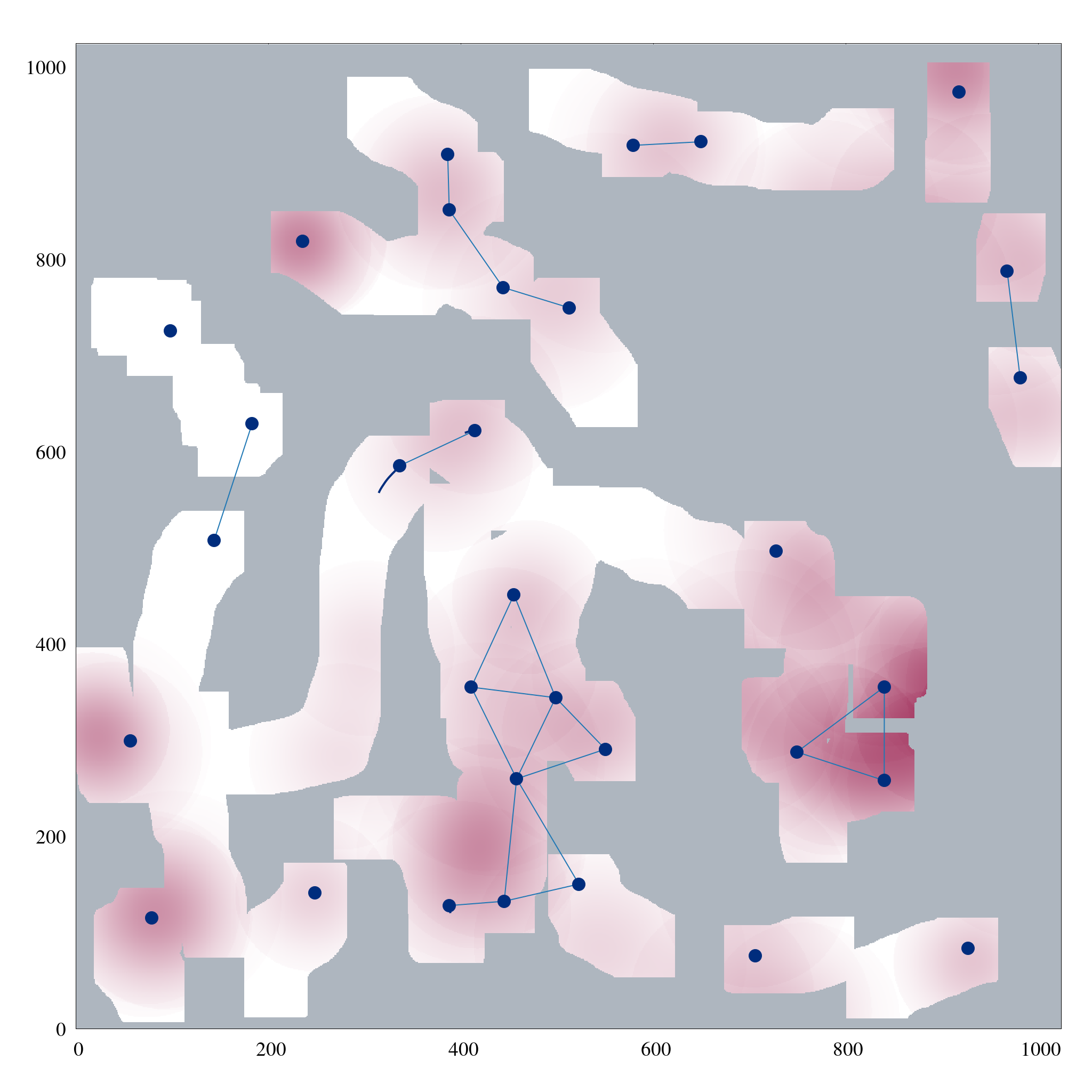}}\hspace{0.02cm}
  \subfloat[Time Step = 1200]{\includegraphics[width=\figwidth,trim={4.9cm 4cm 1.9cm 2.0cm},clip]{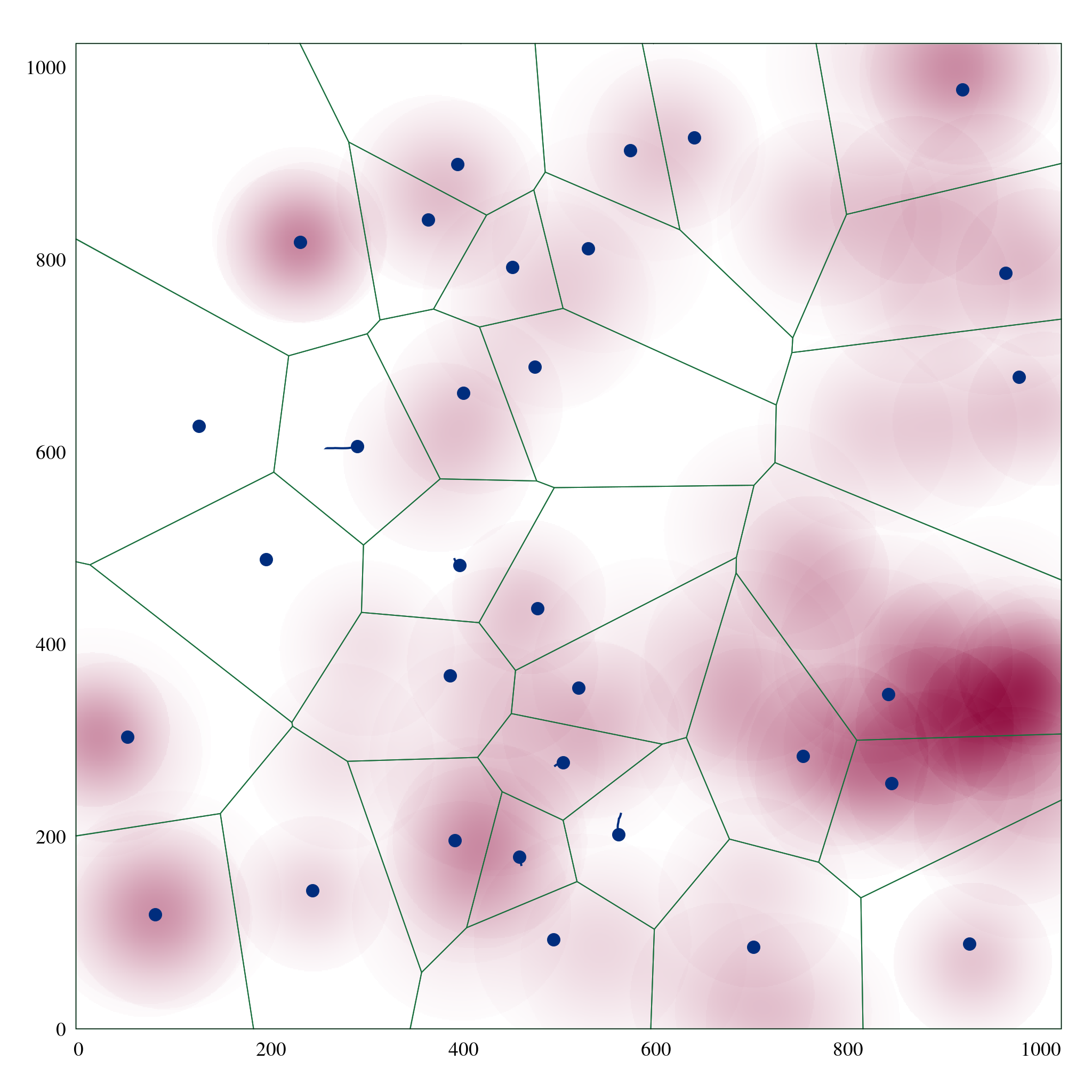}}
  \caption{Progression of the LPAC-K3 model on three real-world environments:
    The traffic signals, shown in the first column, form the features for generating the IDF.
    Each such feature is modeled as a 2-D Gaussian distribution with a uniformly sampled standard deviation $\sigma\in[40, 60]$ and scaled with a value sampled uniformly at random from $[6, 10]$.
    The number of features for Portland, San Francisco, and Boston are 9, 30, and 86, respectively.
    The robots in the swarm are initialized at random positions in the environment.
    The second and third columns show the cumulative observations of all robots up to time steps 100 and 600, respectively.
    The last column shows the positions of the robots at time step 1200 with the Voronoi cells of the robots on the entire IDF.
  \label{fig:realworld}}
\end{figure*}

\fgref{fig:realworld} shows the progression of the LPAC-K3 model in three environments from the dataset.
Since the performance of the coverage control algorithms is sensitive to the initial positions of the robots, we evaluated the performance of the algorithms with 10 different initial positions of the robots.
\fgref{fig:semantic_heatmap} shows the performance of the LPAC-K3 model with respect to the decentralized and centralized CVT algorithms.
The performance is measured as the ratio of the average cost with respect to the clairvoyant algorithm.
The environments are split into buckets of size 10 based on the number of features in each environment.
The results show that the LPAC-K3 model performs poorly, around 3.5 times the clairvoyant algorithm, on environments with a small number of features, as also observed in \scref{sc:generalization}.
The model, however, performs well in environments with more than 20 features, with an average performance of around 1.9 times the clairvoyant algorithm for more than 30 features.

The results show that the LPAC architecture performs well for environments that may be very different from the training distribution.
It is important to note that the LPAC-K3 model was not retrained or fine-tuned on the real-world dataset, and the model was trained on synthetic datasets with randomly generated features.
Furthermore, the clairvoyant algorithm has knowledge of the entire IDF, and the LPAC model is able to perform well with only local observations of the IDF and communication with neighboring robots.
The performance of the learned LPAC policy in the traffic light realistic environments further validates that the LPAC architecture does not merely overfit to synthetic uniform conditions, but can effectively address real-world scenarios with non-uniform feature distributions.

\subsection{Demonstration on Realistic Simulator}
We demonstrate our approach on a simulation framework that combines PX4, Gazebo, and ROS2 Humble to facilitate a direct and practical approach to prototyping aerial robots.
A large number of platforms use PX4, and thus, using PX4 ensures that code developed in simulation can be transferred to most real platforms with minimal modifications.
Gazebo introduces physics-based realism, including collision detection and sensor noise, reducing discrepancies between simulated experiments and real-world field tests.

\begin{figure*}[htbp]
  \centering
  \includegraphics[width=0.8\textwidth]{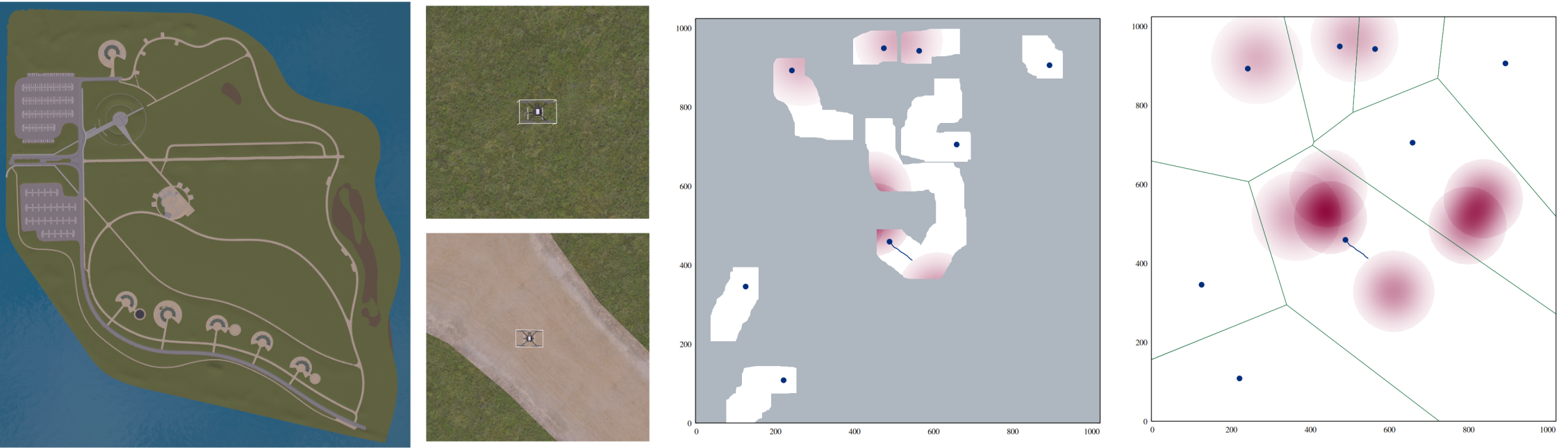}
  \caption{Gazebo and PX4 based simulation environment using ROS2. The environment is a \numproduct{250x250} square with 8 robots and 8 simulated features. \label{fig:sim}}
\end{figure*}

We deployed the Learnable Perception-Action-Communication (LPAC) policy operating within the integrated simulation environment.
In this demonstration, the system relies solely on GPS for localization, although the simulation setup allows for introducing additional sensors and more sophisticated perception methods.
The environment is a \numproduct{250x250} square with 8 robots and 8 features.
The effective communication range was set to \SIm{64} with all other parameters the same.
The \fgref{fig:sim} shows the simulation environment and the coverage of the robots at the final time-step.
Note that the performance of any coverage control policy is subject to the initial positions of the robots, as the field-of-view is limited.
Future work will involve performing large-scale simulations on various environments and more number of robots, along with real-world experiments, to validate the performance of the LPAC architecture.

\begin{figure}[t]
  \begin{tikzpicture}
	\pgfplotsset{
		colormap={violet}{color=(mDarkBlue) color=(white) color=(mDarkRed)}
	}
	\footnotesize
	\begin{axis}[
		xlabel={\small Number of Features},
		width=0.95\columnwidth,
		height=0.63\columnwidth,
		xmin=-0.5,
		xmax=8.5,
		ymin=-0.5,
		ymax=2.5,
		xtick={-0.5, 0.5, ..., 8.5},
		ytick={0, 1, 2},
		xticklabels={0, 10, ..., 90},
		yticklabels={D-CVT, C-CVT, LPAC-K3},
		y dir=reverse,
		yticklabel style={rotate=90},
		colorbar,
		colorbar style={
			yticklabel style={
				/pgf/number format/.cd,
				fixed,
				precision=2,
				fixed zerofill,
			},
			colormap name=violet,
		},
		enlargelimits=false,
		axis on top,
		point meta min=1.95,
		point meta max=3.55,
		]
		\addplot [matrix plot*,point meta=explicit] table [x=x,y=y,meta=C] {./figures/data/semantic_heatmap.dat};
	\end{axis}
\end{tikzpicture}
  \caption{Performance of coverage algorithms on the real-world traffic light datasets from 50 cities in the US spanning \SImSqrDim{1024}:
    The values in the heatmap are the ratio of the average costs of the coverage algorithms with respect to the clairvoyant algorithm.
    The 50 environments are organized in the heatmap in buckets of size 10, based on the number of features (traffic signals) in each environment.
    For each environment, 10 trials with random initial positions of 32 robots are run, and the average cost is computed.
    The LPAC-K3 model performs poorly on environments with a small number of features, as they are out of distribution for the model.
  However, the model is able to generalize well to environments with a large number of features, and it outperforms both the decentralized and centralized CVT algorithms, on average, for environments with more than 20 features.\label{fig:semantic_heatmap}}
\end{figure}

\subsection{Communication Analysis}
In this section, we analyze the performance of LPAC policies with different communication ranges, and discuss the communication requirements for both centralized and decentralized deployment of such policies.

\begin{figure}[t]
  \begin{tikzpicture}
	\pgfplotsset{
		colormap={violet}{color=(mDarkBlue) color=(white) color=(mDarkRed)}
	}
	\footnotesize
	\begin{axis}[
		view={0}{90},   
		xlabel={\small Communication Range (m)},
		width=0.96\columnwidth,
		height=0.65\columnwidth,
		xmin=-0.5,
		xmax=4.5,
		ymin=-0.5,
		ymax=2.5,
		xtick={0, 1, 2, 3, 4},
		ytick={0, 1, 2},
		xticklabels={128, 256, 512, 768, 1024},
		yticklabels={D-CVT, C-CVT, LPAC-K3},
		y dir=reverse,
		yticklabel style={rotate=90},
		colorbar,
		colorbar style={
			yticklabel style={
				/pgf/number format/.cd,
				fixed,
				precision=2,
				fixed zerofill,
			},
			colormap name=violet,
		},
		enlargelimits=false,
		axis on top,
		]
		\addplot [matrix plot*,point meta=explicit] table [x=x,y=y,meta=C] {./figures/data/comm_range_heatmap.dat};
	\end{axis}
\end{tikzpicture}
  \caption{Performance of the LPAC architecture with different communication ranges:
    Models with LPAC architecture, with 3 hops of communication, are trained on environments with 32 robots and \SImSqrDim{1024} size with different communication ranges.
    The models are evaluated on environments with the same size and number of robots but with different communication ranges.
    Counterintuitively, the LPAC architecture does not perform well when the communication range is closer to the size of the environment.
  It, however, performs well when the communication range is relatively smaller than the size of the environment, peaking at a communication range of \SIm{512}.\label{fig:comm_range_heatmap}}
\end{figure}

\subsubsection{Performance with Different Communication Ranges}
We trained models with the LPAC architecture with different communication ranges and compared them with the CVT algorithms.
\fgref{fig:comm_range_heatmap} shows the performance of the controllers as a ratio of the average cost with respect to the clairvoyant algorithm.
The same set of environments is used for all controllers and communication ranges.
The performance of the centralized CVT algorithm does not change with the communication range, as expected.
Whereas the decentralized CVT algorithm improves until a communication range of \SIm{512} and then stagnates.
This is also expected, as the Voronoi cell of a particular robot is generally not affected by the robots that are far away.
The performance of the decentralized CVT algorithm is still worse than the centralized one, even with a communication range of \SIm{1024}, as the robots only have knowledge of their own observations of the IDF.
In contrast, the centralized algorithm uses the combined observations of all robots.

The LPAC-K3 model outperforms the CVT algorithms for all communication ranges.
Interestingly, the performance of the LPAC-K3 model improves until a communication range of \SIm{512} and then decreases.
This is counterintuitive, as one would expect the performance to improve with a larger communication range and the performance to be closer to the clairvoyant algorithm with a communication range of \SIm{1024}.
We hypothesize three reasons for this behavior of the LPAC model.
First, the neighbor maps in the CNN architecture are of the fixed size of \numproduct{32x32}.
For a large communication range, there would be several robots in a cell of the neighbor map due to the low resolution of the map.
We take the sum of the values in the cell when we have multiple robots in a cell, leading to a loss of information.
Second, in a very similar way, the GNN architecture uses summation as the aggregation function, which is again a lossy operation.
The model may be poorly suited to disambiguating features coming from different robots when the communication range is large.
Third, the communication graph becomes almost fully connected as the communication range increases, and GNNs perform better with sparse graphs.
Investigating these hypotheses and improving the performance of the LPAC model with a larger communication range is an interesting direction for future work.
One possible solution is to select only the closest robot when a cell in the neighbor map has multiple robots and use a more sophisticated aggregation function in the GNN architecture.

Nevertheless, the LPAC model is reliable for communication ranges up to \SIm{512} and outperforms the CVT algorithms for all communication ranges.

\subsubsection{Bandwidth Requirements}

In the coverage control problem, each robot makes localized observations based on the sensor size and maintains an ego-centric local map over time.
In real-world deployment, each robot would need to process raw sensor images and generate importance values.
In our simulations, we simulate the generation of importance values to keep the system agnostic to the sensor hardware and the computer vision module.
We now analyze the communication bandwidth requirements for the baseline coverage control algorithms and the proposed LPAC-based policies.

{\bf D-CVT}: The decentralized-CVT algorithm is executed independently on each robot and requires only the positions of the neighboring robots.
This means that there in no communication requirement for a robot when there is no neighbor.
On the other hand, in the worst-case scenario, from a communication load perspective, a robot would have all other robots as neighbors.
Thus, the worst-case requirement is $\mathcal{O}(n)$ messages for a robot.
When all robots are within the communication radius of each other, the entire system will be exchanging $\mathcal O(n^2)$ messages.
The message is only the positions of the robots, which in the simplest case can be represented as a vector of two floating-point numbers.
Hence, the communication load for the D-CVT algorithm is very low.

{\bf C-CVT}: The centralized-CVT requires each robot to send their local maps and positions to a central server.
In our implementation, each robot maintains a local map of size \numproduct{256x256}.
This requires a message of $2^{16} + 2$ floating-point numbers.
The central server receives these messages from all $n$ robots.
A large-scale system can overload the communication bandwidth due to both large message size and the number of robots.

{\bf LPAC}: The LPAC policy can be deployed in both centralized and decentralized setting.
In the centralized setting, each robot communicates their input to the GNN module, which is the concatenation of the CNN output with the normalized position of the robots.
For our LPAC policy, this message is a ($2^{5} + 2$) size floating-point vector.
This is substantially lower than the C-CVT algorithm by a factor of approximately $2^{11}=2048$.
As the results show, the LPAC policy performs significantly better than the C-CVT algorithms, even with a lower communication requirement.

In the decentralized setting, the communication load is strongly tied to the architecture design, i.e., the number of hops, the number of GNN layers, and the latent size of the GNN layers.
In our primary LPAC policy, the input to the first GNN layer is a $34$-sized vector, and the output of all layers is a $2^8 = 256$-sized vector.
As described in \scref{sec:gnn} and in Equations~(\ref{eqn:yilk}) and~(\ref{eqn:agg_msg}),
the transmitted aggregated message is given by a set of vectors $(\mv y_i)_{lk}$ for a robot $i$.
Except for $(\mv y_i)_{00}$, which has a size of 34 and corresponds to the input to the GNN, all other vectors have size $2^8=256$.
The architecture has $L=5$ GNN layers with $K=3$ hops.
Thus, the total message size is $14 \times 2^8 + 34 = 3618$ floating-point numbers.
This is lower than the C-CVT algorithm by approximately a factor of $18$.
Similar to the D-CVT algorithm, a robot only communicates directly with its neighbors, and when all robots are neighbors to each other in the worst-case scenario, they make $\mathcal O(n)$ communications.
In practice, the number of neighbors is relatively low for the coverage control problem.
For environment of size \numproduct{1024x1024} with $32$ robots and $256$ communication radius, the average number of neighbors is $5.30$ with a standard deviation of $2.35$, averaged over simulations of LPAC policy on $100$ unique importance density fields and initial robot positions.

We highlight the following features of the GNN-based decentralized LPAC policy:
\emph{(i)}~The communication is completely decentralized and hence, there is no single point of failure or communication bottleneck.
\emph{(ii)}~The GNN abstracts the observations and received information into a fixed sized message, which is independent of the number of robots, i.e., it does not increase with the number of neighbors seen by a robot.
Hence, the system scales well with the number of robots.
\emph{(iii)}~The number of message exchanges is always with the neighboring robots.
In general, instead of peer-to-peer communication, a broadcasting protocol can also be used to reduce the communication load~\cite{agarwal2023asynchronous,BlumenkampGNN2022}. 
\emph{(iv)}~The LPAC-policy performs better than the centralized-CVT baseline even though the communication requirement is significantly lower.

\section{Discussion} \label{sc:conclusion}
This article presented a learning-based approach to the decentralized coverage control problem, where a robot swarm is deployed in an environment with an underlying importance density field (IDF), not known \textit{a priori}.
The problem requires a robot swarm to be placed optimally in the environment so that robots efficiently cover the IDF with their sensors.
The problem is challenging in the decentralized setting as the robots need to decide what to communicate with the neighboring robots and how to incorporate the received information with their own observations.
However, in contrast to a centralized system, a decentralized system is robust to individual failures and scales well with large teams in larger environments.

Motivated by the advantages of a decentralized system, the article proposed a learnable Perception-Action-Communication (LPAC) architecture.
The perception module uses a convolutional neural network (CNN) to process the local observations of the robot and generates an abstract representation of the local observations.
In the communication module, the GNN computes the messages to be sent to other robots and incorporates the received messages with the output of the perception module.
Finally, a shallow multi-layer perceptron (MLP) in the action module computes the velocity controls for each robot.
All three modules are executed independently on each robot, and the communication model enables collaboration in the robot swarm with the aid of the GNN.

Models based on the LPAC architecture were trained using imitation learning with a clairvoyant algorithm that uses the centroidal Voronoi tessellation (CVT).
We extensively evaluated the LPAC models with the decentralized and the centralized CVT algorithms and showed that the models consistently outperform both CVT algorithms.
It is interesting to note that the centralized CVT algorithm knows the positions of all robots to compute the Voronoi partition; it also has the cumulative knowledge of the IDF from all robots to compute the centroids of the Voronoi cell.
Yet, the decentralized LPAC models do better, which indicates that they are able to learn features beyond what is used by the CVT-based algorithms.

The evaluations further establish that the models are able to perform well in environments with a larger number of robots and features.
The models also transfer to larger environments with bigger robot swarms.
The models were robust to noisy estimates of the robot positions, with only a small degradation in performance as the noise levels increased.
A real-world dataset was generated from traffic signals as features to generate the IDF.
The models were tested without any further training or fine-tuning.
Except for the cases when the number of features was very low, the models outperformed the CVT algorithms, even though these datasets have IDFs that are very different from the training set of randomly and uniformly generated features.

In \textit{conclusion}, the results establish that the LPAC models are transferable to larger environments, scalable to larger robot swarms, and robust to noisy position estimates and varying IDFs for the decentralized coverage control problem.
The success of the LPAC architecture indicates that it is a promising approach to solving other decentralized navigation problems.
Furthermore, the architecture learns a decentralized policy from a centralized clairvoyant algorithm, thereby alleviating the need for carefully designed decentralized algorithms.

Future work involves testing the models in physical experiments with asynchronous and decentralized communication infrastructure~\cite{agarwal2023asynchronous}.
Additional testing with communication dropouts, latency, and robot failures is essential to characterize the resiliency of the system to failures.
Another essential task is to investigate further the poor performance of the models with large communication ranges, which may require minor changes in how aggregation is performed in the GNN architecture.
When we do not have any prior knowledge of the IDF, the algorithm can benefit from exploring the environment to discover more features.
In contrast, in this article, the robots react to the features they observe without any explicit exploration.
However, the exploration is antithetical to the coverage control problem, as the former requires visiting new regions, whereas the latter is focused on providing coverage to the already observed high-importance regions.
In future work, we plan to investigate the trade-off between exploration and coverage control.
The promising results motivate the theoretical study of the LPAC architecture in relation to transferability, robustness, and scalability.
Additionally, the use of the clairvoyant algorithm to generate actions needs further investigation to ensure general applicability to other decentralized navigation problems and to determine the approach's limitations.

The primary motivation for the design of the PAC architecture was to solve decentralized navigation problems with robot swarms where there is a collaborative goal to be achieved.
We plan to investigate more navigation problems and assess the generalizability of the LPAC architecture.

\section*{Acknowledgments}
The map data used to generate the dataset in \scref{sc:real-world} is copyrighted by OpenStreetMap contributors and is available from \url{https://www.openstreetmap.org}.

\IEEEtriggeratref{36}
\bibliographystyle{IEEEtran}

\end{document}